\documentclass[manuscript,screen]{acmart}

\AtBeginDocument{%
  }

\setcopyright{acmcopyright}
\copyrightyear{2023}
\acmYear{2023}
\acmDOI{10.1145/3626235}

\acmPrice{15.00}
\acmISBN{978-1-4503-XXXX-X/18/06}

\usepackage{bm}
\usepackage{subcaption}
\newcommand{\bepsilon}{{\boldsymbol{\epsilon}}}
\usepackage{multirow}
\usepackage{mathtools,amssymb,euscript,yfonts,latexsym}
\usepackage[capitalise]{cleveref}
\usepackage{makecell}
\usepackage{amsmath}
\usepackage{color,colortbl}
\usepackage{textcomp}
\definecolor{LightCyan}{rgb}{0.8,1,1}
\definecolor{Gray}{gray}{0.9}

\newcommand{\E}{\mathbb{E}}

\newcommand{\Ea}[1]{\E\left[#1\right]}
\newcommand{\Eb}[2]{\E_{#1}\!\left[#2\right]}

\makeatletter
\usepackage{xspace}
\def\@onedot{\ifx\@let@token.\else.\null\fi\xspace}
\DeclareRobustCommand\onedot{\futurelet\@let@token\@onedot}
\def\eg{\emph{e.g}\onedot}

\def\ie{\emph{i.e}\onedot}

\def\cf{\emph{cf}\onedot}

\def\aka{a.k.a\onedot}

\begin{document}

\title{Diffusion Models: A Comprehensive Survey of Methods and Applications}

\author{Ling Yang}
\affiliation{%
  \institution{Peking University}
  \country{China}
}
\email{yangling0818@163.com}

\author{Zhilong Zhang}
\authornote{Contributed equally.}
\affiliation{%
  \institution{Peking University}
  \country{China}
}
\email{zhilong.zhang@bjmu.edu.cn}

\author{Yang Song}
\affiliation{%
  \institution{OpenAI}
  \country{USA}
}
\email{songyang@openai.com}

\author{Shenda Hong}
\affiliation{%
  \institution{Peking University}
  \country{China}
}
\email{hongshenda@pku.edu.cn}

\author{Runsheng Xu}
\affiliation{%
  \institution{University of California, Los Angeles}
  \country{USA}
}
\email{rxx3386@ucla.edu}

\author{Yue Zhao}
\affiliation{%
  \institution{Carnegie Mellon University}
  \country{USA}
}
\email{zhaoy@cmu.edu}


\author{Wentao Zhang}
\affiliation{%
  \institution{Peking University}
  \country{China}
}
\email{wentao.zhang@pku.edu.cn}

\author{Bin Cui}
\affiliation{%
  \institution{Peking University}
  \country{China}
}
\email{bin.cui@pku.edu.cn}

\author{Ming-Hsuan Yang}
\authornote{Ling Yang, Wentao Zhang, Bin Cui, and Ming-Hsuan Yang are corresponding authors.}
\affiliation{%
  \institution{University of California at Merced}
  \country{USA}
}
\email{mhyang@ucmerced.edu}

\renewcommand{\shortauthors}{Yang et al.}

\begin{abstract}
Diffusion models have emerged as a powerful new family of deep generative models with record-breaking performance in many applications, including image synthesis, video generation, and molecule design. In this survey, we provide an overview of the rapidly expanding body of work on diffusion models, categorizing the research into three key areas: efficient sampling, improved likelihood estimation, and handling data with special structures. We also discuss the potential for combining diffusion models with other generative models for enhanced results. We further review the wide-ranging applications of diffusion models in fields spanning from computer vision, natural language processing, temporal data modeling, to interdisciplinary applications in other scientific disciplines. This survey aims to provide a contextualized, in-depth look at the state of diffusion models, identifying the key areas of focus and pointing to potential areas for further exploration. Github: \href{https://github.com/YangLing0818/Diffusion-Models-Papers-Survey-Taxonomy}{https://github.com/YangLing0818/Diffusion-Models-Papers-Survey-Taxonomy}.

\end{abstract}

\ccsdesc[500]{Computing methodologies~Computer vision tasks}
\ccsdesc[300]{Computing methodologies~Natural language generation}
\ccsdesc{Computing methodologies~Machine learning approaches}


\keywords{Generative Models, Diffusion Models, Score-Based Generative Models, Stochastic Differential Equations}

\maketitle
\newpage
\tableofcontents
\section{Introduction}
\begin{figure*}[htp]
\centering
    \includegraphics[width=0.93\linewidth]{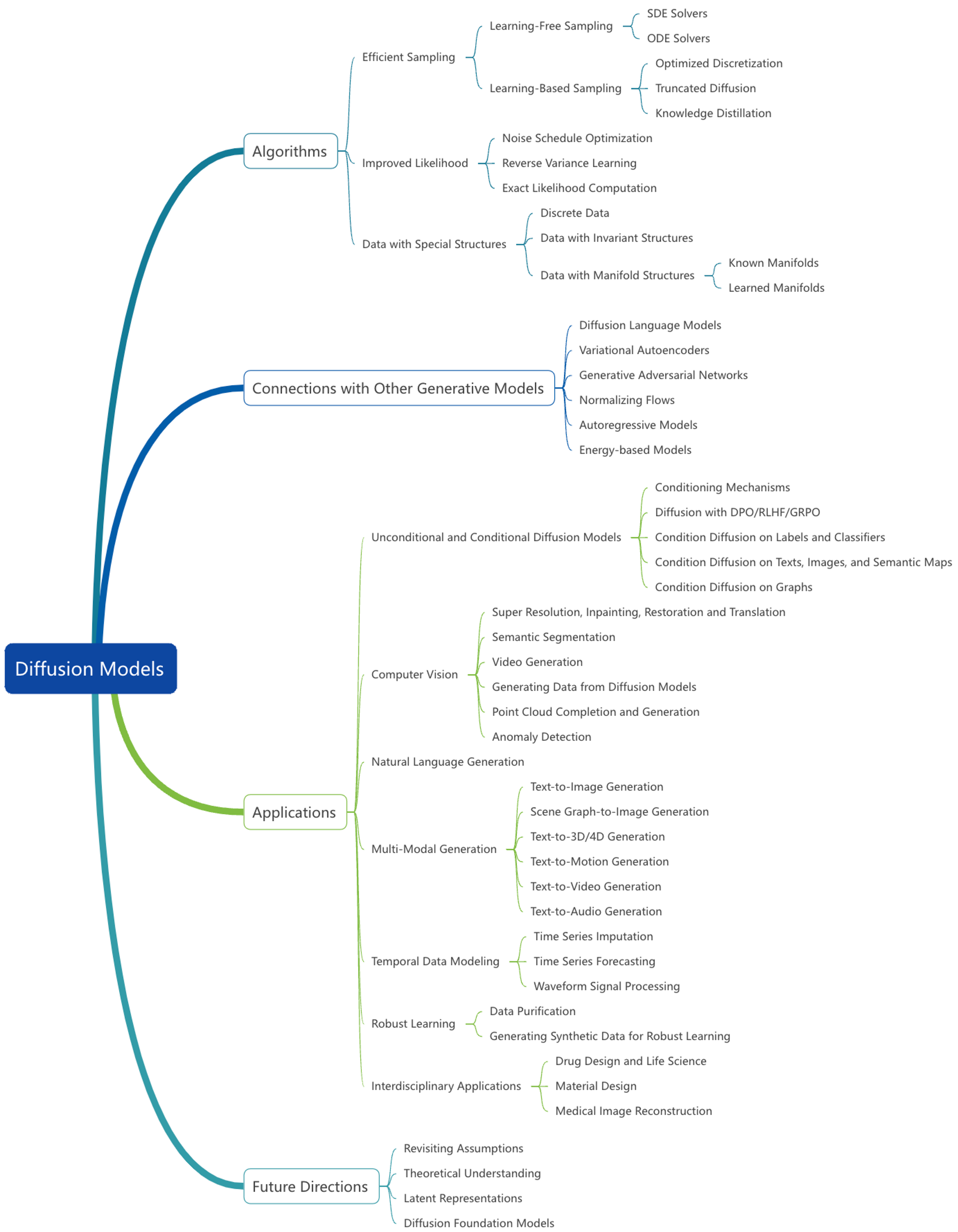}
        \caption{Taxonomy of diffusion models variants (in \crefrange{sec3}{sec5}), connections with other generative models (in \cref{sec6}), applications of diffusion models (in \cref{sec7}), and future directions (in \cref{sec8}).}
\label{fig:intro}
\end{figure*}

Diffusion models~\cite{sohl2015deep,song2019generative,ho2020denoising,song2020score} have emerged as the new state-of-the-art family of deep generative models. They have broken the long-time dominance of generative adversarial networks (GANs) \cite{goodfellow2014generative} in the challenging task of image synthesis \cite{song2019generative,song2020score,ho2020denoising,dhariwal2021diffusion} and have also shown potential in a variety of domains, ranging from computer vision \cite{cai2020learning,luo2021score,meng2021sdedit,amit2021segdiff,kawar2021stochastic,zimmermann2021score,whang2022deblurring,kim2021diffusemorph,brempong2022denoising,yang2022diffusion,yang2022lossy,Li2022SRDiffSI,baranchuk2021label,ozbey2022unsupervised,zhao2022egsde,saharia2022image,ho2022cascaded,saharia2022palette,ho2022video}, natural language processing \cite{austin2021structured,hoogeboom2021argmax,savinov2021step,li2022diffusion,yu2022latent}, temporal data modeling \cite{rasul2020multivariate,tashiro2021csdi,chen2020wavegrad,kong2020diffwave,yan2021scoregrad,alcaraz2022diffusion}, multi-modal modeling \cite{avrahami2022blended,saharia2022photorealistic,zhu2022discrete,ramesh2022hierarchical,rombach2022high}, robust machine learning \cite{blau2022threat,kawar2022enhancing,wang2022guided,yoon2021adversarial,carlini2022certified}, to interdisciplinary applications in fields such as computational chemistry \cite{jing2022torsional,hoogeboom2022equivariant,lee2022proteinsgm,anand2022protein,luo2022antigen,xie2021crystal,lee2022exploring} and medical image reconstruction \cite{song2021solving,chung2022come,chung2022mr,chung2022score,peng2022towards,mei2022metal,cao2022high,xie2022measurement,dar2022adaptive}.

Numerous methods have been developed to improve diffusion models, either by enhancing empirical performance~\cite{nichol2021improved,song2020denoising,song2020improved} or by extending the model's capacity from a theoretical perspective~\cite{song2020score,song2021maximum,lu2022dpm,lu2022maximum,zhang2022fast}. Over the past two years, the body of research on diffusion models has grown significantly, making it increasingly challenging for new researchers to stay abreast of the recent developments in the field. Additionally, the sheer volume of work can obscure major trends and hinder further research progress. This survey aims to address these problems by providing a comprehensive overview of the state of diffusion model research, categorizing various approaches, and highlighting key advances. We hope this survey to serve as a helpful entry point for researchers new to the field while providing a broader perspective for experienced researchers.

In this paper, we first explain the foundations of diffusion models (\cref{sec2}), providing a brief but self-contained introduction to three predominant formulations: denoising diffusion probabilistic models (DDPMs) \cite{sohl2015deep,ho2020denoising}, score-based generative models (SGMs) \cite{song2019generative,song2020improved}, and stochastic differential equations (Score SDEs) \cite{song2020score,song2021maximum,karras2022elucidating}. Key to all these approaches is to progressively perturb data with intensifying random noise (called the ``diffusion'' process), then successively remove noise to generate new data samples. We clarify how they work under the same principle of diffusion and explain how these three models are connected and can be reduced to one another.

Next, we present a taxonomy of recent research that maps out the field of diffusion models, categorizing it into three key areas: efficient sampling (\cref{sec3}), improved likelihood estimation (\cref{sec4}), and methods for handling data with special structures (\cref{sec5}), such as relational data, data with permutation/rotational invariance, and data residing on manifolds. We further examine the models by breaking each category into more detailed sub-categories, as illustrated in \cref{fig:intro}. In addition, we discuss the connections of diffusion models to other deep generative models (\cref{sec6}), including variational autoencoders (VAEs) \cite{kingma2013auto,rezende2014stochastic}, generative adversarial networks (GANs) \cite{goodfellow2014generative}, normalizing flows \cite{rippel2013high,dinh2014nice,dinh2017density,papamakarios2021normalizing}, autoregressive models \cite{oord2016pixelrnn}, and energy-based models (EBMs) \cite{song2021train,lecun2006tutorial}. By combining these models with diffusion models, researchers have the potential to achieve even stronger performance.

Following that, our survey reviews six major categories of application that diffusion models have been applied to in the existing research (\cref{sec7}): computer vision, natural language process, temporal data modeling, multi-modal learning, robust learning, and interdisciplinary applications. For each task, we provide a definition, describe how diffusion models can be employed to address it and summarize relevant previous work. We conclude our paper (\cref{sec8,sec9}) by providing an outlook on possible future directions for this exciting new area of research.

\section{Foundations of Diffusion Models}\label{sec2}

\begin{figure*}
    \centering
    \includegraphics[width=\linewidth]{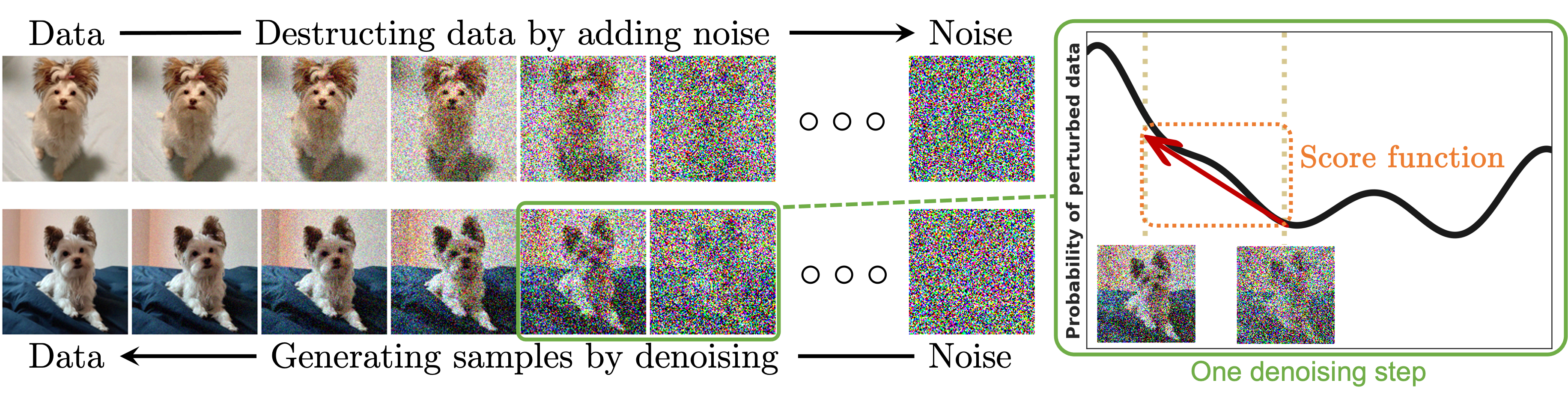}
    \caption{Diffusion models smoothly perturb data by adding noise, then reverse this process to generate new data from noise. Each denoising step in the reverse process typically requires estimating the score function (see the illustrative figure on the right), which is a gradient pointing to the directions of data with higher likelihood and less noise. \label{fig:schematic}}
\end{figure*}

Diffusion models are a family of probabilistic generative models that progressively destruct data by injecting noise, then learn to reverse this process for sample generation. We present the intuition of diffusion models in \cref{fig:schematic}. Current research on diffusion models is mostly based on three predominant formulations: denoising diffusion probabilistic models (DDPMs)~\cite{sohl2015deep,ho2020denoising,nichol2021improved}, score-based generative models (SGMs)~\cite{song2019generative,song2020improved}, and stochastic differential equations (Score SDEs)~\cite{song2020score,song2021maximum}. We give a self-contained introduction to these three formulations in this section, while discussing their connections with each other along the way.

\subsection{Denoising Diffusion Probabilistic Models (DDPMs)}\label{sec:back_ddpm}
A \emph{denoising diffusion probabilistic model} (DDPM) \cite{sohl2015deep,ho2020denoising} makes use of two Markov chains: a forward chain that perturbs data to noise, and a reverse chain that converts noise back to data. The former is typically hand-designed with the goal to transform any data distribution into a simple prior distribution (\eg, standard Gaussian), while the latter Markov chain reverses the former by learning transition kernels parameterized by deep neural networks. New data points are subsequently generated by first sampling a random vector from the prior distribution, followed by ancestral sampling through the reverse Markov chain~\cite{koller2009probabilistic}.

Formally, given a data distribution $\mathbf{x}_0 \sim q(\mathbf{x}_0)$, the forward Markov process generates a sequence of random variables $\mathbf{x}_1, \mathbf{x}_2 \dots \mathbf{x}_T$ with transition kernel $q(\mathbf{x}_t\mid\mathbf{x}_{t-1})$. Using the chain rule of probability and the Markov property, we can factorize the joint distribution of $\mathbf{x}_1, \mathbf{x}_2 \dots \mathbf{x}_T$ conditioned on $\mathbf{x}_0$, denoted as $q(\mathbf{x}_1, \ldots, \mathbf{x}_T\mid\mathbf{x}_0)$, into
\begin{align}
    q(\mathbf{x}_1, \ldots, \mathbf{x}_T\mid\mathbf{x}_0) &= \prod_{t=1}^{T} q(\mathbf{x}_t\mid\mathbf{x}_{t-1}).\label{eq:forward_joint}
\end{align}
In DDPMs, we handcraft the transition kernel $q(\mathbf{x}_t \mid \mathbf{x}_{t-1})$ to incrementally transform the data distribution $q(\mathbf{x}_0)$ into a tractable prior distribution. One typical design for the transition kernel is Gaussian perturbation, and the most common choice for the transition kernel is
\begin{align}
    q(\mathbf{x}_t\mid\mathbf{x}_{t-1}) &= \mathcal{N}(\mathbf{x}_t; \sqrt{1-\beta_t} \mathbf{x}_{t-1}, \beta_t \mathbf{I}), \label{eq:2}
\end{align}
where $\beta_t \in (0,1)$ is a hyperparameter chosen ahead of model training. We use this kernel to simply our discussion here, although other types of kernels are also applicable in the same vein. As observed by Sohl-Dickstein et al. (2015) \cite{sohl2015deep}, this Gaussian transition kernel allows us to marginalize the joint distribution in \cref{eq:forward_joint} to obtain the analytical form of $q(\mathbf{x}_t \mid \mathbf{x}_0)$ for all $t \in \{0,1,\cdots,T\}$. Specifically, with $\alpha_t \coloneqq 1 - \beta_t$ and $\bar{\alpha}_t \coloneqq \prod_{s=0}^{t} \alpha_s$, we have
\begin{align}
        q(\mathbf{x}_t\mid\mathbf{x}_0) &= \mathcal{N}(\mathbf{x}_t; \sqrt{\bar{\alpha}_t} \mathbf{x}_0, (1-\bar{\alpha}_t) \mathbf{I}) \label{eq:3}.
\end{align}
Given $\mathbf x_0$, we can easily obtain a sample of $\mathbf{x}_t$ by sampling a Gaussian vector $\bepsilon \sim \mathcal{N}(\mathbf{0}, \mathbf{I})$ and applying the transformation
\begin{align}
    \mathbf{x}_t &=\sqrt{\bar{\alpha}_t} \mathbf{x}_0 + \sqrt{1-\bar{\alpha}_t} \bepsilon. \label{eq:4}
\end{align}
When \(\bar{\alpha}_T \approx 0 \), \(\mathbf{x}_{T}\) is almost Gaussian in distribution, so we have $q(\mathbf x_T) \coloneqq \int q(\mathbf x_T \mid \mathbf x_0) q(\mathbf x_0) \textrm{d} \mathbf x_0 \approx \mathcal{N}(\mathbf x_T; \mathbf{0}, \mathbf{I})$.

Intuitively speaking, this forward process slowly injects noise to data until all structures are lost. For generating new data samples, DDPMs start by first generating an unstructured noise vector from the prior distribution (which is typically trivial to obtain), then gradually remove noise therein by running a learnable Markov chain in the reverse time direction. Specifically, the reverse Markov chain is parameterized by a prior distribution $p(\mathbf x_T) = \mathcal{N}(\mathbf x_T; \mathbf{0}, \mathbf{I})$ and a learnable transition kernel $p_\theta(\mathbf{x}_{t-1}\mid\mathbf{x}_t)$. We choose the prior distribution $p(\mathbf x_T) = \mathcal{N}(\mathbf x_T; \mathbf 0, \mathbf I)$ because the forward process is constructed such that $q(\mathbf x_T) \approx \mathcal{N}(\mathbf x_T; \mathbf{0}, \mathbf{I})$. The learnable transition kernel $p_\theta(\mathbf{x}_{t-1}\mid\mathbf{x}_t)$ takes the form of
\begin{align}
  p_\theta(\mathbf x_{t-1}\mid\mathbf x_t) &= \mathcal{N}(\mathbf x_{t-1}; \mu_{\theta}(\mathbf x_t, t), \Sigma_{\theta}(\mathbf x_t, t))
  \label{4}
\end{align}
where $\theta$ denotes model parameters, and the mean $\mu_{\theta}(\mathbf x_t, t)$ and variance $\Sigma_{\theta}(\mathbf x_t, t)$ are parameterized by deep neural networks. With this reverse Markov chain in hand, we can generate a data sample $\mathbf x_0$ by first sampling a noise vector $\mathbf x_T \sim p(\mathbf x_T)$, then iteratively sampling from the learnable transition kernel $\mathbf x_{t-1} \sim p_\theta(\mathbf x_{t-1} \mid \mathbf x_t)$ until $t = 1$.

Key to the success of this sampling process is training the reverse Markov chain to match the actual time reversal of the forward Markov chain. That is, we have to adjust the parameter $\theta$ so that the joint distribution of the reverse Markov chain $p_\theta(\mathbf x_0, \mathbf x_1, \cdots, \mathbf x_T)\coloneqq p(\mathbf x_T)\prod_{t=1}^T p_\theta(\mathbf x_{t-1}\mid\mathbf x_t)$ closely approximates that of the forward process $q(\mathbf x_0, \mathbf x_1, \cdots, \mathbf x_T) \coloneqq q(\mathbf x_0) \prod_{t=1}^T q(\mathbf x_t \mid \mathbf x_{t-1})$ (\cref{eq:forward_joint}). This is achieved by minimizing the Kullback-Leibler (KL) divergence between these two:
\begin{align}
    &\operatorname{KL}(q(\mathbf x_0, \mathbf x_1, \cdots, \mathbf x_T) \mid\mid p_\theta(\mathbf x_0, \mathbf x_1, \cdots, \mathbf x_T))\\
    \stackrel{(i)}{=} &-\mathbb{E}_{q(\mathbf x_0, \mathbf x_1, \cdots, \mathbf x_T)}[\log p_\theta(\mathbf x_0, \mathbf x_1, \cdots, \mathbf x_T)] + \text{const}\\
    \stackrel{(ii)}{=} & \underbrace{\mathbb{E}_{q(\mathbf x_0, \mathbf x_1, \cdots, \mathbf x_T)}\bigg[ -\log p(\mathbf x_T) - \sum_{t=1}^T \log \frac{p_\theta(\mathbf x_{t-1}\mid\mathbf x_t)}{q(\mathbf x_t\mid\mathbf x_{t-1})} \bigg]}_{\coloneqq -L_{\textrm{VLB}}(\mathbf x_0)} + \text{const} \label{eq:7} \\
    \stackrel{(iii)}{\geq} & \Ea{-\log p_\theta(\mathbf x_0)} + \text{const},
\end{align}
where (i) is from the definition of KL divergence, (ii) is from the fact that $q(\mathbf x_0, \mathbf x_1, \cdots, \mathbf x_T)$ and $p_\theta(\mathbf x_0, \mathbf x_1, \cdots, \mathbf x_T)$ are both products of distributions, and (iii) is from Jensen's inequality. The first term in \cref{eq:7} is the variational lower bound (VLB) of the log-likelihood of the data $\mathbf x_0$, a common objective for training probabilistic generative models. We use ``$\text{const}$'' to symbolize a constant that does not depend on the model parameter $\theta$ and hence does not affect optimization. The objective of DDPM training is to maximize the VLB (or equivalently, minimizing the negative VLB), which is particularly easy to optimize because it is a sum of independent terms, and can thus be estimated efficiently by Monte Carlo sampling~\cite{metropolis1949monte} and optimized effectively by stochastic optimization~\cite{spall2012stochastic}.

Ho et al. (2020) \cite{ho2020denoising} propose to reweight various terms in $L_\textrm{VLB}$ for better sample quality and noticed an important equivalence between the resulting loss function and the training objective for noise-conditional score networks (NCSNs), one type of \emph{score-based generative models}, in Song and Ermon~\cite{song2019generative}. The loss in \cite{ho2020denoising} takes the form of
\begin{align}
    \Eb{t \sim \mathcal{U}\llbracket 1,T \rrbracket, \mathbf x_0 \sim q(\mathbf x_0), \bepsilon \sim \mathcal{N}(\mathbf{0},\mathbf{I})}{ \lambda(t)  \left\| \bepsilon - \bepsilon_\theta(\mathbf{x}_t, t) \right\|^2} \label{8}
\end{align}
where $\lambda(t)$ is a positive weighting function, $\mathbf x_t$ is computed from $\mathbf x_0$ and $\bepsilon$ by \cref{eq:4}, $\mathcal{U}\llbracket 1, T \rrbracket$ is a uniform distribution over the set $\{1, 2, \cdots, T\}$, and \(\bepsilon_{\theta}\) is a deep neural network with parameter $\theta$ that predicts the noise vector \(\bepsilon\) given \(\mathbf{x}_{t}\) and $t$. This objective reduces to \cref{eq:7} for a particular choice of the weighting function $\lambda(t)$, and has the same form as the loss of denoising score matching over multiple noise scales for training score-based generative models~\cite{song2019generative}, another formulation of diffusion models to be discussed in the next section.

\subsection{Score-Based Generative Models (SGMs)}\label{sec:back_sgm}
At the core of score-based generative models~\cite{song2019generative,song2020improved} is the concept of \emph{(Stein) score} (\aka, score or score function)~\cite{Hyvrinen2005EstimationON}. Given a probability density function $p(\mathbf x)$, its score function is defined as the gradient of the log probability density $\nabla_{\mathbf x} \log p(\mathbf x)$. Unlike the commonly used \emph{Fisher score} $\nabla_\theta \log p_\theta(\mathbf x)$ in statistics, the Stein score considered here is a function of the data $\mathbf x$ rather than the model parameter $\theta$. It is a vector field that points to directions along which the probability density function has the largest growth rate.

The key idea of score-based generative models (SGMs) \cite{song2019generative} is to perturb data with a sequence of intensifying Gaussian noise and jointly estimate the score functions for all noisy data distributions by training a deep neural network model conditioned on noise levels (called a noise-conditional score network, NCSN, in \cite{song2019generative}). Samples are generated by chaining the score functions at decreasing noise levels with score-based sampling approaches, including Langevin Monte Carlo~\cite{parisi1981correlation,grenander1994representations,song2019generative,song2020score,JolicoeurMartineau2021AdversarialSM}, stochastic differential equations~\cite{song2020score,jolicoeur2021gotta}, ordinary differential equations~\cite{song2020score,song2021maximum,lu2022dpm,zhang2022fast,karras2022elucidating}, and their various combinations~\cite{song2020score}. Training and sampling are completely decoupled in the formulation of score-based generative models, so one can use a multitude of sampling techniques after the estimation of score functions.

With similar notations in \cref{sec:back_ddpm}, we let $q(\mathbf x_0)$ be the data distribution, and $0 < \sigma_{1} < \sigma_{2} < \cdots < \sigma_t < \cdots < \sigma_T$ be a sequence of noise levels. A typical example of SGMs involves perturbing a data point $\mathbf x_0$ to $\mathbf x_t$ by the Gaussian noise distribution $q(\mathbf x_t \mid \mathbf x_0) = \mathcal{N}(\mathbf x_t ; \mathbf x_0, \mathbf \sigma_t^2 I)$. This yields a sequence of noisy data densities $q(\mathbf x_1), q(\mathbf x_2), \cdots, q(\mathbf x_T)$, where $q(\mathbf x_t) \coloneqq \int q(\mathbf x_t) q(\mathbf x_0) \textrm{d} \mathbf x_0$. A noise-conditional score network is a deep neural network $\mathbf s_\theta(\mathbf x, t)$ trained to estimate the score function $\nabla_{\mathbf x_t} \log q(\mathbf x_t)$. Learning score functions from data (\aka, score estimate) has established techniques such as score matching~\cite{Hyvrinen2005EstimationON}, denoising score matching~\cite{vincent2011connection,raphan2007learning,raphan2011least}, and sliced score matching~\cite{song2019sliced}, so we can directly employ one of them to train our noise-conditional score networks from perturbed data points. For example, with denoising score matching and similar notations in \cref{8}, the training objective is given by
\begin{align}
    &\Eb{t \sim \mathcal{U}\llbracket 1, T\rrbracket, \mathbf x_0 \sim q(\mathbf x_0), \mathbf x_t \sim q(\mathbf x_t \mid \mathbf x_0)}{\lambda(t) \sigma_t^2 \left\| \nabla_{\mathbf x_t} \log q(\mathbf x_t) - \mathbf s_\theta(\mathbf x_t, t) \right\|^2}\\
    \stackrel{(i)}{=}&\Eb{t \sim \mathcal{U}\llbracket 1, T\rrbracket, \mathbf x_0 \sim q(\mathbf x_0), \mathbf x_t \sim q(\mathbf x_t \mid \mathbf x_0)}{\lambda(t) \sigma_t^2 \left\| \nabla_{\mathbf x_t} \log q(\mathbf x_t \mid \mathbf x_0) - \mathbf s_\theta(\mathbf x_t, t) \right\|^2} + \text{const}\\
    \stackrel{(ii)}{=}&\Eb{t \sim \mathcal{U}\llbracket 1, T\rrbracket, \mathbf x_0 \sim q(\mathbf x_0), \mathbf x_t \sim q(\mathbf x_t \mid \mathbf x_0)}{\lambda(t) \left\| -\frac{\mathbf x_t - \mathbf x_0}{\sigma_t} - \sigma_t \mathbf s_\theta(\mathbf x_t, t) \right\|^2} + \text{const}\\
    \stackrel{(iii)}{=}&\Eb{t \sim \mathcal{U}\llbracket 1, T\rrbracket, \mathbf x_0 \sim q(\mathbf x_0), \bepsilon \sim \mathcal{N}(\mathbf 0, \mathbf I)}{\lambda(t) \left\| \bepsilon + \sigma_t \mathbf s_\theta(\mathbf x_t, t) \right\|^2} + \text{const} \label{eq:dsm},
\end{align}
where (i) is derived by \cite{vincent2011connection}, (ii) is from the assumption that $q(\mathbf x_t \mid \mathbf x_0) = \mathbf{N}(\mathbf x_t; \mathbf x_0, \sigma_t^2 \mathbf I)$, and (iii) is from the fact that $\mathbf x_t = \mathbf x_0 + \sigma_t \bepsilon$. Again, we denote by $\lambda(t)$ a positive weighting function, and ``const'' a constant that does not depend on the trainable parameter $\theta$. Comparing \cref{eq:dsm} with \cref{8}, it is clear that the training objectives of DDPMs and SGMs are equivalent, once we set $\mathbf \bepsilon_\theta(\mathbf x, t) = -\sigma_t \mathbf s_\theta(\mathbf x, t)$. Moreover, one can generalize the score matching with higher order. High-order derivatives of data density provide additional local information about the data distribution. Meng et al. \cite{meng2021estimating} proposes a generalized denoising score matching method to efficiently estimate the high-order score function. The proposed model can improve the mixing speed of Langevin dynamics and thus the sampling efficiency of diffusion models.

For sample generation, SGMs leverage iterative approaches to produce samples from $\mathbf s_\theta(\mathbf x, T)$, $\mathbf s_\theta(\mathbf x, T-1)$, $\cdots, \mathbf s_\theta(\mathbf x, 0)$ in succession. Many sampling approaches exist due to the decoupling of training and inference in SGMs, some of which are discussed in the next section. Here we introduce the first sampling method for SGMs, called annealed Langevin dynamics (ALD) \cite{song2019generative}. Let $N$ be the number of iterations per time step and $s_t>0$ be the step size. We first initialize ALD with $\mathbf x_T^{(N)} \sim \mathcal{N}(\mathbf 0, \mathbf I)$, then apply Langevin Monte Carlo for $t = T, T-1, \cdots, 1$ one after the other. At each time step $0 \leq t < T$, we start with $\mathbf x_t^{(0)} = \mathbf x_{t+1}^{(N)}$, before iterating according to the following update rule for $i=0,1,\cdots,N-1$:
\begin{align*}
    \bepsilon^{(i)} &\leftarrow \mathcal{N}(\mathbf 0, \mathbf I)\\
    \mathbf x_t^{(i+1)} &\leftarrow \mathbf x_{t}^{(i)} + \frac{1}{2} s_t \mathbf s_\theta(\mathbf x_{t}^{(i)}, t) + \sqrt{s_t} \bepsilon^{(i)}.
\end{align*}
The theory of Langevin Monte Carlo \cite{parisi1981correlation} guarantees that as $s_t \to 0$ and $N \to \infty$, $\mathbf x_0^{(N)}$ becomes a valid sample from the data distribution $q(\mathbf x_0)$.

\subsection{Stochastic Differential Equations (Score SDEs)}\label{sec:back_sde}
DDPMs and SGMs can be further generalized to the case of infinite time steps or noise levels, where the perturbation and denoising processes are solutions to stochastic differential equations (SDEs). We call this formulation \emph{Score SDE}~\cite{song2020score}, as it leverages SDEs for noise perturbation and sample generation, and the denoising process requires estimating score functions of noisy data distributions.

Score SDEs perturb data to noise with a diffusion process governed by the following stochastic differential equation (SDE) \cite{song2020score}:
\begin{align}
    \textrm{d} \mathbf{x} = \mathbf{f}(\mathbf{x},t) \textrm{d} t + g(t) \textrm{d} \mathbf{w}
    \label{9}
\end{align}
where $\mathbf f(\mathbf x, t)$ and $g(t)$ are diffusion and drift functions of the SDE, and \(\mathbf{w}\) is a standard Wiener process (\aka, Brownian motion). The forward processes in DDPMs and SGMs are both discretizations of this SDE. As demonstrated in Song et al. (2020) \cite{song2020score}, for DDPMs, the corresponding SDE is:
\begin{align}
    \textrm{d} \mathbf{x} = -\frac{1}{2}\mathbf{\beta}(t)\mathbf{x}dt + \sqrt{\beta(t)}\textrm{d}\mathbf{w} \label{ddpmsde}
\end{align}
where \(\mathbf{\beta}(\frac{t}{T}) = T\beta_{t}\) as $T$ goes to infinity; and for SGMs, the corresponding SDE is given by
\begin{align}
    \textrm{d} \mathbf{x} = \sqrt{\frac{\textrm{d} [\sigma(t)^2]}{\textrm{d} t}}\textrm{d}\mathbf{w} \label{eqn:sgmsde},
\end{align}
where $\sigma(\frac{t}{T}) = \sigma_t$ as $T$ goes to infinity. Here we use $q_t(\mathbf{x})$ to denote the distribution of $\mathbf{x}_t$ in the forward process.

Crucially, for any diffusion process in the form of \cref{9}, Anderson~\cite{anderson1982reverse} shows that it can be reversed by solving the following reverse-time SDE:
\begin{align}
    \textrm{d} \mathbf{x} = \left[\mathbf{f}(\mathbf{x},t) - g(t)^2\nabla_{\mathbf{x}} \log q_t(\mathbf{x})\right] \textrm{d}t + g(t) \textrm{d} \bar{\mathbf{w}}
    \label{eq:rsde}
\end{align}
where \(\bar{\mathbf{w}}\) is a standard Wiener process when time flows backwards, and \(\textrm{d}t\) denotes an infinitesimal negative time step. The solution trajectories of this reverse SDE share the same marginal densities as those of the forward SDE, except that they evolve in the opposite time direction \cite{song2020score}. Intuitively, solutions to the reverse-time SDE are diffusion processes that gradually convert noise to data. Moreover, Song et al. (2020) \cite{song2020score} prove the existence of an ordinary differential equation (ODE), namely the \emph{probability flow ODE}, whose trajectories have the same marginals as the reverse-time SDE. The probability flow ODE is given by:
\begin{align}
    \textrm{d} \mathbf{x} = \left[\mathbf{f}(\mathbf{x},t) - \frac{1}{2} g(t)^2\nabla_{\mathbf{x}} \log q_t(\mathbf{x})\right] \textrm{d}t.
    \label{eq:prob_flow_ode}
\end{align}
Both the reverse-time SDE and the probability flow ODE allow sampling from the same data distribution as their trajectories have the same marginals.

Once the score function at each time step t, $\nabla_{\mathbf{x}} \log q_t(\mathbf{x})$, is known, we unlock both the reverse-time SDE (\cref{eq:rsde}) and the probability flow ODE (\cref{eq:prob_flow_ode}) and can subsequently generate samples by solving them with various numerical techniques, such as annealed Langevin dynamics~\cite{song2019generative} (\cf, \cref{sec:back_sgm}), numerical SDE solvers~\cite{song2020score,jolicoeur2021gotta}, numerical ODE solvers~\cite{song2020score,song2020denoising,lu2022dpm,zhang2022fast,karras2022elucidating}, and predictor-corrector methods (combination of MCMC and numerical ODE/SDE solvers)~\cite{song2020score}. Like in SGMs, we parameterize a time-dependent score model \(\mathbf{s}_{\theta}(\mathbf{x}_t,t)\) to estimate the score function by generalizing the score matching objective in \cref{eq:dsm} to continuous time, leading to the following objective:
\begin{align}
    \Eb{t \sim \mathcal{U}[0, T], \mathbf x_0 \sim q(\mathbf x_0), \mathbf x_t \sim q(\mathbf x_t \mid \mathbf x_0)} { \lambda(t)  \left\| \mathbf{s}_{\theta}(\mathbf{x}_t,t) - \nabla_{\mathbf{x}_t} \log q_{0t}(\mathbf{x}_t \mid \mathbf{x}_0) \right\|^2}, \label{eq:11}
\end{align}
where $\mathcal{U}[0, T]$ denotes the uniform distribution over $[0, T]$, and the remaining notations follow \cref{eq:dsm}.

Subsequent research on diffusion models focuses on improving these classical approaches (DDPMs, SGMs, and Score SDEs) from three major directions: faster and more efficient sampling, more accurate likelihood and density estimation, and handling data with special structures (such as permutation invariance, manifold structures, and discrete data). We survey each direction extensively in the next three sections (\crefrange{sec3}{sec5}).
In \cref{tab2}, we list the three types of diffusion models with more detailed categorization, corresponding articles and years, under continuous and discrete time settings.

\begin{table}[htp]
\center

\caption{Three types of diffusion models are listed with corresponding articles and years, under continuous and discrete settings.}
\resizebox{0.95\textwidth}{!}{
\begin{tabular}{llllll}
\hline
\rowcolor{LightCyan}
Primary   & Secondary & Tertiary & Article & Year&Setting\\ \hline
\multirow{18}{*} {Efficient Sampling} &  \multirow{13}{*} {Learning-Free Sampling} &\multirow{7}{*}{SDE Solvers} &
Song et al. \cite{song2020score} &
 2020& Continuous\\
\multirow{19}{*} {} &\multirow{9}{*}{} & \multirow{4}{*}{} & \cellcolor{Gray} Dockhorn  et al. \cite{dockhorn2021score} & \cellcolor{Gray} 2021& \cellcolor{Gray} Continuous\\
\multirow{19}{*} {}  &\multirow{9}{*}{} & \multirow{4}{*}{} & Jolicoeur et al. \cite{JolicoeurMartineau2021AdversarialSM} & 2021 & Continuous\\

\multirow{19}{*} {}  &\multirow{9}{*}{} & \multirow{4}{*}{} & \cellcolor{Gray} Jolicoeur et al. \cite{jolicoeur2021gotta} & \cellcolor{Gray} 2021 & \cellcolor{Gray} Continuous\\

\multirow{19}{*} {}  &\multirow{9}{*}{} & \multirow{4}{*}{} &
Chuang et al. \cite{chung2022come} & 2022&Continuous\\

\multirow{19}{*} {}  &\multirow{9}{*}{} & \multirow{4}{*}{} &
\cellcolor{Gray} Song et al. \cite{song2019generative} & \cellcolor{Gray} 2019& \cellcolor{Gray} Continuous\\
\multirow{19}{*} {}  &\multirow{9}{*}{} & \multirow{4}{*}{} & Karras et al. \cite{karras2022elucidating} & 2022& Continuous\\
 \cline{3-6}
\multirow{19}{*} {} &\multirow{9}{*}{} & \multirow{6}{*}{ODE Solvers}&\cellcolor{Gray} Liu et al. \cite{liu2021pseudo} &\cellcolor{Gray} 2021& \cellcolor{Gray}Continuous\\
\multirow{11}{*} {} & \multirow{5}{*}{} &\multirow{2}{*}{} & Song et al. \cite{song2020denoising} & 2020& Continuous\\
\multirow{19}{*} {} &\multirow{9}{*}{} & \multirow{5}{*}{} &\cellcolor{Gray}
Zhang et al. \cite{zhang2022gddim} & \cellcolor{Gray}2022& \cellcolor{Gray} Continuous\\
\multirow{19}{*} {} &\multirow{9}{*}{} & \multirow{5}{*}{} & Karras et al. \cite{karras2022elucidating} & 2022&Continuous\\
\multirow{19}{*} {} &  \multirow{9}{*}{} & \multirow{5}{*}{} &\cellcolor{Gray}
 Lu et al. \cite{lu2022dpm} & \cellcolor{Gray}2022& \cellcolor{Gray}Continuous\\
\multirow{19}{*} {}  &\multirow{9}{*}{} &\multirow{5}{*}{} &
Zhang et al. \cite{zhang2022fast} & 2022& Continuous\\
 \cline{2-6}

\multirow{19}{*} {} &\multirow{7}{*}{Learning-Based Sampling}  &\multirow{3}{*}{Optimized Discretization} & \cellcolor{Gray}Watson et al. \cite{watson2021learning1} &  \cellcolor{Gray}2021& \cellcolor{Gray}Discrete\\ 

\multirow{19}{*} {} &\multirow{5}{*}{}  &\multirow{3}{*}{} & Watson et al. \cite{watson2021learning} &2021& Discrete\\

\multirow{19}{*} {} &\multirow{5}{*}{}  &\multirow{3}{*}{} & \cellcolor{Gray} Dockhorn et al. \cite{dockhorn2022genie} & \cellcolor{Gray} 2021& \cellcolor{Gray}Continuous\\ \cline{3-6}

\multirow{19}{*} {}  &\multirow{6}{*}{}&\multirow{2}{*}{Knowledge Distillation} &  Salimans et al. \cite{salimans2021progressive}  &   2021&  Discrete\\
\multirow{19}{*} {}  &\multirow{5}{*}{} &\multirow{2}{*}& \cellcolor{Gray}Luhman et al. \cite{luhman2021knowledge} & \cellcolor{Gray}2021 & \cellcolor{Gray}Discrete\\
\multirow{19}{*} {} &\multirow{5}{*}{}& \multirow{2}{*}{} &  Meng et al. \cite{meng2022distillation} & 2022& Discrete\\ \cline{3-6}
\multirow{19}{*} {}  &\multirow{5}{*}{}& \multirow{2}{*}{Truncated Diffusion} & \cellcolor{Gray} Lyu et al. \cite{lyu2022accelerating} & \cellcolor{Gray} 2022& \cellcolor{Gray} Discrete\\
\multirow{19}{*} {} &\multirow{5}{*}{}& \multirow{2}{*}{} &  Zheng et al. \cite{zheng2022truncated} &  2022& Discrete\\  \cline{1-6}

\multirow{9}{*} {Improved Likelihood}  &\multirow{4}{*}{Noise Schedule Optimization} &\multirow{4}{*}{Noise Schedule Optimization} & \cellcolor{Gray} Nichol et al. \cite{nichol2021improved} &  \cellcolor{Gray}2021& \cellcolor{Gray}Discrete\\ 
\multirow{9}{*} {}  & \multirow{2}{*}{} &\multirow{2}{*}{}& Kingma et al. \cite{kingma2021variational} &  2021&Discrete\\ 
\multirow{9}{*} {}  & \multirow{2}{*}{} &\multirow{2}{*}{}&\cellcolor{Gray} Huang et al. \cite{huang2024proteinligand} &  \cellcolor{Gray} 2024&\cellcolor{Gray} Discrete\\ 
\multirow{9}{*} {}  & \multirow{2}{*}{} &\multirow{2}{*}{}& Yang et al. \cite{yang2024crossmodal} &  2024&Discrete\\ \cline{2-6}
\multirow{9}{*} {}  & \multirow{2}{*}{Reverse Variance Learning}& \multirow{2}{*}{Reverse Variance Learning}& \cellcolor{Gray} Bao et al.\cite{bao2021analytic} & \cellcolor{Gray} 2021& \cellcolor{Gray}Discrete\\

\multirow{9}{*} {}  & \multirow{2}{*}{} &\multirow{2}{*}{}& Nichol et al. \cite{nichol2021improved} &  2021&Discrete\\ \cline{2-6}
\multirow{9}{*} {}  & \multirow{4}{*}{Exact Likelihood Computation} & \multirow{4}{*}{Exact Likelihood Computation}& \cellcolor{Gray} Song et al. \cite{song2021maximum} & \cellcolor{Gray} 2021& \cellcolor{Gray}Continuous\\ 

\multirow{9}{*} {}  & \multirow{2}{*}{} &\multirow{2}{*}{}& Huang et al. \cite{huang2021variational} &  2021&Continuous\\ 

\multirow{9}{*} {}  & \multirow{2}{*}{} & \multirow{2}{*}{}& \cellcolor{Gray} Song et al. \cite{song2020score} & \cellcolor{Gray} 2020& \cellcolor{Gray}Continuous\\ 
\multirow{9}{*} {}  & \multirow{2}{*}{} &\multirow{2}{*}{}&  Lu et al. \cite{lu2022maximum} &  2022 &Continuous\\ \cline{1-6}

\multirow{15}{*} {{Data with Special Structures}}  & \multirow{6}{*}{Manifold Structures}&\multirow{4}{*}{Learned Manifolds}& \cellcolor{Gray} Vahdat et al. \cite{vahdat2021score} & \cellcolor{Gray}2021& \cellcolor{Gray}Continuous\\
\multirow{11}{*} {} & \multirow{5}{*}{} & \multirow{2}{*}{} &  Yang et al. \cite{yang2024structure} &  2024& Discrete\\
\multirow{11}{*} {} & \multirow{5}{*}{} & \multirow{2}{*}{} &  \cellcolor{Gray} Ramesh et al. \cite{ramesh2022hierarchical} & \cellcolor{Gray} 2022& \cellcolor{Gray}Discrete\\

\multirow{11}{*} {} & \multirow{5}{*}{} & \multirow{2}{*}{} &  Rombach et al. \cite{rombach2022high} &  2022& Discrete\\\cline{3-6}

\multirow{11}{*} {} & \multirow{5}{*}{}& \multirow{2}{*}{Known Manifolds} &\cellcolor{Gray} Bortoli et al. \cite{de2022riemannian} & \cellcolor{Gray}2022&\cellcolor{Gray} Continuous\\

\multirow{11}{*} {} & \multirow{5}{*}{}& \multirow{2}{*}{} & Huang et al. \cite{Huang2022RiemannianDM} &2022& Continuous\\\cline{2-6}

\multirow{15}{*}{}  & \multirow{5}{*}{Data with Invariant Structures}& \multirow{4}{*}{Data with Invariant Structures}& \cellcolor{Gray} Niu et al. \cite{niu2020permutation} & \cellcolor{Gray}2020& \cellcolor{Gray}Discrete\\ 
\multirow{11}{*} {}  & \multirow{5}{*}{} &\multirow{2}{*}{} &Jo et al. \cite{jo2022score}  & 2022&Continuous\\
\multirow{11}{*} {}  & \multirow{5}{*}{} &\multirow{2}{*}{} &\cellcolor{Gray} Shi et al. \cite{shi2021learning}  & \cellcolor{Gray} 2022&\cellcolor{Gray}Continuous\\
\multirow{11}{*} {}  & \multirow{5}{*}{} &\multirow{2}{*}{} &Xu et al. \cite{xu2021geodiff}  & 2021&Discrete\\ \cline{2-6}
\multirow{11}{*} {}   & \multirow{7}{*}{Discrete Data} &\multirow{7}{*}{Discrete Data}
&  \cellcolor{Gray}Meng et al. \cite{meng2022concrete} & \cellcolor{Gray}2022& \cellcolor{Gray} Discrete\\
\multirow{11}{*} {}   & \multirow{4}{*}{} &\multirow{4}{*}{} & liu et al. \cite{liu2023learning} & 2023&Continuous\\ 
\multirow{11}{*} {}   & \multirow{4}{*}{} &\multirow{4}{*}{}&  \cellcolor{Gray}Sohl et al. \cite{sohl2015deep} & \cellcolor{Gray}2015& \cellcolor{Gray} Discrete\\
\multirow{11}{*} {}   & \multirow{4}{*}{} &\multirow{4}{*}{} & Austin et al. \cite{austin2021structured} & 2021&Discrete\\ 
\multirow{11}{*} {}   & \multirow{5}{*}{} &\multirow{3}{*}{} & \cellcolor{Gray}Yang et al. \cite{yang2025mmada} &\cellcolor{Gray} 2025& \cellcolor{Gray}Discrete\\
\multirow{11}{*} {}   & \multirow{5}{*}{} &\multirow{3}{*}{} & Wang et al. \cite{wang2025revolutionizing} &2025&Discrete\\ 
\multirow{11}{*} {}   & \multirow{5}{*}{} &\multirow{3}{*}{}& \cellcolor{Gray} Campbell et al. \cite{campbell2022continuous} & \cellcolor{Gray}2022&\cellcolor{Gray} Continuous\\
\bottomrule
\end{tabular}}
\label{tab2}
\end{table}

\section{Diffusion Models with Efficient Sampling}
\label{sec3}

Generating samples from diffusion models typically demands iterative approaches that involve a large number of evaluation steps. A great deal of recent work has focused on speeding up the sampling process while also improving quality of the resulting samples. We classify these efficient sampling methods into two main categories: those that do not involve learning (learning-free sampling) and those that require an additional learning process after the diffusion model has been trained (learning-based sampling).

\subsection{Learning-Free Sampling}
Many samplers for diffusion models rely on discretizing either the reverse-time SDE present in \cref{eq:rsde} or the probability flow ODE from \cref{eq:prob_flow_ode}. Since the cost of sampling increases proportionally with the number of discretized time steps, many researchers have focused on developing discretization schemes that reduce the number of time steps while also minimizing discretization errors.

\subsubsection{SDE Solvers}

The generation process of DDPM \cite{sohl2015deep,ho2020denoising} can be viewed as a particular discretization of the reverse-time SDE. As discussed in \cref{sec:back_sde}, the forward process of DDPM discretizes the SDE in \cref{ddpmsde}, whose corresponding reverse SDE takes the form of
\begin{align}
    \textrm{d} \mathbf{x} = -\frac{1}{2}\mathbf{\beta}(t)(\mathbf{x}_t-\nabla_{\mathbf x_t} \log q_t(\mathbf{x}_t)) \textrm{d}t + \sqrt{\beta(t)}\textrm{d}\mathbf{w} \label{rddpm}
\end{align}
Song et al. (2020) \cite{song2020score} show that the reverse Markov chain defined by \cref{4} amounts to a numerical SDE solver for \cref{rddpm}.

Noise-Conditional Score Networks (NCSNs) \cite{song2019generative} and Critically-Damped Langevin Diffusion (CLD) \cite{dockhorn2021score} both solve the reverse-time SDE with inspirations from Langevin dynamics. In particular, NCSNs leverage annealed Langevin dynamics (ALD, \cf, \cref{sec:back_sgm}) to iteratively generate data while smoothly reducing noise level until the generated data distribution converges to the original data distribution. Although the sampling trajectories of ALD are not exact solutions to the reverse-time SDE, they have the correct marginals and hence produce correct samples under the assumption that Langevin dynamics converges to its equilibrium at every noise level. The method of ALD is further improved by Consistent Annealed Sampling (CAS) \cite{JolicoeurMartineau2021AdversarialSM}, a score-based MCMC approach with better scaling of time steps and added noise. Inspired by statistical mechanics, CLD proposes an augmented SDE with an auxiliary velocity term resembling underdamped Langevin diffusion. To obtain the time reversal of the extended SDE, CLD only needs to learn the score function of the conditional distribution of velocity given data, arguably easier than learning scores of data directly. The added velocity term is reported to improve sampling speed as well as quality.

The reverse diffusion method proposed in \cite{song2020score} discretizes the reverse-time SDE in the same way as the forward one. For any one-step discretization of the forward SDE, one may write the general form below:
\begin{align}
     \mathbf{x}_{i+1} = \mathbf{x}_i + \mathbf{f}_i(\mathbf{x}_i) + \mathbf{g}_i \mathbf{z}_i, \quad i = 0, 1, \cdots, N-1 \label{eq:13}
 \end{align}
where $\mathbf{z}_i \sim \mathcal{N}(\mathbf{0}, \mathbf{I})$, $\mathbf{f}_i$ and $\mathbf{g}_i$ are determined by drift/diffusion coefficients of the SDE and the discretization scheme. Reverse diffusion proposes to discretize the reverse-time SDE similarly to the forward SDE, \ie,
 \begin{align}
     \mathbf{x}_{i} = \mathbf{x}_{i+1} -\mathbf{f}_{i+1}(\mathbf{x}_{i+1}) + \mathbf{g}_{i+1}\mathbf{g}_{i+1}^{t} \mathbf{s}_{\mathbf{\theta^*}}(\mathbf{x}_{i+1}, t_{i+1}) + \mathbf{g}_{i+1}\mathbf{z}_i \quad i = 0, 1, \cdots, N-1
 \end{align}
where $\mathbf{s}_{\mathbf{\theta^*}}(\mathbf{x}_i, t_{i})$ is the trained noise-conditional score model. Song et al. (2020) \cite{song2020score} prove that the reverse diffusion method is a numerical SDE solver for the reverse-time SDE in \cref{eq:rsde}. This process can be applied to any types of forward SDEs, and empirical results indicate this sampler performs slightly better than DDPM \cite{song2020score} for a particular type of SDEs called the VP-SDE.

Jolicoeur-Martineau et al. (2021) \cite{jolicoeur2021gotta} develop an SDE solver with adaptive step sizes for faster generation. The step size is controlled by comparing the output of a high-order SDE solver versus the output of an low-order SDE solver. At each time step, the high- and low-order solvers generate new sample $\mathbf{x}'_{\text{high}}$ and $\mathbf{x}'_{\text{low}}$  from the previous sample $\mathbf{x}'_{prev}$ respectively. The step size is then adjusted by comparing the difference between the two samples. If $\mathbf{x}'_{\text{high}}$ and $\mathbf{x}'_{\text{low}}$ are similar, the algorithm will return $\mathbf{x}'_{\text{high}}$ and then increase the step size. 
The similarity between $\mathbf{x}'_{\text{high}}$ and $\mathbf{x}'_{\text{low}}$ is measured by:
 \begin{equation}
E_{q} = \left\|\frac{\mathbf{x}'_{\text{low}} - \mathbf{x}'_{\text{high}}}{\mathbf{\delta}(\mathbf{x}', \mathbf{x}'_{\text{prev}})}\right\|^2
\end{equation}
where $\mathbf{\delta}(\mathbf{x}'_{\text{low}}, \mathbf{x}'_{\text{prev}}) \coloneqq \max(\epsilon_{abs}, \epsilon_{rel}\max(\mid\mathbf{x}'_{\text{low}}, \mid\mathbf{x}'_{\text{prev}}|))$, and
$\epsilon_{abs}$ and $\epsilon_{rel}$ are absolute and relative tolerances. 

The predictor-corrector method proposed in \cite{song2020score} solves the reverse SDE by combining numerical SDE solvers (``predictor'') and iterative Markov chain Monte Carlo (MCMC) approaches (``corrector"). At each time step, the predictor-corrector method first employs a numerical SDE solver to produce a coarse sample, followed by a "corrector" that corrects the sample' marginal distribution with score-based MCMC. The resulting samples have the same time-marginals as solution trajectories of the reverse-time SDE, \ie, they are equivalent in distribution at all time steps. Empirical results demonstrate that adding a corrector based on Langevin Monte Carlo is more efficient than using an additional predictor without correctors \cite{song2020score}. Karras et al. (2022) \cite{karras2022elucidating} further improve the Langevin dynamics corrector in \cite{song2020score} by proposing a Langevin-like ``churn'' step of adding and removing noise, achieving new state-of-the-art sample quality on datasets like CIFAR-10 \cite{krizhevsky2009learning} and ImageNet-64 \cite{deng2009imagenet}.

\subsubsection{ODE solvers}
A large body of works on faster diffusion samplers are based on solving the probability flow ODE (\cref{eq:prob_flow_ode}) introduced in \cref{sec:back_sde}. In contrast to SDE solvers, the trajectories of ODE solvers are deterministic and thus not affected by stochastic fluctuations. These deterministic ODE solvers typically converge much faster than their stochastic counterparts at the cost of slightly inferior sample quality. 

Denoising Diffusion Implicit Models (DDIM) \cite{song2020denoising} is one of the earliest work on accelerating diffusion model sampling. The original motivation was to extend the original DDPM to non-Markovian case with the following Markov chain
\begin{align}
    & q(\mathbf{x}_1, \ldots, \mathbf{x}_T\mid\mathbf{x}_0) = \prod_{t=1}^{T} q(\mathbf{x}_t\mid\mathbf{x}_{t-1},\mathbf{x}_0)\\
    & q_\sigma(\mathbf{x}_{t-1}\mid\mathbf{x}_t, \mathbf{x}_0) = \mathcal{N}(\mathbf{x}_{t-1}|\tilde{\mu}_t(\mathbf{x}_t, \mathbf{x}_0), \sigma_t^2 \mathbf{I}) \\
    & \tilde{\mu}_t(\mathbf{x}_t, \mathbf{x}_0) \coloneqq \sqrt{\overline{\alpha}_{t-1}} \mathbf{x}_0 + \sqrt{1 - \overline{\alpha}_{t-1} - \sigma_t^2} \cdot \frac{\mathbf{x}_t - \sqrt{\overline{\alpha}_t}\mathbf{x}_0}{\sqrt{1-\overline{\alpha}_t}}
\end{align}
This formulation encapsulates DDPM and DDIM as special cases, where DDPM corresponds to setting $\sigma_t^2 = \frac{\hat{\beta}_{t-1}}{\hat{\beta_t}}\beta_t$ and DDIM corresponds to setting $\sigma_t^2 = 0$. DDIM learns a Markov chain to reverse this non-Markov perturbation process, which is fully deterministic when $\sigma_t^2 = 0$. It is observed in \cite{song2020denoising,salimans2021progressive,lu2022dpm,karras2022elucidating} that the DDIM sampling process amounts to a special discretization scheme of the probability flow ODE. Inspired by an analysis of DDIM on a singleton dataset, generalized Denoising Diffusion Implicit Models (gDDIM) \cite{zhang2022gddim} proposes a modified parameterization of the score network that enables deterministic sampling for more general diffusion processes, such as the one in Critically-Damped Langevin Diffusion (CLD) \cite{dockhorn2021score}. PNDM \cite{liu2021pseudo} proposes a pseudo numerical method to generate sample along a specific manifold in $\mathcal{R}^N$. It uses numerical solver with nonlinear transfer part to solve differential equation on manifolds and then generates sample, which encapsulates DDIM as a special case.

Through extensive experimental investigations, Karras et al. (2022) \cite{karras2022elucidating} show that Heun's $2^{nd}$ order method \cite{ascher1998computer} provides an excellent trade off between sample quality and sampling speed. The higher-order solver leads to smaller discretization error at the cost of one additional evaluation of the learned score function per time step. Heun's method generates samples of comparable, if not better quality than Euler's method with fewer sampling steps.

Diffusion Exponential Integrator Sampler \cite{zhang2022fast} and DPM-solver \cite{lu2022dpm} leverage the semi-linear structure of probability flow ODE to develop customized ODE solvers that are more efficient than general-purpose Runge-Kutta methods. Specifically, the linear part of probability flow ODE can be analytically computed, while the non-linear part can be solved with techniques similar to exponential integrators in the field of ODE solvers. These methods contain DDIM as a first-order approximation. However, they also allow for higher order integrators, which can produce high-quality samples in just 10 to 20 iterations---far fewer than the hundreds of iterations typically required by diffusion models without accelerated sampling.

\subsection{Learning-Based Sampling}
Learning-based sampling is another efficient approach for diffusion models. By using partial steps or training a sampler for the reverse process, this method achieves faster sampling speeds at the expense of slight degradation in sample quality. Unlike learning-free approaches that use handcrafted steps, learning-based sampling typically involves selecting steps by optimizing certain learning objectives.

\subsubsection{Optimized Discretization}

Given a pre-trained diffusion model, Watson et al. (2021) \cite{watson2021learning1} put forth a strategy for finding the optimal discretization scheme by selecting the best $K$ time steps to maximize the training objective for DDPMs. Key to this approach is the observation that the DDPM objective can be broken down into a sum of individual terms, making it well suited for dynamic programming. However, it is well known that the variational lower bound used for DDPM training does not correlate directly with sample quality \cite{theis2015note}. A subsequent work, called Differentiable Diffusion Sampler Search \cite{watson2021learning}, addresses this issue by directly optimizing a common metric for sample quality called the Kernel Inception Distance (KID) \cite{binkowski2018demystifying}. This optimization is feasible with the help of reparameterization \cite{kingma2013auto,rezende2014stochastic} and gradient rematerialization. Based on truncated Taylor methods, Dockhorn et al. (2022) \cite{dockhorn2022genie} derive a second-order solver for accelerating synthesis by training a additional head on top of the first-order
score network.

\subsubsection{Truncated Diffusion}
One can improve sampling speed by truncating the forward and reverse diffusion processes \cite{lyu2022accelerating, zheng2022truncated}. The key idea is to halt the forward diffusion process early on, after just a few steps, and to begin the reverse denoising process with a non-Gaussian distribution. Samples from this distribution can be obtained efficiently by diffusing samples from pre-trained generative models, such as variational autoencoders \cite{kingma2013auto,rezende2014stochastic} or generative adversarial networks \cite{goodfellow2014generative}.

\subsubsection{Knowledge Distillation}

Approaches that use knowledge distillation \cite{luhman2021knowledge,salimans2021progressive,meng2022distillation} can significantly improve the sampling speed of diffusion models. Specifically, in Progressive Distillation \cite{salimans2021progressive}, the authors propose distilling the full sampling process into a faster sampler that requires only half as many steps. By parameterizing the new sampler as a deep neural network, authors are able to train the sampler to match the input and output of the DDIM sampling process. Repeating this procedure can further reduce sampling steps, although fewer steps can result in reduced sample quality. To address this issue, the authors suggest new parameterizations for diffusion models and new weighting schemes for the objective function. 

\section{Diffusion Models with Improved Likelihood}
\label{sec4}

As discussed in \cref{sec:back_ddpm}, the training objective for diffusion models is a (negative) variational lower bound (VLB) on the log-likelihood. This bound, however, may not be tight in many cases \cite{kingma2021variational}, leading to potentially suboptimal log-likelihoods from diffusion models. In this section, we survey recent works on likelihood maximization for diffusion models. We focus on three types of methods: noise schedule optimization, reverse variance learning, and exact log-likelihood evaluation.

\subsection{Noise Schedule Optimization}
In the classical formulation of diffusion models, noise schedules in the forward process are handcrafted without trainable parameters. By optimizing the forward noise schedule jointly with other parameters of diffusion models, one can further maximize the VLB in order to achieve higher log-likelihood values \cite{nichol2021improved,kingma2021variational}.

The work of iDDPM \cite{nichol2021improved} demonstrates that a certain cosine noise schedule can improve log-likelihoods. Specifically, the cosine noise schedule in their work takes the form of
\begin{equation}
\bar{\alpha}_t = \frac{h(t)}{h(0)},\quad h(t) = \cos\left(\frac{t/T + m}{1+m}\cdot\frac{\pi}{2}\right)^2
\end{equation}
where $\bar{\alpha}_t$ and $\beta_t$ are defined in \cref{eq:2,eq:3}, and $m$ is a hyperparameter to control the noise scale at $t=0$. They also propose a parameterization of the reverse variance with an interpolation between $\beta_t$ and $1-\bar{\alpha}_t$ in the log domain.

In Variational Diffusion Models (VDMs) \cite{kingma2021variational}, authors propose to improve the likelihood of continuous-time diffusion models by jointly training the noise schedule and other diffusion model parameters to maximize the VLB. They parameterize the noise schedule using a monotonic neural network $\gamma_\eta(t)$, and build the forward perturbation process according to $\sigma_t^2 = \operatorname{sigmoid}(\gamma_\eta(t))$, $q(\mathbf{x}_t\mid\mathbf{x}_0) = \mathcal{N}(\bar{\alpha}_t\mathbf{x}_0,\sigma_t^2\mathbf{I})$, and $\bar{\alpha}_t = \sqrt{(1-\sigma_t^2)}$. Moreover, authors prove that the VLB for data point $\mathbf{x}$ can be simplified to a form that only depends on the signal-to-noise ratio \(\text{R}(t)\coloneqq \frac{\bar{\alpha}_t^2}{\sigma^2_t}\). In particular, the $L_{VLB}$ can be decomposed to
\begin{equation}
    L_{VLB} = -\E_{\mathbf{x}_0} \operatorname{KL}(q(\mathbf{x}_T|\mathbf{x}_0)\mid\mid p(\mathbf{x}_T)) + \E_{\mathbf{x}_0,\mathbf{x}_1}\log p(\mathbf{x}_0|\mathbf{x}_1) - L_{D},
\end{equation}
where the first and second terms can be optimized directly in analogy to training variational autoencoders. The third term can be further simplified to the following:
\begin{align}
    L_{D} = \frac{1}{2}\E_{\mathbf{x}_0,\mathbf{\epsilon}\sim\mathcal{N}(0,\mathbf{I})} \int_{\text{R}_{\text{min}}}^{\text{R}_{\text{max}}}\left\rVert \mathbf{x}_0 - \tilde{\mathbf{x} }_{\theta}(\mathbf{x}_v, v) \right\lVert_{2}^{2} dv,
    \label{30}
\end{align}
where $\text{R}_{\text{max}} = R(1), \text{R}_{\text{min}} = R(T)$, \(\mathbf{x}_v = \bar{\alpha}_v \mathbf{x}_0 + \sigma_v \epsilon\) denotes a noisy data point obtained by diffusing $\mathbf x_0$ with the forward perturbation process until $t = R^{-1}(v)$, and $\tilde{\mathbf x}_\theta$ denotes the predicted noise-free data point by the diffusion model. As a result, noise schedules do not affect the VLB as long as they share the same values at $\text{R}_{\text{min}}$ and $\text{R}_{\text{max}}$, and will only affect the variance of Monte Carlo estimators for VLB.

Another line of works \citep{huang2023protein,yang2024crossmodal} propose to the modify diffusion trajectory through the integration of cross-modality information. Specifically, the cross-modal information, denoted as $r_\phi(y,x_0)$, is extracted from any conditional input $y$ and original sample $x_0$ with relational network $r_\phi(\cdot)$. And then it can be injected to the forward process as an additional bias to adapt diffusion trajectory: 
\begin{equation}
    q_t(x_t|x_0,y) = \mathcal{N}(x_t,\sqrt{\bar{\alpha}_t}x_0+k_t r_\phi(x_0,y),(1-\bar{\alpha}_t) I)
\end{equation}
where $k_t$ is a non-negative scalar that control the magnitude of the bias term. It is important to note that with this modification, the forward process ceases to be a Markovian chain. ContextDiff \citep{yang2024crossmodal} introduces a general framework to jointly learn the cross-modal relational network $r_\phi$ and the diffusion model, and derives the VLB and sampling procedure for this modified diffusion process. 
\subsection{Reverse Variance Learning}
The classical formulation of diffusion models assumes that Gaussian transition kernels in the reverse Markov chain have fixed variance parameters. Recall that we formulated the reverse kernel as $q_\theta(\mathbf{x}_{t-1}\mid\mathbf{x}_t) = \mathcal{N}(\mu_\theta(\mathbf{x}_t, t), \Sigma_\theta(\mathbf x_t, t))$ in \cref{4} but often fixed the reverse variance $\Sigma_\theta(\mathbf x_t, t)$ to $\beta_t \mathbf{I}$. Many methods propose to train the reverse variances as well to further maximize VLB and log-likelihood values.

In iDDPM \cite{nichol2021improved}, Nichol and Dhariwal propose to learn the reverse variances by parameterizing them with a form of linear interpolation and training them using a hybrid objective. This results in higher log-likelihoods and faster sampling without losing sample quality. In particular, they parameterize the reverse variance in \cref{4} as:
\begin{align}
\Sigma_{\theta}(\mathbf{x}_t,t) = \exp(\theta \cdot\log\beta_t + (1-\theta )\cdot\log \tilde{\beta}_t),
\end{align}
where $\tilde{\beta}_t \coloneqq \frac{1-\bar{\alpha}_{t-1}}{1-\bar{\alpha}_t}\cdot \beta_t$ and $\theta$ is jointly trained to maximize VLB.
This simple parameterization avoids the instability of estimating more complicated forms of $\Sigma_{\theta}(\mathbf{x}_t,t)$ and is reported to improve likelihood values.

Analytic-DPM \cite{bao2021analytic} shows a remarkable result that the optimal reverse variance can be obtained from a pre-trained score function, with the analytic form below:
\begin{equation}
    \Sigma_\theta(\mathbf{x}_t,t) =
    \sigma_t^2 + \left( \sqrt{\frac{\overline{\beta}_t }{\alpha_t}} - \sqrt{\overline{\beta}_{t-1} - \sigma_t^2} \right)^2 \cdot
    \left(1 - \overline{\beta}_t \E_{q_t(\mathbf{x}_t)} \frac{||\nabla_{\mathbf{x}_t} \log q_t(\mathbf{x}_t)||^2}{d} \right) \label{eq:opt_sigma}
\end{equation}
As a result, given a pre-traied score model, we can estimate its first- and second-order moments to obtain the optimal reverse variances. Plugging them into the VLB can lead to tighter VLBs and higher likelihood values.

\subsection{Exact Likelihood Computation}
In the Score SDE \cite{song2020score} formulation, samples are generated by solving the following reverse SDE, where $\nabla_{\mathbf x_t} \log p_\theta (\mathbf x_t, t)$ in \cref{eq:rsde} is replaced by the learned noise-conditional score model \(\mathbf{s}_{\theta}(\mathbf{x}_t,t)\):
\begin{align}
    \textrm{d} \mathbf{x} &= f(\mathbf{x}_t,t) - g(t)^2\mathbf{s}_{\theta}(\mathbf{x}_t,t) \textrm{d}t + g(t) \textrm{d}\mathbf{w}.
\end{align}
Here we use $p_\theta^{\text{sde}}$ to denote the distribution of samples generated by solving the above SDE. One can also generate data by plugging the score model into the probability flow ODE in \cref{eq:prob_flow_ode}, which gives:
\begin{align}
    \frac{\textrm{d} \mathbf{x}_t}{\textrm{d}t} &= \underbrace{f(\mathbf{x}_t,t) - \frac{1}{2}g^2(t)\mathbf{s}_{\theta}(\mathbf{x}_t,t)}_{\coloneqq \tilde{f}_\theta(\mathbf{x}_t,t)} \label{eq:40}
\end{align}
Similarly, we use $p_\theta^{\text{ode}}$ to denote the distribution of samples generated via solving this ODE. The theory of neural ODEs \cite{CheTubBet18} and continuous normalizing flows \cite{grathwohl2018scalable} indicates that $p_\theta^{\text{ode}}$ can be computed accurately albeit with high computational cost. For $p_\theta^{\text{sde}}$, several concurrent works \cite{song2021maximum,lu2022maximum,huang2021variational} demonstrate that there exists an efficiently computable variational lower bound, and we can directly train our diffusion models to maximize $p_\theta^{\text{sde}}$ using modified diffusion losses.

Specifically, Song et al. (2021) \cite{song2021maximum} prove that with a special weighting function (likelihood weighting), the objective used for training score SDEs implicitly maximizes the expected value of $p_\theta^{\text{sde}}$ on data. It is shown that
\begin{align}
    \mathbf{D}_{KL}(q_0\mathrel\| p_\theta^{\text{sde}}) \leq \mathcal{L}(\theta; g(\cdot)^2) + \mathbf{D}_{KL}(q_T\mathrel\| \pi)\label{eqn:39},
\end{align}
where $\mathcal{L}(\theta; g(\cdot)^2)$ is the Score SDE objective in \cref{eq:11} with $\lambda(t) = g(t)^2$. Since $\mathbf{D}_{KL}(q_0\mathrel\| p^{\text{sde}}_\theta) = -\E_{q_0}\log(p^{\text{sde}}_\theta) + \text{const}$, and $\mathbf{D}_{KL}(q_T\mathrel\| \pi)$ is a constant, training with $\mathcal{L}(\theta; g(\cdot)^2)$ amounts to minimizing $-\E_{q_0}\log(p^{\text{sde}}_\theta)$, the expected negative log-likelihood on data. Moreover, Song et al. (2021) and Huang et al. (2021) \cite{huang2021variational,song2021maximum} provide the following bound for $p^{\text{sde}}_\theta(\mathbf x)$:
\begin{align}
-\log  p^{\text{sde}}_\theta(\mathbf x) \leq \mathcal{L}^{'}(\mathbf x),
\end{align}
where $\mathcal{L}^{'}(\mathbf x)$ is defined by
\begin{align}
    \mathcal{L}^{'}(\mathbf x)\coloneqq\int_0^T \E\left[ \frac{1}{2} ||g(t)\mathbf{s}_{\theta}(\mathbf{x}_t,t)||^2 + \nabla \cdot (g(t)^2\mathbf{s}_{\theta}(\mathbf{x}_t,t) \sc - f(\mathbf{x}_t),t) \,\bigg|\, \mathbf{x}_0=x \right] dt
     -\E_{\mathbf{x}_T}[\log p^{\text{sde}}_\theta(\mathbf{x}_T)\,|\,\mathbf{x}_0=x]
    \label{eq:41}
\end{align}
The first part of \cref{eq:41} is reminiscent of implicit score matching \cite{Hyvrinen2005EstimationON} and the whole bound can be efficiently estimated with Monte Carlo methods.

Since the probability flow ODE is a special case of neural ODEs or continuous normalizing flows, we can use well-established approaches in those fields to compute $\log p_\theta^{\text{ode}}$ accurately. Specifically, we have
\begin{align}
    \log p_\theta^{\text{ode}}(\mathbf x_0) = \log p_T(\mathbf{x}_T) + \int_{t=0}^T \nabla \cdot \tilde{f}_\theta(\mathbf{x}_t,t) \textrm{d}t.
\end{align}
One can compute the one-dimensional integral above with numerical ODE solvers and the Skilling-Hutchinson trace estimator \cite{skilling1989eigenvalues,hutchinson1989stochastic}. Unfortunately, this formula cannot be directly optimized to maximize $p_\theta^{\text{ode}}$ on data, as it requires calling expensive ODE solvers for each data point $\mathbf x_0$. To reduce the cost of directly maximizing $p_\theta^{\text{ode}}$ with the above formula, Song et al. (2021) \cite{song2021maximum} propose to maximize the variational lower bound of $p_\theta^{\text{sde}}$ as a proxy for maximizing $p_\theta^{\text{ode}}$, giving rise to a family of diffusion models called \emph{ScoreFlows}.

Lu et al. (2022) \cite{lu2022maximum} further improve ScoreFlows by proposing to minimize not just the vanilla score matching loss function, but also its higher order generalizations.
They prove that $\log p_\theta^{\text{ode}}$ can be bounded with the first, second, and third-order score matching errors. Building upon this theoretical result, authors further propose efficient training algorithms for minimizing high order score matching losses and reported improved $p_\theta^{\text{ode}}$ on data.

\section{Diffusion Models for Data with Special Structures}
\label{sec5}

While diffusion models have achieved great success for data domains like images and audio, they do not necessarily translate seamlessly to other modalities. Many important data domains have special structures that must be taken into account for diffusion models to function effectively. Difficulties may arise, for example, when models rely on score functions that are only defined on continuous data domains, or when data reside on low dimensional manifolds. To cope with these challenges, diffusion models have to be adapted in various ways.

\subsection{Discrete Data}
Most diffusion models are geared towards continuous data domains, because Gaussian noise perturbation as used in DDPMs is not a natural fit for discrete data, and the score functions required by SGMs and Score SDEs are only defined on continuous data domains. To overcome this difficulty, several works \cite{hoogeboom2021argmax,gu2022vector,xie2022vector, austin2021structured} build on Sohl-Dickstein et al. (2015) \cite{sohl2015deep} to generate discrete data of high dimensions. Specifically, VQ-Diffusion \cite{gu2022vector} replaces Gaussian noise with a random walk on the discrete data space, or a random masking operation. The resulting transition kernel for the forward process takes the form of
\begin{equation}
q(\mathbf{x}_t\mid\mathbf{x}_{t-1}) = \mathbf{v}^\top(\mathbf{x}_{t})\mathbf{Q}_t \mathbf{v}(\mathbf{x}_{t-1})
\end{equation}
where $\mathbf{v}(\mathbf{x})$ is a one-hot column vector, and $\mathbf{Q}_t$ is the transition kernel of a lazy random walk.

Building on discrete diffusion principles, MMaDA \cite{yang2025mmada} introduces a novel approach for multi-modal discrete data generation by incorporating masked autoencoder architectures within the diffusion framework. This method demonstrates improved performance on complex discrete sequences by leveraging cross-modal information. Similarly, TraceRL \cite{wang2025revolutionizing} extends discrete diffusion models to more comprehensive reinforcement learning settings, where the discrete state-action sequences are generated through a specialized diffusion process that incorporates reward signals and policy constraints.

D3PM \cite{austin2021structured} accommodates discrete data in diffusion models by constructing the forward noising process with absorbing state kernels or discretized Gaussian kernels. Campbell et al. (2022) \cite{campbell2022continuous} present the first continuous-time framework for discrete diffusion models. Leveraging Continuous Time Markov Chains, they are able to derive efficient samplers that outperform discrete counterparts, while providing a theoretical analysis on the error between the sample distribution and the true data distribution.

Concrete Score Matching (CSM) \cite{meng2022concrete} proposes a generalization of the score function for discrete random variables. Concrete score is defined by the rate of change of the probabilities with respect to directional changes of the input, which can be seen as a finite-difference approximation to the continuous (Stein) score. The concrete score can be efficiently trained and applied to MCMC.

Based on the theory of stochastic calculus, Liu et al. (2023) \cite{liu2023learning} proposes a framework for diffusion models to generate data on constrained and structured domains, including discrete data as a special case. Using a fundamental theorem in stochastic calculus, the Doob's h-transform, one can constrain the data distribution on a specific area by including a special force term in the reverse diffusion process. They use a parameterization of the force term with an EM-based optimization algorithm. Furthermore, the loss function can be transformed to $L_2$ loss using Girsanov theorem.

Inspired by the success of reinforcement learning in discrete language modeling \cite{guo2025deepseek}, recent works like MMaDA \cite{yang2025mmada} and TraceRL \cite{wang2025revolutionizing} learn to model discrete textual and visual data with masked diffusion model  revolutionize

\subsection{Data with Invariant Structures}
Data in many important domains have invariant structures. For example, graphs are permutation invariant, and point clouds are both translation and rotation invariant. In diffusion models, these invariances are often ignored, which can lead to suboptimal performance. To address this issue, several works \cite{niu2020permutation,de2022riemannian} propose to endow diffusion models with the ability to account for invariance in data.

Niu et al. (2020) \cite{niu2020permutation} first tackle the problem of permutation invariant graph generation with diffusion models. They achieve this by using a permutation equivariant graph neural network \cite{gori2005new,scarselli2008graph,wu2020comprehensive}, called EDP-GNN, to parameterize the noise-conditional score model. GDSS \cite{jo2022score} further develops this idea by proposing a continuous-time graph diffusion process. This process models both the joint distribution of nodes and edges through a system of stochastic differential equations (SDEs), where message-passing operations are used to guarantee permutation invariance.

Similarly, Shi et al. (2021) \cite{shi2021learning} and Xu et al. (2022) \cite{xu2021geodiff} enable diffusion models to generate molecular conformations that are invariant to both translation and rotation.
For example, Xu et al. (2022) \cite{xu2021geodiff} shows that Markov chains starting with an invariant prior and evolving with equivariant Markov kernels can induce an invariant marginal distribution, which can be used to enforce appropriate data invariance in molecular conformation generation. Formally, let $\mathcal{T}$ be a rotation or translation operation. Given that $
     p(\mathbf{x}_T) = p(\mathcal{T}(\mathbf{x}_T)),\ p_\theta(\mathbf{x}_{t-1}\mid\mathbf{x}_t) = p_\theta(\mathcal{T}(\mathbf{x}_{t-1})\mid \mathcal{T}(\mathbf{x}_t))$,
Xu et al. (2022) \cite{xu2021geodiff} prove that the distribution of samples is guaranteed to be invariant to $\mathcal{T}$, that is, $p_0(\mathbf{x}) = p_0(\mathcal{T}(\mathbf{x}))$. As a result, one can build a diffusion model that generates rotation and translation invariant molecular conformations as long as the prior and transition kernels enjoy the same invariance.

\subsection{Data with Manifold Structures}
Data with manifold structures are ubiquitous in machine learning. As the manifold hypothesis \cite{fefferman2016testing} posits, natural data often reside on manifolds with lower intrinsic dimensionality. In addition, many data domains have well-known manifold structures. For instance, climate and earth data naturally lie on the sphere because that is the shape of our planet. Many works have focused on developing diffusion models for data on manifolds. We categorize them based on whether the manifolds are known or learned, and introduce some representative works below.
\subsubsection{Known Manifolds}

Recent studies have extended the Score SDE formulation to various known manifolds. This adaptation parallels the generalization of neural ODEs \cite{CheTubBet18} and continuous normalizing flows \cite{grathwohl2018scalable} to Riemannian manifolds \cite{lou2020neural,mathieu2020riemannian}. To train these models, researchers have also adapted score matching and score functions to Riemannian manifolds.

The Riemannian Score-Based Generative Model (RSGM) \cite{de2022riemannian} accommodates a wide range of manifolds, including spheres and toruses, provided they satisfy mild conditions. The RSGM demonstrates that it is possible to extend diffusion models to compact Riemannian manifolds. The model also provides a formula for reversing diffusion on a manifold. Taking an intrinsic view, the RSGM approximates the sampling process on Riemannian manifolds using a Geodesic Random Walk. It is trained with a generalized denoising score matching objective.

In contrast, the Riemannian Diffusion Model (RDM) \cite{Huang2022RiemannianDM} employs a variational framework to generalize the continuous-time diffusion model to Riemannian manifolds. The RDM uses a variational lower bound (VLB) of the log-likelihood as its loss function. The authors of the RDM model have shown that maximizing this VLB is equivalent to minimizing a Riemannian score-matching loss. Unlike the RSGM, the RDM takes an extrinsic view, assuming that the relevant Riemannian manifold is embedded in a higher dimensional Euclidean space.

\subsubsection{Learned Manifolds}

According to the manifold hypothesis \cite{fefferman2016testing}, most natural data lies on manifolds with significantly reduced intrinsic dimensionality. Consequently, identifying these manifolds and training diffusion models directly on them can be advantageous due to the lower data dimensionality. Many recent works have built on this idea, starting by using an autoencoder to condense the data into a lower dimensional manifold, followed by training diffusion models in this latent space. In these cases, the manifold is implicitly defined by the autoencoder and learned through the reconstruction loss. In order to be successful, it is crucial to design a loss function that allows for the joint training of the autoencoder and the diffusion models.

The Latent Score-Based Generative Model (LSGM) \cite{vahdat2021score} seeks to address the problem of joint training by pairing a Score SDE diffusion model with a variational autoencoder (VAE) \cite{kingma2013auto,rezende2014stochastic}. In this configuration, the diffusion model is responsible for learning the prior distribution. The authors of the LSGM propose a joint training objective that merges the VAE's evidence lower bound with the diffusion model's score matching objective. This results in a new lower bound for the data log-likelihood. By situating the diffusion model within the latent space, the LSGM achieves faster sample generation than conventional diffusion models. Additionally, the LSGM can manage discrete data by converting it into continuous latent codes.

Rather than jointly training the autoencoder and diffusion model, the Latent Diffusion Model (LDM) \cite{rombach2022high} addresses each component separately. First, an autoencoder is trained to produce a low-dimensional latent space. Then, the diffusion model is trained to generate latent codes. DALLE-2 \cite{ramesh2022hierarchical} employs a similar strategy by training a diffusion model on the CLIP image embedding space, followed by training a separate decoder to create images based on the CLIP image embeddings.

Structure-guided Adversarial training of Diffusion Models (SADMs) \citep{yang2024structure}, for the first time, propose to utilize the structural information within the sample batch. Specifically, SADMs incorporate an adversarially-trained structural discriminator to enforce the preservation of manifold structure among samples within each training batch. This approach leverages the intrinsic data manifold to facilitate the generation of realistic samples, thereby significantly advancing the capabilities of previous diffusion models in tasks such as image synthesis and cross-domain fine-tuning.
\section{Connections with Other Generative Models}
\label{sec6}
\begin{figure*}[ht]
\centering
    \includegraphics[width=15cm]{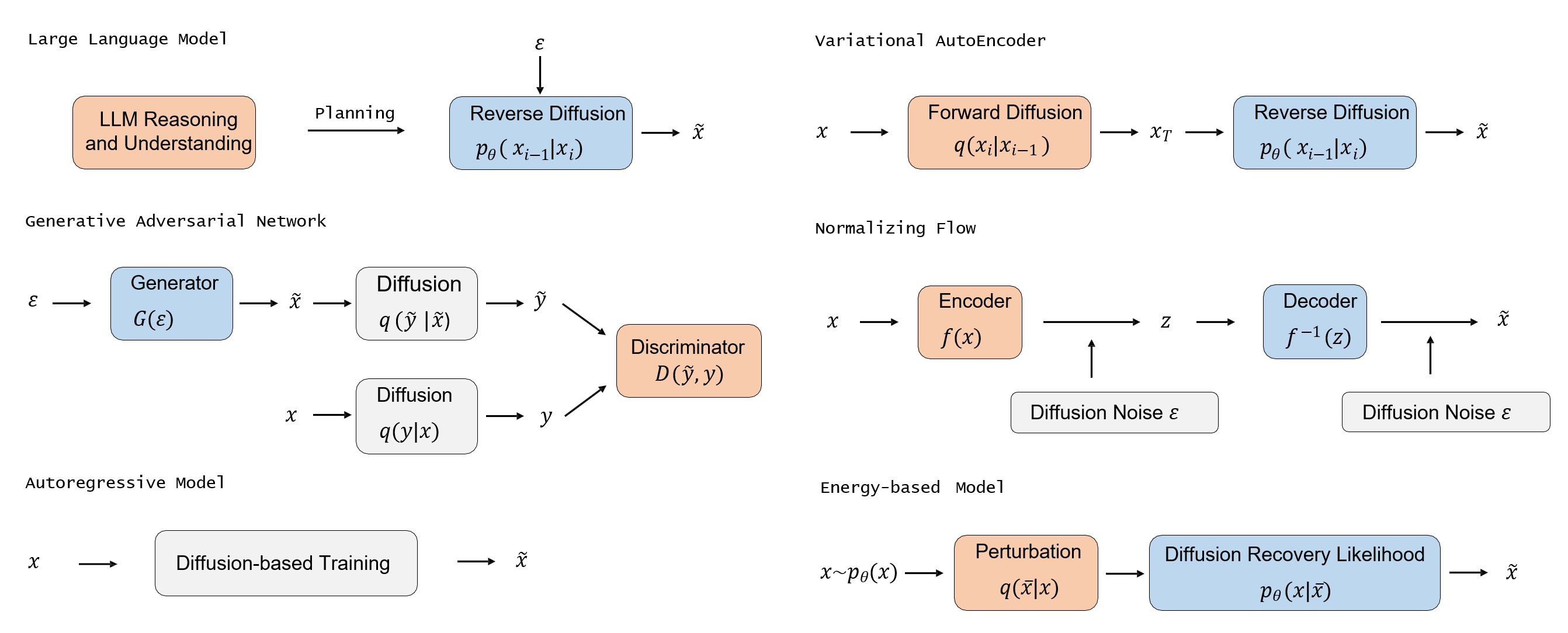}
        \caption{Illustrations of works incorporating diffusion models with other generative models, such as : LLM \cite{yang2024mastering} where a diffusion model is guided by the LLM planning, VAE \cite{rombach2022high} where a diffusion model is applied on a latent space, 
        GAN \cite{wang2022diffusion} where noise is injected to the discriminator input,
        normalizing flow \cite{zhang2021diffusion} where noise is injected in both forward and backward processes in the flow, autoregressive model \cite{hoogeboom2021autoregressive} where the training objective is similar to diffusion models, and EBM \cite{gao2020learning} where a sequence of EBMs is learned by diffusion recovery likelihood.}
        \vspace{-4mm}
\label{fig:connection}
\end{figure*}

In this section, we first introduce five other important classes of generative models and analyze their advantages and limitations.
Then we introduce how diffusion models are connected with them, and illustrate how these generative models are promoted by incorporating diffusion models.
The algorithms that integrate diffusion models with other generative models are summarized in \cref{tab3}, and we also provide a schematic illustration in \cref{fig:connection}.
\begin{table}[htp]
\center
\caption{Diffusion models are incorporated into different generative models.}
\begin{tabular}{lll}
\hline
\rowcolor{LightCyan}
Model & Article & Year \\ \hline
\multirow{6}{*}{Large Language Model} &
\cellcolor{Gray} Zhang et al. \cite{zhang2024realcompo} & \cellcolor{Gray}  2024 \\
&Yang et al. \cite{yang2025mmada}&2025\\&
\cellcolor{Gray} Yang et al. \cite{yang2024mastering} & \cellcolor{Gray}  2024 \\
&Tian et al. \cite{tian2024videotetris}&2024\\&
\cellcolor{Gray} Wang et al. \cite{wang2025revolutionizing} & \cellcolor{Gray}  2025 \\
&Zeng et al. \cite{zeng2024trans4d}&2024\\
\cline{1-3}
\multirow{3}{*}{Variational Auto-Encoder} &
\cellcolor{Gray} Luo et al. \cite{luo2022understanding} & \cellcolor{Gray}  2022 \\
& Hunag et al. \cite{huang2021variational} & 2021 \\
& \cellcolor{Gray} Vadhat et al. \cite{vahdat2021score} & \cellcolor{Gray} 2021 \\ \cline{1-3}
\multirow{2}{*}{{Generative Adversarial Network}} & Wang et al. \cite{wang2022diffusion}& 2022\\
& \cellcolor{Gray} Yang et al. \cite{yang2024structure} & \cellcolor{Gray} 2021\\ \cline{1-3}
\multirow{5}{*}{Normalizing Flow} & Zhang et al.\cite{zhang2021diffusion} & 2021\\
& \cellcolor{Gray} Gong et al. \cite{gong2021interpreting} & \cellcolor{Gray} 2021 \\ 
& kim et al. \cite{kimmaximum} & 2022 \\&
\cellcolor{Gray} Wang et al. \cite{wang2024rectified} & \cellcolor{Gray}  2024 \\
&Yang et al. \cite{yang2024consistencyfm}&2024\\\cline{1-3}
\multirow{4}{*}{Autoregressive Model} 
&\cellcolor{Gray} Meng et al.\cite{meng2020autoregressive} & \cellcolor{Gray} 2020 \\
& Meng et al.\cite{meng2021improved} & 2021 \\
& \cellcolor{Gray}Hoogeboom et al.\cite{hoogeboom2021autoregressive} & \cellcolor{Gray}2021 \\
&  Rasul et al. \cite{pmlr-v139-rasul21a} &  2021\\ \cline{1-3}
\multirow{2}{*}{Energy-based Model} & \cellcolor{Gray}Gao et al. \cite{gao2020learning} & \cellcolor{Gray}2021 \\
& Yu et al. \cite{yu2022latent} & 2022 \\ \bottomrule

\end{tabular}
\label{tab3}
\end{table}

\subsection{Large Language Models and Connections with Diffusion Models}
Large Language Models (LLMs) \citep{brown2020language,anil2023palm,achiam2023gpt,jiang2024mixtral,yang2024buffer,yang2025reasonflux,zou2025reasonflux,wang2025co} have profoundly impacted the AI community, showcasing advanced language comprehension and reasoning abilities. Recent works have begun extending these impressive reasoning capabilities to visual generative tasks and overall generation planning, establishing powerful synergies between language understanding and multimodal generation.

The collaboration between LLMs \citep{chatgpt2022introducing,openai2023gpt,yang2024buffer} and diffusion models \citep{ramesh2022hierarchical,betker2023improving,yang2023improving,yang2024crossmodal} can significantly improve text-image alignment and the quality of generated content \citep{zhang2024realcompo,lian2023llm,tian2024videotetris}. For instance, RealCompo \citep{zhang2024realcompo} utilizes LLMs to enhance the compositional generation of diffusion models by generating images grounded on bounding box layouts from the LLM. EditWorld \cite{yang2024editworld} composes a set of LLMs and pretrained diffusion models to generate an image editing dataset containing numerous instructions with world knowledge \cite{ha2018world}. VideoTetris \citep{tian2024videotetris} leverages LLMs to decompose text prompts along the temporal axis for guiding video generation with smoother and more reasonable transitions.

Beyond traditional text-to-image synthesis, recent advances have explored more sophisticated integrations. MMaDA \cite{yang2025mmada} demonstrates how LLMs can guide multi-modal discrete diffusion processes, where the language model's understanding of cross-modal relationships enhances the generation of complex discrete sequences across different modalities. In reinforcement learning contexts, TraceRL \cite{wang2025revolutionizing} combines the reasoning capabilities of LLMs with discrete diffusion models to generate optimal action sequences, where the language model provides high-level policy guidance while the diffusion process handles fine-grained action sampling.

Notably, RPG \cite{yang2024mastering} leverages the vision-language prior of multimodal LLMs to reason out complementary spatial layouts from text prompts, manipulating object compositions for diffusion models in both text-guided image generation and editing processes, achieving state-of-the-art performance in compositional synthesis scenarios.

These developments highlight a broader trend toward intelligent generation systems where LLMs serve as high-level planners and coordinators, while diffusion models handle the detailed generation process, creating more controllable and semantically coherent outputs across various domains and modalities.

\subsection{Variational Autoencoders and Connections with Diffusion Models}
Variational Autoencoders \cite{rezende2014stochastic,doersch2016tutorial,kingma2019introduction} aim to learn both an encoder and a decoder
to map input data to values in a continuous latent space.
In these models, the embedding can be interpreted as a latent variable in a probabilistic generative model, and a probabilistic decoder can be formulated by a parameterized likelihood function.
In addition, the data $\mathbf{x}$ is assumed to be generated by some unobserved latent variable $\mathbf{z}$ using conditional distribution $p_\theta(\mathbf{x}\mid\mathbf{z})$, and $ q_\phi(\mathbf{z}\mid\mathbf{x})$ is used to approximately inference $\mathbf{z}$.
To guarantee an effective inference, a variational Bayes approach is used to maximize the evidence lower bound:
\begin{align}
\mathcal{L}(\phi, \theta; \mathbf{x})
&=
\E_{q(\mathbf{z}\mid\mathbf{x})}\left[ \log p_\theta(\mathbf{x},\mathbf{z}) - \log q_\phi(\mathbf{z}\mid\mathbf{x}) \right]
\label{eq:elbo}
\end{align}
with $\mathcal{L}(\phi, \theta; \mathbf{x}) \leq \log p_\theta(\mathbf{x})$.
Provided that the parameterized likelihood function $p_\theta(\mathbf x\mid\mathbf z)$ and the parameterized posterior approximation $q_\phi(\mathbf z\mid\mathbf x)$ can be computed in a point-wise way and are
differentiable with their parameters, the ELBO can be maximized with gradient descent.
This formulation allows flexible choices of encoder and decoder models.
Typically, these models are represented by exponential family distributions whose parameters are generated by multi-layer neural networks.

The DDPM can be conceptualized as a hierarchical Markovian VAE with a fixed encoder. Specifically, DDPM's forward process functions as the encoder, and this process is structured as a linear Gaussian model (as described by \cref{eq:2}). The DDPM's reverse process, on the other hand, corresponds to the decoder, which is shared across multiple decoding steps. The latent variables within the decoder are all the same size as the sample data.

In a continuous-time setting, Song et al. (2021) \cite{song2020score}, Huang et al. (2021) \cite{huang2021variational}, and Kingma et al. (2021) \cite{kingma2021variational} demonstrate that the score matching objective may be approximated by the Evidence Lower Bound (ELBO) of a deep hierarchical VAE. Consequently, optimizing a diffusion model can be seen as training an infinitely deep hierarchical VAE---a finding that supports the common belief that Score SDE diffusion models can be interpreted as the continuous limit of hierarchical VAEs.

The Latent Score-Based Generative Model (LSGM) \cite{vahdat2021score} furthers this line of research by illustrating that the ELBO can be considered a specialized score matching objective in the context of latent space diffusion. Though the cross-entropy term in the ELBO is intractable, it can be transformed into a tractable score matching objective by viewing the score-based generative model as an infinitely deep VAE.

\subsection{Generative Adversarial Networks and Connections with Diffusion Models}
Generative Adversarial Networks (GANs) \cite{goodfellow2014generative,creswell2018generative,gui2021review}  mainly consist of two models: a generator $G$ and a discriminator $D$. These two models are typically constructed by neural networks but could be implemented in any form of a differentiable system that maps input data from one space to another.
The optimization of GANs can be viewed as a mini-max optimization problem with value function $V(G, D)$:
\begin{equation}
\min_G \max_D \mathbb{E}_{\mathbf{x} \sim p_{\text{data}}(\mathbf {x})}[\log D(\mathbf{x})] + \mathbb{E}_{\mathbf{z} \sim p_{\mathbf{z}}(\mathbf{z})}[\log (1 - D(G(\mathbf{z})))].
\end{equation}
The generator $G$ aims to generate new examples and implicitly model the data distribution.
The discriminator $D$ is usually a binary classifier that is used to identify generated examples from true examples with maximally possible accuracy.
The optimization process ends at a saddle point that produces a minimum about the generator and a maximum about the discriminator.
Namely, the goal of GAN optimization is to achieve Nash equilibrium \cite{ratliff2013characterization}.
At that point, the generator can be considered that it has captured the accurate distribution of real examples.

One of the issues of GAN is the instability in the training process, which is mainly caused by the non-overlapping between the distribution of input data and that of the generated data.
One solution is to inject noise into the discriminator input for widening the support of both the generator and discriminator distributions.
Taking advantage of the flexible diffusion model, Wang et al. (2022) \cite{wang2022diffusion} inject noise to the discriminator with an adaptive noise schedule determined by a diffusion model.
On the other hand, GAN can facilitate sampling speed of diffusion models.
Xiao et al. (2021) \cite{xiao2021tackling} show that slow sampling is caused by the Gaussian assumption in the denoising step, which is justified only for small step sizes.
As such, each denoising step is modeled by a conditional GAN, allowing larger step size.
To ensure the diffusion model captures authentic manifold structures in the data distribution, SADM \citep{yang2024structure}  advocates adversarial training of the diffusion generator against a novel structure discriminator in a minimax game, distinguishing real manifold structures from the generated ones.

\subsection{Normalizing Flows and Connections with Diffusion Models}
Normalizing flows~\cite{DinSohBen16,RezMoh15}
are generative models that generate tractable distributions to model high-dimensional data~\cite{KinDha18,DinSohLar19}.
Normalizing flows can transform simple probability distribution into an extremely complex probability distribution, which can be used in generative models, reinforcement learning, variational inference, and other fields.
Existing normalizing flows are constructed based on the change of variable formula~\cite{DinSohBen16,RezMoh15}.
The trajectory in normalizing flows is formulated by a differential equation.
In the discrete-time setting, the mapping from data $\mathbf{x}$ to latent $\mathbf{z}$ in normalizing flows is a composition of a sequence of bijections, taking the form of $ F = F_N \circ F_{N-1} \circ \ldots \circ F_1$. The trajectory $\{ \mathbf{x}_1,\mathbf{x}_2,\ldots\mathbf{x}_N \} $ in normalizing flows satisfies :
\begin{equation}
    \mathbf{x}_i = F_i(\mathbf{x}_{i-1}, \theta), \ \mathbf{x}_{i-1} = F^{-1}_i(\mathbf{x}_{i}, \theta)
\end{equation}
for all $i \leq N$.

Similar to the continuous setting, normalizing flows allow for the retrieval of the exact log-likelihood through a change of variable formula. However, the bijection requirement limits the modeling of complex data in both practical and theoretical contexts \cite{WuKohNoe20,CorCateDel20}. Several works attempt to relax this bijection requirement \cite{WuKohNoe20,DinSohLar19}. For example, DiffFlow \cite{zhang2021diffusion} introduces a generative modeling algorithm that combines the benefits of both flow-based and diffusion models. As a result, DiffFlow produces sharper boundaries than normalizing flow and learns more general distributions with fewer discretization steps compared to diffusion probabilistic models. Implicit Nonlinear Diffusion Model (INDM) \cite{kimmaximum} optimizes the pre-encoding process of latent diffusion, which first encodes the original data into the latent space using normalizing flow, and then performs diffusion in the latent space. Using a non-linear diffusion process, INDM can effectively improve the likelihood and the sampling speed.

To scale up the training of CNFs, recent works propose efficient simulation-free approaches \citep{lipman2022flow,albergo2022building,liu2022flow} by parameterizing
a vector field which flows from noise samples to data samples. Lipman et al. (2022) \cite{lipman2022flow} propose Flow Matching
(FM) to train CNFs based on constructing explicit conditional probability paths between the noise distribution and each data sample. Wang et al. (2024) \cite{wang2024rectified} conduct an in-depth analysis of the essence of rectification in rectified flow \cite{liu2022flow} and extend it to rectified diffusion. Besides, they identify that it is not straightness but first-order property is the essential training target of rectified diffusion with theoretical derivations.
Yang et al. (2024) further propose Consistency Flow Matching \cite{yang2024consistencyfm}, a novel FM method that explicitly enforces self-consistency in the velocity field. Consistency Flow Matching \cite{yang2024consistencyfm} directly defines
straight flows starting from different times to the same endpoint, imposing constraints on their velocity values:
\begin{equation}\label{loss-1}
    \begin{aligned}
        & \mathcal{L}_{\theta} = E_{t\sim \mathcal{U}}E_{x_t,x_{t+\Delta t}} || f_\theta(t,x_{t}) - f_{\theta^-}(t+\Delta t,x_{t+\Delta t})||_2^2+ \alpha|| v_\theta(t,x_{t}) - v_{\theta^-}(t+\Delta t,x_{t+\Delta t})||_2^2, \\
        & f_\theta(t,x_t) = x_t + (1-t)*v_\theta(t,x_t), \\
    \end{aligned}
\end{equation}
where $\mathcal{U}$ is the uniform distribution on $[0,1-\Delta t]$, $\alpha$ is a positive scalar, $\Delta t$ denotes a time interval which is a small and positive scalar. $\theta^-$ denotes the running average of past values of $\theta$ using  exponential moving average (EMA), $x_t$ and $x_{t+\Delta t}$ follows a pre-defined distribution which can be efficiently sampled, for example, VP-SDE \citep{ho2020denoising} or OT path \citep{lipman2022flow}. In this way, Consistency Flow Matching \cite{yang2024consistencyfm}  innovatively bridges consistency models and flow matching models through the novel concept of straight flows characterized by velocity consistency. 

\subsection{Autoregressive Models and Connections with Diffusion Models}
Autoregressive Models (ARMs) work by decomposing the joint distribution of data into a product of conditional distributions using the probability chain rule:
\begin{equation}
    \log p(\mathbf{x}_{1:T}) = \sum_{t=1}^{ T} \log p(x_t\mid\mathbf{x}_{<t})
\end{equation}
where $\mathbf{x}_{< t}$ is a shorthand for $x_1,x_2,\ldots, x_{t-1}$ \cite{bengio2000takingcurse,larochelle2011nade}.
Recent advances in deep learning have facilitated significant progress for various data modalities \cite{meng2020improved,chang2022maskgit,savinov2021step}, such as images \cite{oord2016pixelrnn,child2019sparsetransformer}, audio \cite{oord2016wavenet,kalchbrenner2018efficientneural}, and text \cite{bengio2003neural,graves2013rnn,melis2018on,merity2018regularizing,brown2020language}.
Autoregressive models (ARMs) offer generative capabilities through the use of a single neural network. Sampling from these models requires the same number of network calls as the data's dimensionality. While ARMs are effective density estimators, sampling is a continuous, time-consuming process—particularly for high-dimensional data.

The Autoregressive Diffusion Model (ARDM) \cite{hoogeboom2021autoregressive}, on the other hand, is capable of generating arbitrary-order data, including order-agnostic autoregressive models and discrete diffusion models as special cases \cite{sohldickstein2015diffusion,hoogeboom2021argmax,austin2021structured}. Instead of using causal masking on representations like ARMs, the ARDM is trained with an effective objective that mirrors that of diffusion probabilistic models. At the testing stage, the ARDM is able to generate data in parallel---enabling its application to a range of arbitrary-generation tasks.

Ment et al.(2021) \cite{meng2021improved} incorporates randomized smoothing into
autoregressive generative modeling, in order to improve the sample quality. The original data distribution is smoothed by convolving it with a smooth distribution, e.g., a Gaussian or Laplacian kernel. The smoothed data distribution is learned by autoregressive model, and then the learned distribution is denoised by either applying gradient-based denoising approach or introducing another conditional autoregressive model. By choosing the level of smoothness appropriately, the proposed method can improve the sample quality of existing autoregressive models while retaining reasonable likelihoods.

On the other hand, Autoregressive conditional score models (AR-CSM) \cite{meng2020autoregressive} proposes a score matching method to model the conditional distribution of autoregressive model. The score function of conditional distribution, i.e., $\nabla_{x_t}\log p(x_t\mid\mathbf{x}_{<t})$, does not need to be normalized and thus one can use more flexible and advanced neural networks in the model. Furthermore, the univariate conditional score function can be efficiently estimated, even though the dimension of original data might be very high. For inference, AR-CSM uses Langevin dynamics that only need the score function to sample from a density.
\subsection{Energy-based Models and Connections with Diffusion Models}
Energy-based Models (EBMs) 
\cite{lecun2006tutorial,ngiam2011learning,kim2016deep,zhao2016energy,goyal2017variational,xie2016theory,finn2016connection,gao2018learning,kumar2019maximum,nijkamp2019learning,du2019implicit,grathwohl2019your,desjardins2011tracking,gao2020flow,che2020your,grathwohl2020cutting,qiu2019unbiased,rhodes2020telescoping} 
can be viewed as one generative version of discriminators \cite{jin2017introspective, lazarow2017introspective, lee2018wasserstein, grathwohl2020cutting}, while can be learned from unlabeled input data.
Let $\mathbf x \sim p_{\rm{data}}(\mathbf x)$ denote a training example, and $p_\theta(\mathbf x)$ denote a probability density function that aims to approximates $p_{\rm{data}}(\mathbf x)$.
An energy-based model is defined as:
\begin{equation}
         p_\theta(\mathbf x) = \frac{1}{Z_\theta}\exp(f_\theta(\mathbf x)),
\end{equation}
where $Z_\theta = \int \exp(f_\theta(\mathbf x)) d\mathbf x$ is the partition function, which is analytically intractable for high-dimensional $\mathbf x$.
For images, $f_\theta(\textnormal x)$ is parameterized by a convolutional neural network with a scalar output. Salimans et al. (2021) \cite{salimans2021should} compare both constrained score models and energy-based models for modeling the score of the data distribution, finding
that constrained score models, i.e., energy based models, can perform just as well
as unconstrained models when using a comparable model structure.

Although EBMs have a number of desirable properties, two challenges remain for modeling high-dimensional data.
First, learning EBMs by maximizing the likelihood requires MCMC method to generate samples from the model, which can be very computationally expensive.
Second, as demonstrated in ~\cite{nijkamp2019anatomy}, the energy potentials learned with non-convergent MCMC are not stable, in the sense that samples from long-run Markov chains can be significantly different from the observed samples, and thus it is difficult to evaluate the learned energy potentials.
In a recent study, Gao et al. (2021) \cite{gao2020learning} present a diffusion recovery likelihood method to tractably learn samples from a sequence of EBMs in the reverse process of the diffusion model.
Each EBM is trained with recovery likelihood, which aims to maximize the conditional probability of the data at a certain noise level, given their noisy versions at a higher noise level.
EBMs maximize the recovery likelihood because it is more tractable than marginal likelihood, as sampling from the conditional distributions is much easier than sampling from the marginal distributions.

\section{Applications of Diffusion Models}
\label{sec7}
Diffusion models have recently been employed to address a variety of challenging real-world tasks due to their flexibility and strength. We have grouped these applications into six different categories based on the task: computer vision, natural language processing, temporal data modeling, multi-modal learning, robust learning, and interdisciplinary applications. For each category, we provide a brief introduction to the task, followed by a detailed explanation of how diffusion models have been applied to improve performance. \cref{tab4} summarizes the various applications that have made use of diffusion models.
\begin{table*}[htp]
\center
\caption{Summary of all the applications utilizing the diffusion models.}
\resizebox{0.95\textwidth}{!}{
\begin{tabular}{lll}
\hline
\rowcolor{LightCyan}
Primary & Secondary & Article \\ \hline
\multirow{6}{*} {{Computer Vision}} &
\cellcolor{Gray} &
  \cellcolor{Gray} \cite{Li2022SRDiffSI} \cite{saharia2022image},\cite{rombach2022high},\cite{lugmayr2022repaint},\cite{saharia2022palette},\cite{preechakul2022diffusion}, \cite{ho2022cascaded},\cite{batzolis2021conditional},\cite{ozbey2022unsupervised},\cite{chung2022score}\\  &  \multirow{-2}{*}{Super Resolution, Inpainting, Restoration, Translation, and Editing} \cellcolor{Gray}  & \cellcolor{Gray} \cite{song2021solving},\cite{chung2022mr},\cite{meng2021sdedit},\cite{kawar2022denoising},\cite{yang2023improving},\cite{yang2024mastering}\\ \cline{2-3}
 &  \multirow{1}{*}{Semantic Segmentation} &  \cite{baranchuk2021label},\cite{brempong2022denoising},\cite{graikos2022diffusion},\cite{xu2023open}\\ \cline{2-3}
 & \cellcolor{Gray}\multirow{1}{*}{Video Generation} &\cellcolor{Gray} \cite{harvey2022flexible},\cite{ho2022video},\cite{yang2022diffusion},\cite{zhang2022motiondiffuse},\cite{singer2022make},\cite{ho2022imagen},\cite{wu2022tune},\cite{qi2023fatezero},\cite{tian2024videotetris}\\ \cline{2-3}
 & \multirow{1}{*}{Point Cloud Completion and Generation} &  \cite{zhou20213d},\cite{luo2021diffusion},\cite{lyu2021conditional},\cite{liu2022let},\cite{zeng2022lion}\\ \cline{2-3}
 & \cellcolor{Gray} \multirow{1}{*}{Generating Data from Diffusion Models} & \cellcolor{Gray} \cite{yang2024editworld},\cite{brooks2023instructpix2pix},\cite{zhu2024distribution} \\ \cline{1-3}

\multirow{1}{*} {{Natural Language Generation}} &  \multirow{1}{*}{Natural Language Generation} & \cite{austin2021structured},\cite{li2022diffusion},\cite{chen2022analog},\cite{gong2023sequence},\cite{han2022ssd},\cite{dieleman2022continuous}\\ \cline{1-3}
\multirow{3}{*}{Temporal Data Modeling}&\cellcolor{Gray}
  \multirow{1}{*}{Time Series Imputation} & \cellcolor{Gray} \cite{tashiro2021csdi},\cite{alcaraz2022diffusion},\cite{park2021neural},\cite{liu2024retrieval}\\ \cline{2-3}
 &  \multirow{1}{*}{Time Series Forecasting} & \cite{rasul2021autoregressive},\cite{alcaraz2022diffusion},\cite{liu2024retrieval}\\ \cline{2-3}
   &\cellcolor{Gray}\multirow{1}{*}{Waveform Signal Processing} & \cellcolor{Gray}\cite{chen2020wavegrad},\cite{kong2020diffwave} \\ \cline{1-3}

\multirow{6}{*} {{Multi-Modal Learning}} & \multirow{1}{*}{Text-to-Image Generation} & \cite{avrahami2022blended},\cite{ramesh2022hierarchical},\cite{saharia2022photorealistic},\cite{pmlr-v162-nichol22a},\cite{gu2022vector},\cite{ruiz2022dreambooth},\cite{kawar2022imagic},\cite{zhang2023adding},\cite{yang2023improving},\cite{yang2024crossmodal},\cite{yang2024mastering},\cite{zhang2024itercomp},\cite{yang2025mmada}\\ \cline{2-3}
&\cellcolor{Gray} \multirow{1}{*}{Scene Graph-to-Image Generation} &\cellcolor{Gray} \cite{yang2022sgdiff}\\ \cline{2-3}
& \multirow{1}{*}{Text-to-3D/4D Generation} &\cite{xu2022dream3d},\cite{lin2022magic3d},\cite{poole2022dreamfusion},\cite{zeng2023ipdreamer},\cite{yang2024semanticsds},\cite{zeng2024trans4d} \\ \cline{2-3}
&\cellcolor{Gray} \multirow{1}{*}{Text-to-Motion Generation} & \cellcolor{Gray} \cite{zhang2022motiondiffuse,tevet2022human},\cite{kim2022flame}\\ \cline{2-3}
& \multirow{1}{*}{Text-to-Video Generation} & \cite{singer2022make},\cite{ho2022imagen},\cite{wu2022tune},\cite{qi2023fatezero},\cite{streamingt2v},\cite{yang2024crossmodal},\cite{tian2024videotetris}\\ \cline{2-3}
 & \cellcolor{Gray}  \multirow{1}{*}{Text-to-Audio Generation} & \cellcolor{Gray}\cite{popov2021grad},\cite{yang2022diffsound},\cite{wu2021itotts}, \cite{levkovitch2022zero},\cite{tae2021editts},\cite{huang2022prodiff},\cite{kim2022guided}\\ \cline{1-3}
  Robust Learning&  Robust Learning &   \cite{nie2022diffusion},\cite{yoon2021adversarial},\cite{blau2022threat},\cite{wang2022guided},\cite{wu2022guided},\cite{sun2022pointdp} \\\cline{1-3}
  \multirow{3}{*}{Interdisciplinary Applications} &\cellcolor{Gray}\multirow{1}{*}{Molecular Graph Modeling} &\cellcolor{Gray} \cite{jing2022torsional},\cite{hoogeboom2022equivariant},\cite{yang2024graphusion},\cite{xu2021geodiff},\cite{trippe2022diffusion}, \cite{huang2024binddm},\cite{huang2024proteinligand},\cite{huang2024interactionbased}\\ \cline{2-3}
  &\multirow{1}{*}{Material Design} & \cite{xie2021crystal},\cite{luo2022antigen}\\ \cline{2-3}
  &\cellcolor{Gray}\multirow{1}{*}{Medical Image Reconstruction} &\cellcolor{Gray} \cite{song2021solving},\cite{chung2022mr},\cite{chung2022come},\cite{chung2022score},\cite{peng2022towards},\cite{xie2022measurement}\\\hline
\end{tabular}}
\label{tab4}
\end{table*}
\subsection{Unconditional and Conditional Diffusion Models}
Before we introduce the applications of diffusion models, we illustrate two basic application paradigms of diffusion models, namely unconditional diffusion models and conditional diffusion models. As a generative model, the history of diffusion models is very similar to VAE, GAN, flow models, and other generative models. They all first developed unconditional generation, and then conditional generation followed closely. Unconditional generation is often used to explore the upper limit of the performance of the generative model, while conditional generation is more about application-level content because it can enable us to control the generation results according to our intentions. In addition to promising generation quality and sample diversity, diffusion models are especially superior in their controllability. The main algorithms of unconditional diffusion models have been well discussed in \cref{sec2,sec3,sec4,sec5}, in next part, we mainly discuss how conditional diffusion models are applied to different applications with different forms of conditions, and choose some typical scenarios for demonstrations.
\subsubsection{Conditioning Mechanisms in Diffusion Models}
Utilizing different forms of conditions to guide the generation directions of diffusion models are widely used, such as labels, classifiers, texts, images, semantic maps, graphs and so on. However, some of the conditions are structural and complex, thus the methods to condition on them are deserving discussion. There are mainly four kinds of conditioning mechanisms, including concatenation, gradient-based, cross-attention and adaptive layer normalization (adaLN). The concatenation means diffusion models concatenate informative guidance with intermediate denoised targets in diffusion process, such as label embedding and semantic feature maps. The gradient-based mechanism incorporates task-related gradient into the diffusion sampling process for controllable generation. For example, in image generation, one can train an auxiliary classifier on noisy images, and then use gradients to guide the diffusion sampling process towards an arbitrary class label. The cross-attention performs attentional message passing between the guidance and diffusion targets, which is usually conducted in a layer-wise manner in denoising networks. The adaLN mechanism follows the widespread usage of adaptive normalization layers \cite{perez2018film} in GANs \cite{karras2019style}, Scalable Diffusion Models \cite{peebles2022scalable} explores replacing standard layer norm layers in transformer-based diffusion backbones with adaptive layer normalization. Instead of directly learning dimension-wise scale and shift parameters, it regresses them from the sum of the time embedding and conditions. 
\subsubsection{Diffusion with DPO/RLHF}
\begin{figure}[ht]
\begin{center}
\centerline{\includegraphics[width=0.8\textwidth]{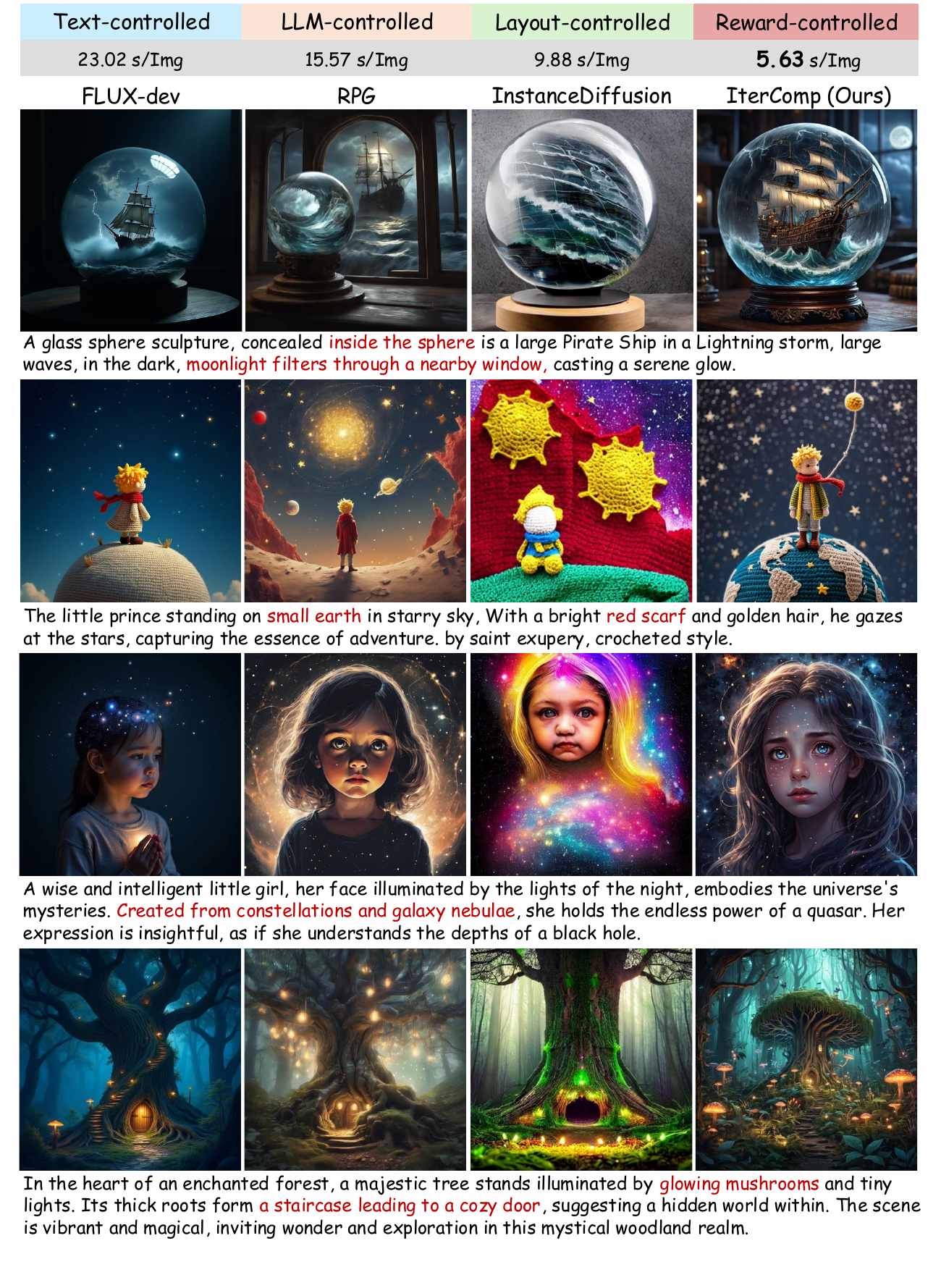}}
\vskip -0.05in
\caption{Qualitative comparison between IterComp \cite{zhang2024itercomp} and three types of compositional generation methods: text-controlled, LLM-controlled, and layout-controlled approaches. Colored text denotes the advantages of IterComp \cite{zhang2024itercomp} in generated images.}
\label{fig:itercomp}
\end{center}
\vskip -0.3in
\end{figure}

Building on the success of reinforcement learning from human feedback (RLHF) in Large Language Models (LLMs) \citep{ouyang2022training, bai2022training}, numerous methods in diffusion models have attempted to use similar approaches for model alignment \citep{lee2023aligning, fan2024reinforcement}. Some methods use a pretrained reward model or train a new one to guide the generation process. For instance, ImageReward \citep{xu2024imagereward} manually annotated a large dataset of human-preferred images and trained a reward model to assess the alignment between images and human preferences. Some methods bypass the training of a reward model and directly finetune diffusion models on human preference datasets \citep{yang2024using}. Diffusion-DPO \citep{wallace2024diffusion} reformulates Direct Preference Optimization (DPO) to account for a diffusion model's notion of likelihood, utilizing the evidence lower bound to derive a differentiable objective. Recently, Zhang et al. (2024) propose IterComp \cite{zhang2024itercomp} to iteratively align the base diffusion model with composition-aware model preferences from the model gallery, consisting of six powerful open-source diffusion models, effectively enhancing the performance of base model on conditional diffusion generation. As demonstrated in \cref{fig:itercomp}, IterComp \cite{zhang2024itercomp} outperforms other three types of conditional diffusion methods while achieving the best inference efficiency. 

\subsubsection{Condition Diffusion on Labels and Classifiers}
Conditioning diffusion process on the guidance of labels is a straight way to add desired properties into generated samples. However, when labels are limited, it is difficult to enable diffusion models to sufficiently capture the whole distribution of data. SGGM \cite{yang2023scorebased} proposes a self-guided diffusion process conditioning on the self-produced hierarchical label set, while You et al. (2023) \cite{you2023diffusion} demonstrate large-scale diffusion models and semi-supervised learners benefit mutually with a few labels via dual pseudo training. Dhariwal and Nichol \cite{dhariwal2021diffusion} proposes \textit{classifier guidance} to boost the sample quality of a diffusion model by using an extra trained classifier. Ho and Salimans \cite{ho2022classifier} jointly train a conditional and an unconditional diffusion model, and find that it is possible to combine the resulting conditional
and unconditional scores to obtain a trade-off between sample quality and diversity similar to that obtained by using classifier guidance.      


\subsubsection{Condition Diffusion on Texts, Images, and Semantic Maps} 
Recent researches begin to condition diffusion process on the guidance of more semantics, such as texts, images, and semantic maps, to better express rich semantics in samples. DiffuSeq \cite{gong2023sequence} conditions on texts and proposes a seq-to-seq diffusion framework that helps with four NLP tasks. SDEdit \cite{meng2021sdedit} conditions on a styled images to make image-to-image translation, while LDM \cite{rombach2022high} unifies these semantic conditions with flexible latent diffusion. Kindly note that if conditions and diffusion targets are of different modalities, pre-alignment \cite{yang2022sgdiff,ramesh2022hierarchical} is a practical way to strengthen the guided diffusion. unCLIP \cite{ramesh2022hierarchical} and ConPreDiff \cite{yang2023improving} leverage CLIP latents in text-to-image generation, which have align the semantics between images and texts. RPG \cite{yang2024mastering}  conditions on complementary rectangle and contour regions to enable compositional text-to-image generation and complex text-guided image editing. ContextDiff \cite{yang2024crossmodal} proposes a universal forward-backward consistent diffusion model for better conditioning on various input modalities.

\subsubsection{Condition Diffusion on Graphs}
Graph-structured data usually exhibits complex relations between nodes, thus conditioning on graphs are extremely hard for diffusion models. SGDiff \cite{yang2022sgdiff} proposes the first diffusion model specifically designed for scene graph to image generation with a novel masked contrastive pre-training. Such masked pre-training paradigm is general and can be extended to any cross-modal diffusion architectures for both coarse- and fine-grained guidance. Other graph-conditioned diffusion models are mainly studied for graph generation. Graphusion \cite{yang2024graphusion} conditions on the latent clusters of graph dataset to generate new 2D graphs that greatly align with data distribution. BindDM \cite{huang2024binddm}, IPDiff \cite{huang2024proteinligand} and IRDiff \cite{huang2024interactionbased} propose to condition on 3D protein graph to generate 3D molecules with equivariant diffusion.

\subsection{Computer Vision}
\subsubsection{Image Super Resolution, Inpainting, Restoration, Translation, and Editing}

Generative models have been used to tackle a variety of image restoration tasks including super-resolution, inpainting, and translation \cite{Li2022SRDiffSI,batzolis2021conditional,deng2009imagenet,esser2021taming,ramesh2021zero,isola2017image,ozbey2022unsupervised,zhao2022egsde}. Image super-resolution aims to restore high-resolution images from low-resolution inputs, while image inpainting revolves around reconstructing missing or damaged regions in an image.

Several methods make use of diffusion models for these tasks. For example, Super-Resolution via Repeated Refinement (SR3) \cite{saharia2022image} uses DDPM to enable conditional image generation. SR3 conducts super-resolution through a stochastic, iterative denoising process. The Cascaded Diffusion Model (CDM) \cite{ho2022cascaded} consists of multiple diffusion models in sequence, each generating images of increasing resolution. Both the SR3 and CDM directly apply the diffusion process to input images, which leads to larger evaluation steps.
\begin{figure}[htp]
\centering
\footnotesize
\includegraphics[width=0.6\textwidth]{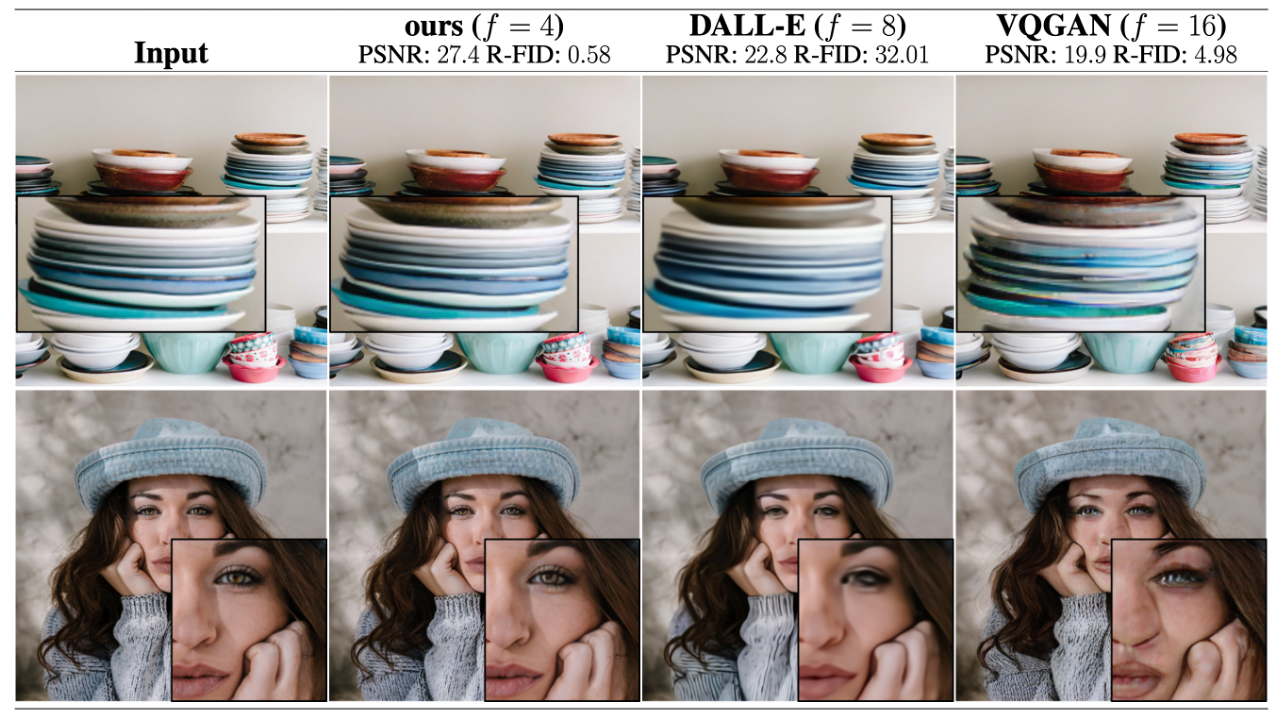}
\caption{\textbf{Image super resolution results produced by LDM~\cite{rombach2022high}}.}
\label{fig:ldm}
\end{figure}
In order to allow for the training of diffusion models with limited computational resources, some methods \cite{rombach2022high,vahdat2021score} have shifted the diffusion process to the latent space using pre-trained autoencoders. The Latent Diffusion Model (LDM) \cite{rombach2022high} streamlines the training and sampling processes for denoising diffusion models without sacrificing quality.

\begin{figure}[ht]
\centering
\footnotesize
\setlength{\tabcolsep}{1pt}\includegraphics[width=0.9\textwidth]{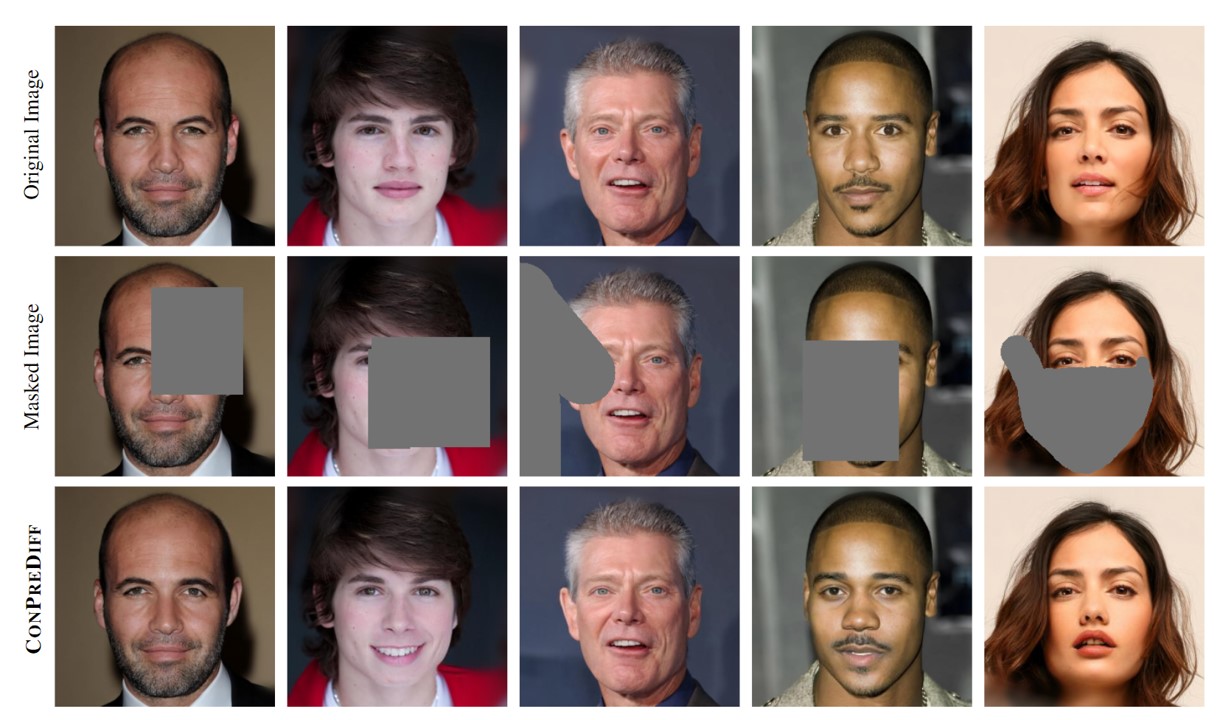}
\caption{\textbf{Image inpainting results produced by ConPreDiff~\cite{yang2023improving}}.}
\label{fig:paint}
\end{figure}

For inpainting tasks, RePaint \cite{lugmayr2022repaint} features an enhanced denoising strategy that uses resampling iterations to better condition the image. ConPreDiff \cite{yang2023improving} proposes a universal diffusion model based on context prediction to consistently improve unconditional/conditional image generation and image inpainting (see Figure \cref{fig:paint}). Meanwhile, Palette \cite{saharia2022palette} employs conditional diffusion models to create a unified framework for four image generation tasks: colorization, inpainting, uncropping, and JPEG restoration.
Image translation focuses on synthesizing images with specific desired styles \cite{isola2017image}. SDEdit \cite{meng2021sdedit} uses a Stochastic Differential Equation (SDE) prior to improve fidelity. Specifically, it begins by adding noise to the input image, then denoises the image through the SDE. Denoising Diffusion Restoration Models (DDRM) \cite{kawar2022denoising} takes advantage of a pre-trained denoising diffusion generative model for solving linear inverse problem, and demonstrates DDRM’s
versatility on several image datasets for super-resolution, deblurring, inpainting, and colorization under various amounts of measurement noise. \textbf{Please refer to \cref{text-to-image} for more text-to-image diffusion models.}

\subsubsection{Semantic Segmentation}
Semantic segmentation aims to label each image pixel according to established object categories. Generative pre-training can enhance the label utilization of semantic segmentation models, and recent work has shown that representations learned through DDPM contain high-level semantic information that is useful for segmentation tasks \cite{baranchuk2021label,graikos2022diffusion}. The few-shot method that leverages these learned representations has outperformed alternatives such as VDVAE \cite{child2020very} and ALAE \cite{pidhorskyi2020adversarial}. Similarly, Decoder Denoising Pretraining (DDeP) \cite{brempong2022denoising} integrates diffusion models with denoising autoencoders \cite{vincent2008extracting} and delivers promising results on label-efficient semantic segmentation. ODISE \cite{xu2023open} explores diffusion models for open-vocabulary segmentation tasks, and proposes a novel implicit captioner to generate captions for images for better utilizing pre-trained large-scale text-to-image diffusion models.

\subsubsection{Video Generation}

Generating high-quality videos remains a challenge in the deep learning era due to the complexity and spatio-temporal continuity of video frames \cite{yu2022generating,yang2022unsupervised}. Recent research has turned to diffusion models to improve the quality of generated videos \cite{ho2022video}. For example, the Flexible Diffusion Model (FDM) \cite{harvey2022flexible} uses a generative model to allow for the sampling of any arbitrary subset of video frames, given any other subset. The FDM also includes a specialized architecture designed for this purpose. Additionally, the Residual Video Diffusion (RVD) model \cite{yang2022diffusion} utilizes an autoregressive, end-to-end optimized video diffusion model. It generates future frames by amending a deterministic next-frame prediction, using a stochastic residual produced through an inverse diffusion process. \textbf{Please refer to \cref{text-to-video} for more text-to-video diffusion models.}

\begin{figure}[ht]
\centering
\footnotesize
\setlength{\tabcolsep}{1pt}\includegraphics[width=0.9\textwidth]{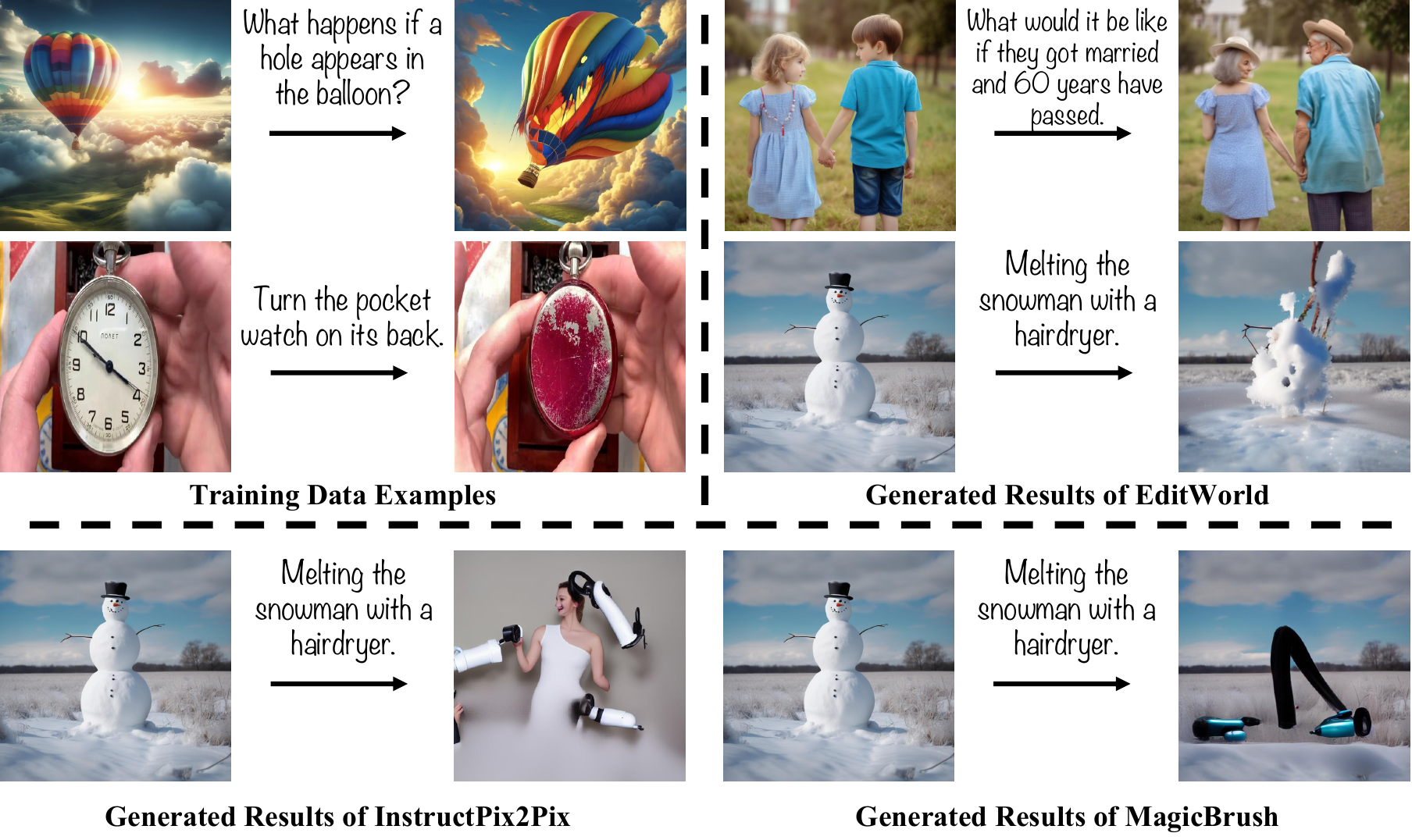}
\caption{\textbf{Comparing EditWorld \cite{yang2024editworld} with InstructPix2Pix and MagicBrush}.}
\label{fig:editworld}
\end{figure}

\subsubsection{Generating Data from Diffusion Models} 
\begin{figure}[ht]
\centering
\footnotesize
\setlength{\tabcolsep}{1pt}\includegraphics[width=0.9\textwidth]{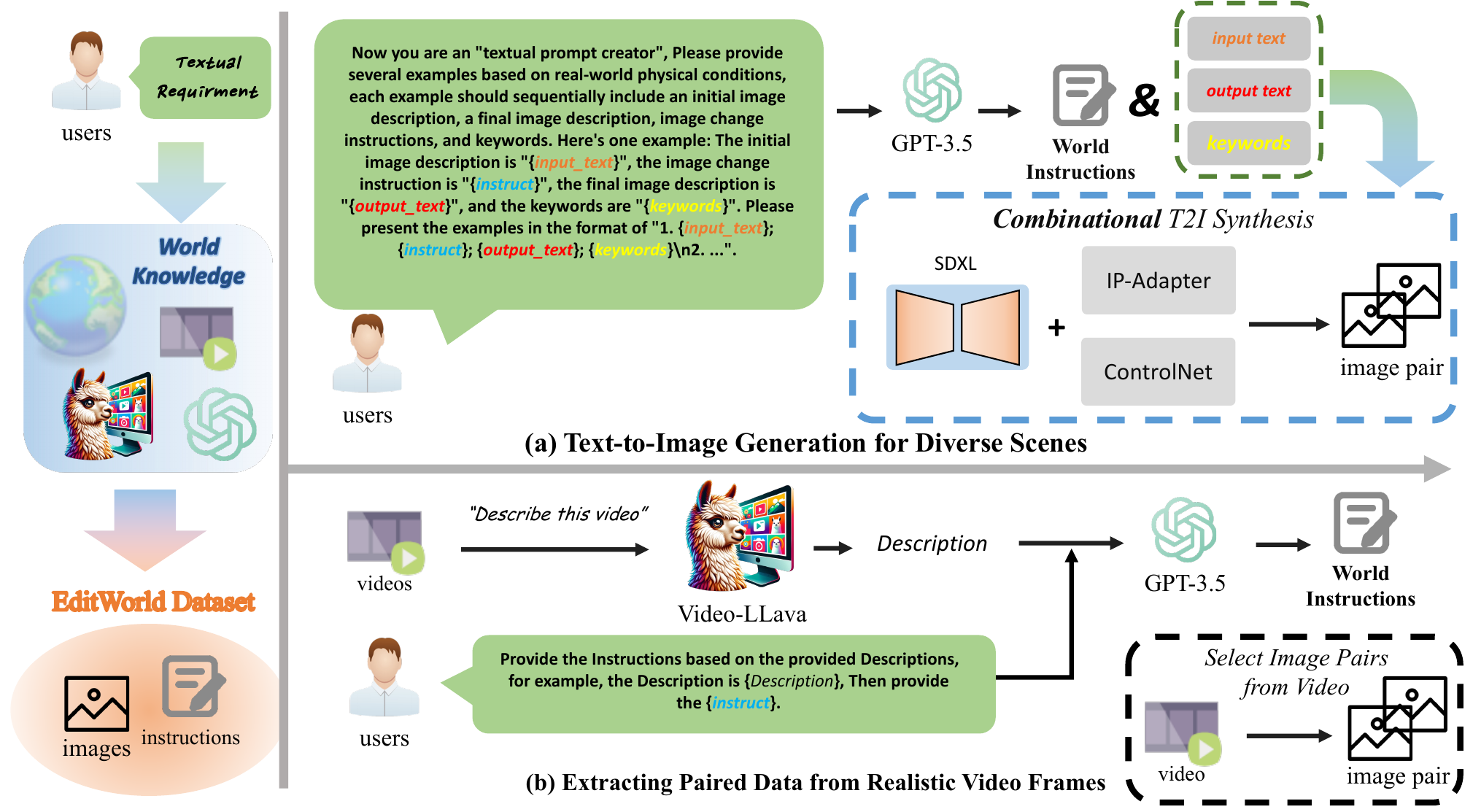}
\caption{\textbf{EditWorld \cite{yang2024editworld} generates a training dataset of world-instructed image editing from two different branches}.}
\label{fig:editworldpipe}
\end{figure}

Synthesizing datasets from generative models can effectively advance various tasks like classification \cite{wang2022knowda,bansal2023leaving,zhu2024distribution}. Recent works have begun to utilize diffusion models to achieve this goal for vision tasks. 
For example, \citet{trabucco2023effective} adopt diffusion models to make effective data augmentation for few-shot image classification. 
DistDiff \cite{zhu2024distribution} proposes a training-free
data expansion framework with a distribution-aware diffusion model. It constructs hierarchical prototypes to approximate the real data distribution, and optimizes latent data points in generation process with hierarchical energy guidance.
InstructPix2Pix \cite{brooks2023instructpix2pix} leverages two large pretrained models (i.e., GPT-3 and Stable Diffusion) to generate a large dataset
of input-goal-instruction triplet examples, and trains an instruction-following image editing model on the dataset. 
To enable image editing to reflect chalenging world knowledge and dynamics from both real physical world and virtual media, EditWorld \cite{yang2024editworld}, introduces a new task named world-instructed image editing, as the data examples presented in \cref{fig:editworld}. EditWorld proposes an innovative compositional framework with a set of pretrained LLMs and Diffusion Models, illustrated in \cref{fig:editworldpipe}, to synthesize a world-instructed training dataset for instruction-following image editing.

\subsubsection{Point Cloud Completion and Generation}
Point clouds are a critical form of 3D representation for capturing real-world objects. However, scans often generate incomplete point clouds due to partial observation or self-occlusion. Recent studies have applied diffusion models to address this challenge, using them to infer missing parts in order to reconstruct complete shapes. This work has implications for many downstream tasks such as 3D reconstruction, augmented reality, and scene understanding \cite{lyu2021conditional,luo2021score,zeng2022lion}.

Luo et al. 2021 \cite{luo2021diffusion} has taken the approach of treating point clouds as particles in a thermodynamic system, using a heat bath to facilitate diffusion from the original distribution to a noise distribution. Meanwhile, the Point-Voxel Diffusion (PVD) model \cite{zhou20213d} joins denoising diffusion models with the pointvoxel representation of 3D shapes. The Point Diffusion-Refinement (PDR) model \cite{lyu2021conditional} uses a conditional DDPM to generate a coarse completion from partial observations; it also establishes a point-wise mapping between the generated point cloud and the ground truth.
\begin{figure*}[!tp]
\centering
\footnotesize
\includegraphics[width=0.7\textwidth]{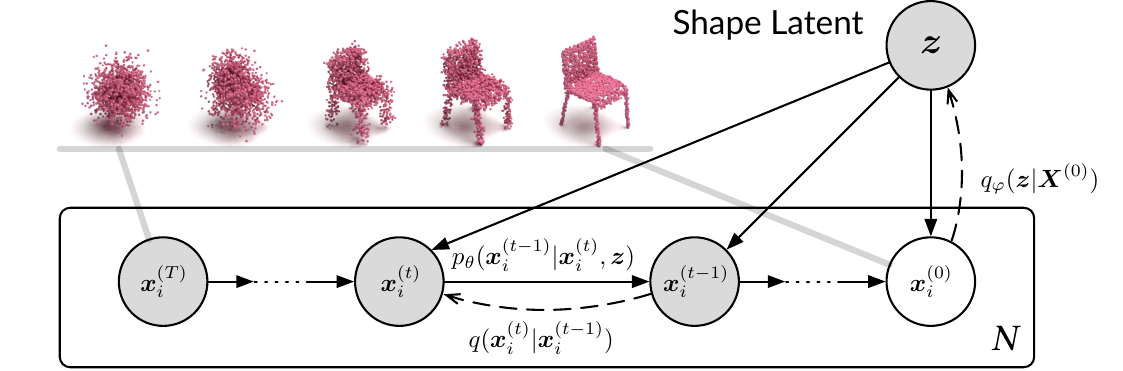}
\caption{\textbf{The directed graphical model of the diffusion process for point clouds~\cite{luo2021diffusion}}.}
\label{fig:pc}
\end{figure*}
\subsubsection{Anomaly Detection} Anomaly detection is a critical and challenging problem in machine learning \cite{zhao2019pyod,schlegl2017unsupervised} and computer vision \cite{yang2021visual}.
Generative models have been shown to own a powerful mechanism for anomaly detection \cite{han2022adbench,wyatt2022anoddpm,gedara2022remote}, modeling normal or healthy reference data. AnoDDPM \cite{wyatt2022anoddpm} utilizes DDPM to corrupt the input image and reconstruct a healthy approximation of the image.
These approaches may perform better than alternatives based on adversarial training as they can better model smaller datasets with effective sampling and stable training schemes.
DDPM-CD \cite{gedara2022remote} incorporates large numbers of unsupervised remote sensing images into the training process through DDPM.
Changes of remote sensed images is detected by utilizing a pre-trained DDPM and applying the multi-scale representations from the diffusion model decoder.

\subsection{Natural Language Generation}
Natural language processing aims to understand, model, and manage human languages from different sources such as text or audio. Text generation has become one of the most critical and challenging tasks in natural language processing \cite{iqbal2020survey,li2021textbox,li2021pretrained}, aiming to compose plausible and readable text in human language given input data (e.g., a sequence and keywords) or random noise.

Numerous approaches based on diffusion models have been developed for text generation. Discrete Denoising Diffusion Probabilistic Models (D3PM) \cite{austin2021structured} introduces diffusion-like generative models for character-level text generation \cite{chelba2013one}, generalizing the multinomial diffusion model \cite{hoogeboom2021argmax} by going beyond corruption processes with uniform transition probabilities. Analog Bits \cite{chen2022analog} generates analog bits to represent discrete variables and further improves sample quality with self-conditioning and asymmetric time intervals.

While large autoregressive language models (LMs) can generate high-quality text \cite{brown2020language,chowdhery2022palm,radford2019language,zhang2022opt}, reliable deployment in real-world applications typically requires controllable text generation to satisfy desired requirements (e.g., topic, syntactic structure). Controlling language model behavior without re-training remains a major challenge in text generation \cite{keskar2019ctrl,dathathri2019plug}. Although recent methods have achieved significant success on controlling simple sentence attributes (e.g., sentiment) \cite{krause2020gedi,yang2021fudge}, there has been limited progress on complex, fine-grained controls (e.g., syntactic structure).

To tackle more complex controls, Diffusion-LM \cite{li2022diffusion} proposes a new language model based on continuous diffusion that starts with Gaussian noise vectors and incrementally denoises them into word-corresponding vectors. The gradual denoising steps help produce hierarchical continuous latent representations, enabling simple gradient-based methods to accomplish complex control. Similarly, DiffuSeq \cite{gong2023sequence} conducts diffusion processes in latent space and proposes a new conditional diffusion model for challenging text-to-text generation tasks.

\begin{figure}[ht]
\centering
\includegraphics[width=1.0\textwidth]{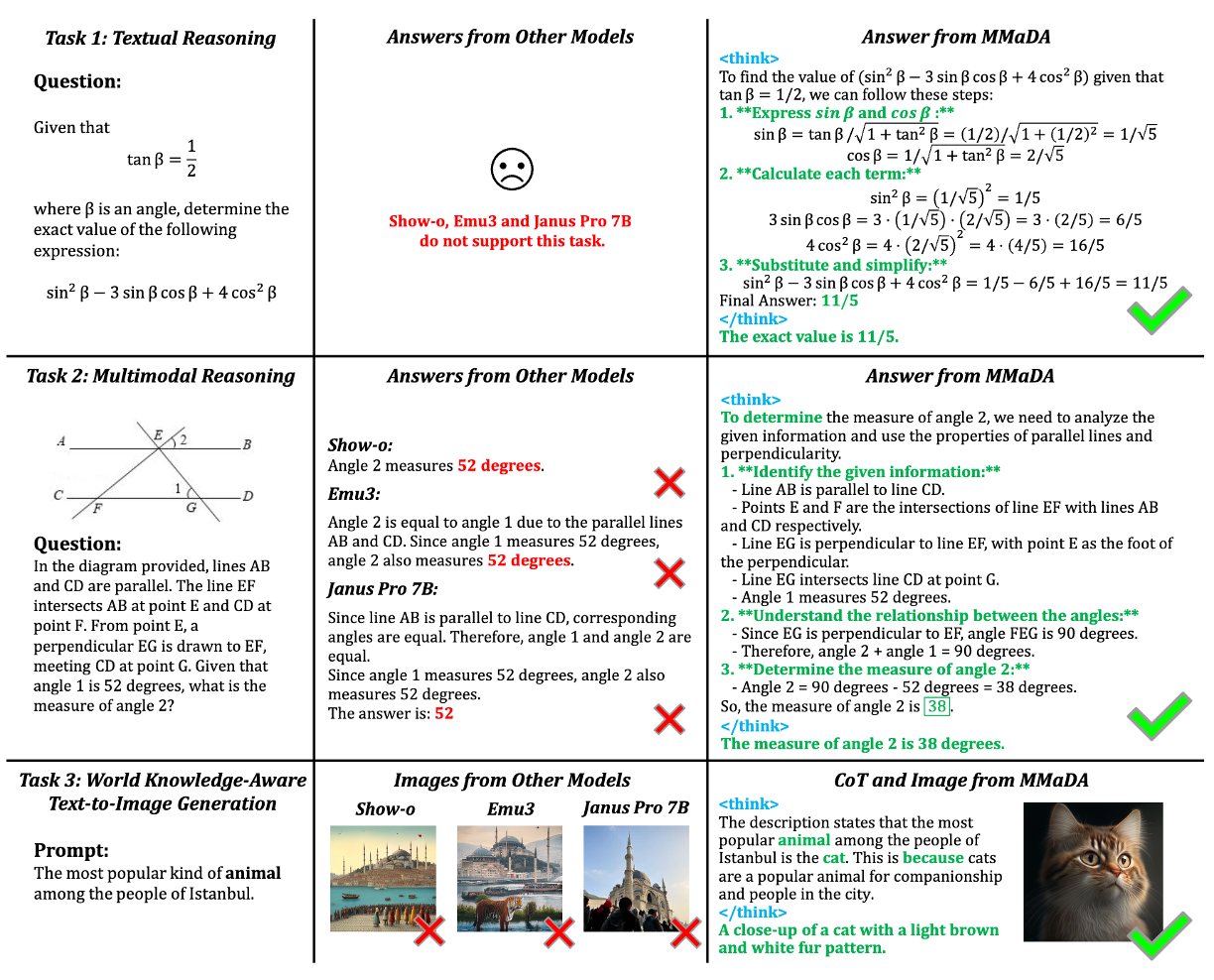}
\caption{\textbf{Performance comparison of MMaDA \cite{yang2025mmada} against other models on textual reasoning, multimodal reasoning, and world knowledge-aware text-to-image generation tasks.}.}
\label{fig:mmada}
\end{figure}

Recent advances have explored more specialized applications of diffusion-based language generation. MMaDA \cite{yang2025mmada} extends diffusion-based text generation to multi-modal settings, where textual content is generated in conjunction with other modalities through masked autoencoder-enhanced diffusion processes (in \cref{fig:mmada}). In reinforcement learning contexts, TraceRL \cite{wang2025revolutionizing} applies diffusion models to generate natural language descriptions of optimal action sequences, bridging the gap between policy learning and natural language explanation.

These developments demonstrate the versatility of diffusion models in natural language generation, from basic text synthesis to complex multi-modal and interactive applications, opening new avenues for controllable and context-aware language generation.

\subsection{Multi-Modal Generation}
\subsubsection{Text-to-Image Generation} \label{text-to-image}
Vision-language models have attracted a lot of attention recently due to the number of potential applications \cite{radford2021learning}.
Text-to-Image generation is the task of generating a corresponding image from a descriptive text \cite{du2022survey,kawar2022imagic,valevski2022unitune}.
%
%
Blended diffusion \cite{avrahami2022blended} utilizes both pre-trained DDPM \cite{dhariwal2021diffusion}
and CLIP \cite{radford2021learning} models, and it proposes a solution for region-based image editing for general purposes, which uses natural language guidance and is applicable to real and diverse images.
On the other hand, unCLIP (DALLE-2) \cite{ramesh2022hierarchical} proposes a two-stage approach, a prior model that can generate a CLIP-based image embedding conditioned on a text caption, and a diffusion-based decoder that can generate an image conditioned on the image embedding.
Recently, Imagen \cite{saharia2022photorealistic} proposes a text-to-image diffusion model and a  comprehensive benchmark for performance evaluation.
It shows that Imagen performs well against the state-of-the-art approaches including VQ-GAN+CLIP \cite{crowson2022vqgan}, Latent Diffusion Models \cite{lu2022dpm}, and DALL-E 2 \cite{ramesh2022hierarchical}.
\begin{figure*}[ht]
\centering
\includegraphics[width=1.\linewidth]{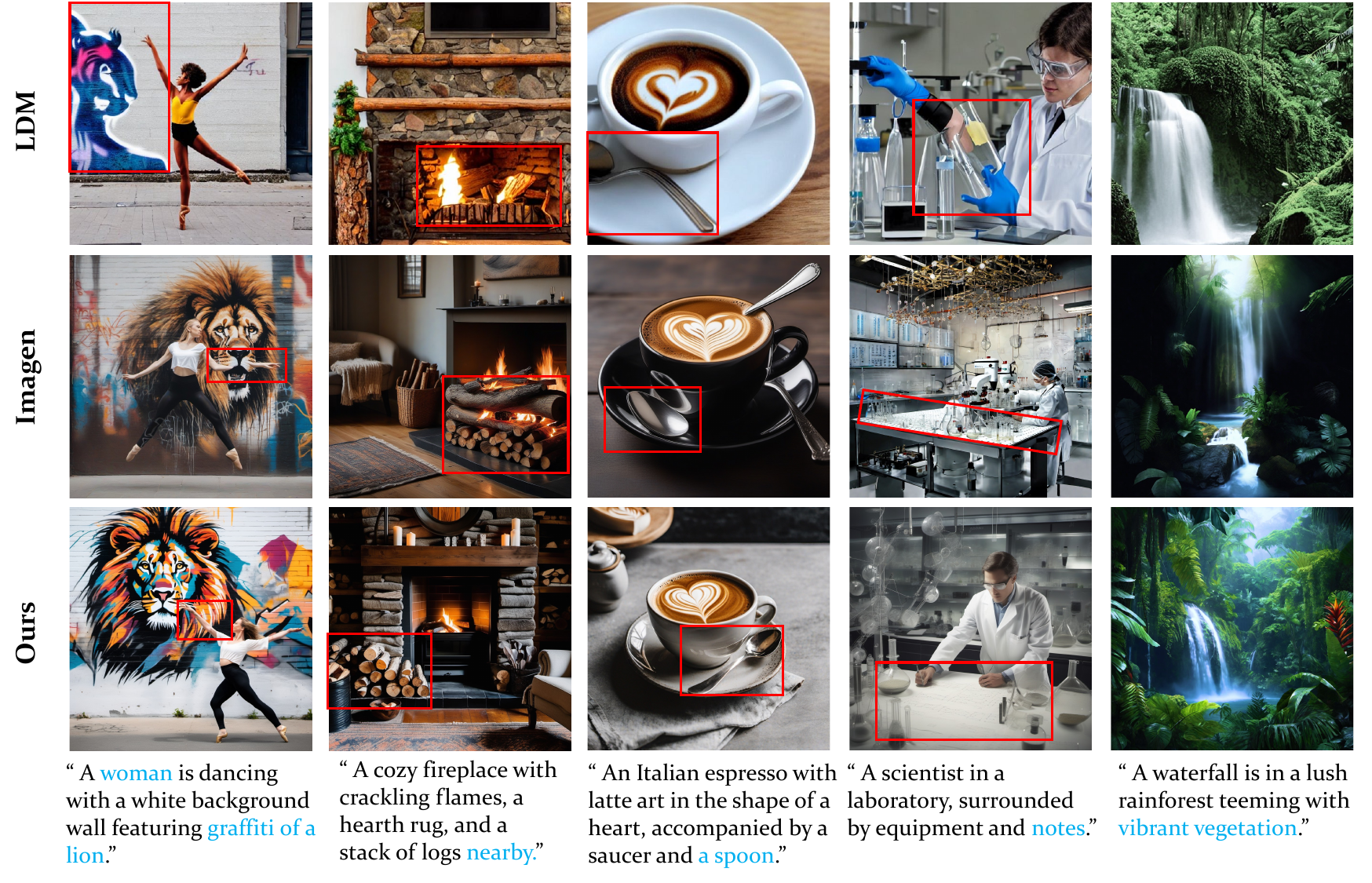}
\caption{\textbf{Synthesis examples demonstrating text-to-image capabilities of for various text prompts with LDM, Imagen, and
ContextDiff \cite{yang2024crossmodal}}.}
\vskip -0.1in
\label{fig:contextdiff}
\end{figure*}
Inspired by the ability of guided diffusion models \cite{dhariwal2021diffusion,ho2022classifier} to generate photorealistic samples and the ability of text-to-image models to handle free-form prompts,
GLIDE \cite{pmlr-v162-nichol22a} applies guided diffusion to the application of text-conditioned image synthesis.
VQ-Diffusion \cite{gu2022vector} proposes a vector-quantized diffusion model for text-to-image generation, and it eliminates the unidirectional bias and avoids accumulative prediction errors.
Versatile Diffusion \cite{xu2022versatile} proposes the first unified multi-flow multimodal diffusion framework, which supports image-to-text, image-variation, text-to-image, and text-variation, and can be further extended to other applications such as semantic-style disentanglement, image-text dual-guided generation, latent image-to-text-to-image editing, and more. Following Versatile Diffusion, UniDiffuser \cite{bao2023one} proposes a unified diffusion model framework based on Transformer, which can fit multimodal data distributions and simultaneously handle text-to-image, image-to-text, and joint image-text generation tasks. 
ConPreDiff \cite{yang2023improving} for the first time incorporates context prediction into text-to-image diffusion models, and significantly improves generation performance without additional inference costs.
ContextDiff \cite{yang2024crossmodal} proposes general contextualized diffusion model by incorporating the cross-modal context encompassing interactions and alignments into forward and reverse processes.
A qualitative comparison between these models are presented in \cref{fig:contextdiff}.

\begin{figure*}[htp]
\centering
\footnotesize
\includegraphics[width=0.9\textwidth]{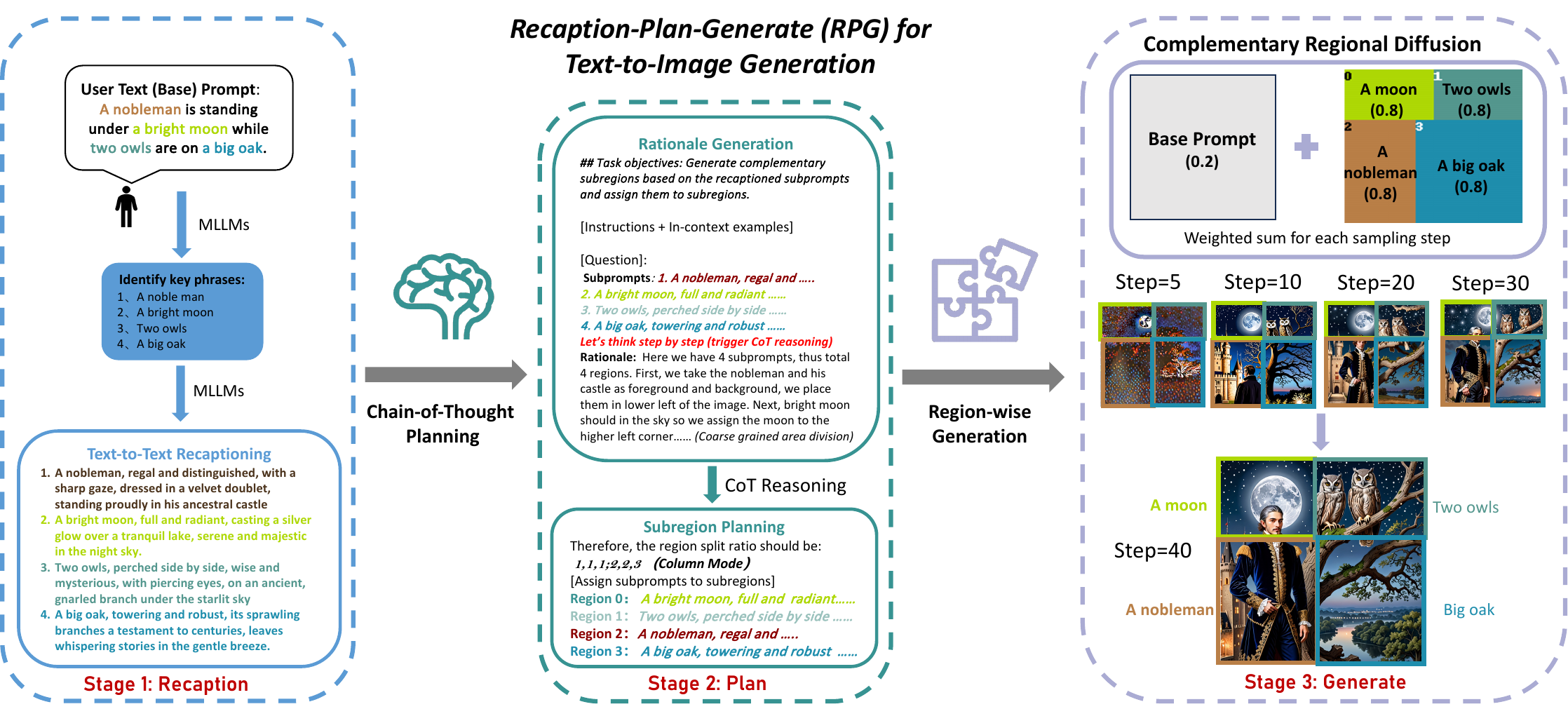}
\caption{\textbf{Overview of RPG \cite{yang2024mastering} framework for text-to-image generation.}}
\label{fig:rpg-generation}
\end{figure*}

\begin{figure*}[htp]
\centering
\footnotesize
\includegraphics[width=0.9\textwidth]{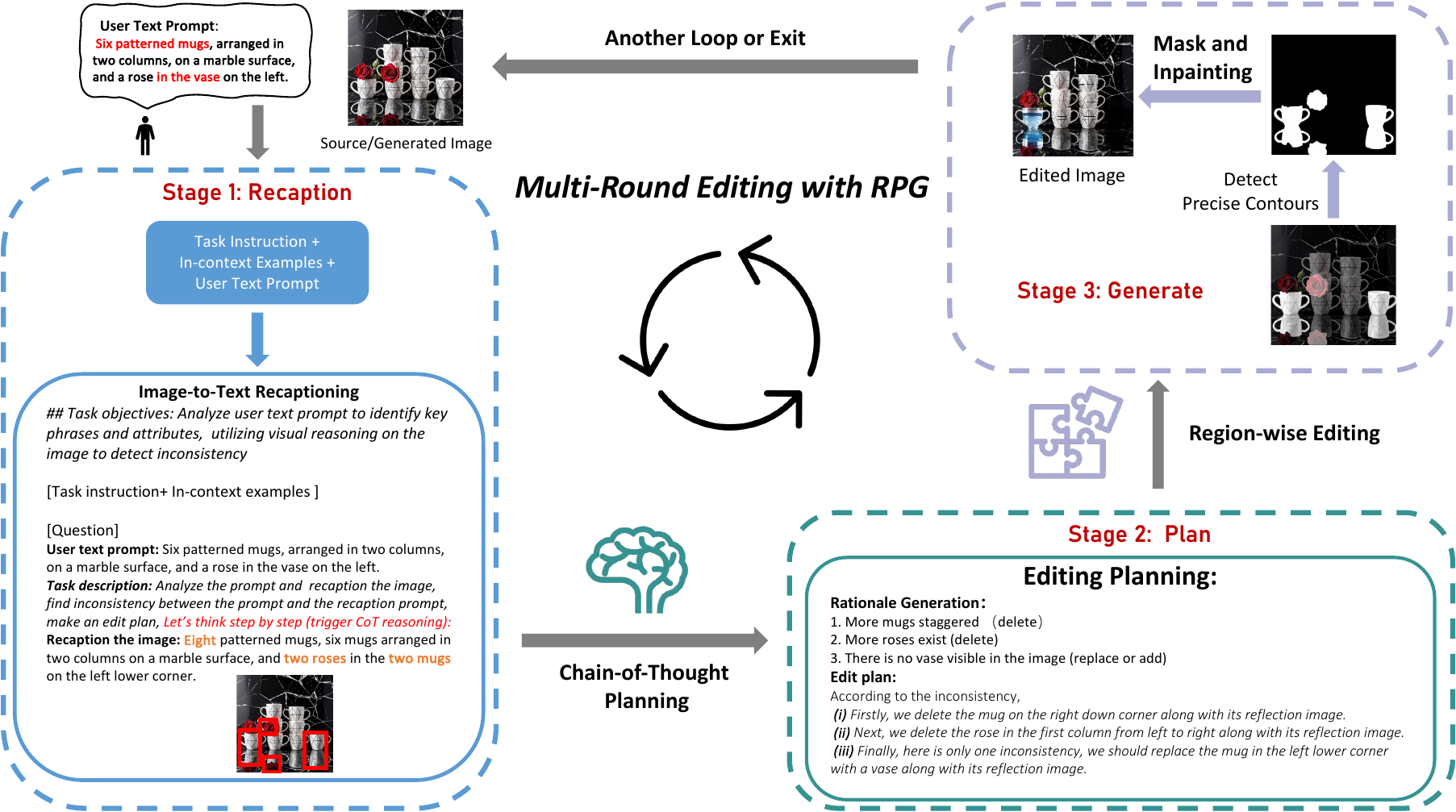}
\caption{\textbf{RPG \cite{yang2024mastering} can unify text-guided image generation and editing in a closed-loop approach.}}
\label{fig:rpg-edit}
\end{figure*}

\begin{figure*}[htp]
\centering
\footnotesize
\includegraphics[width=1.\textwidth]{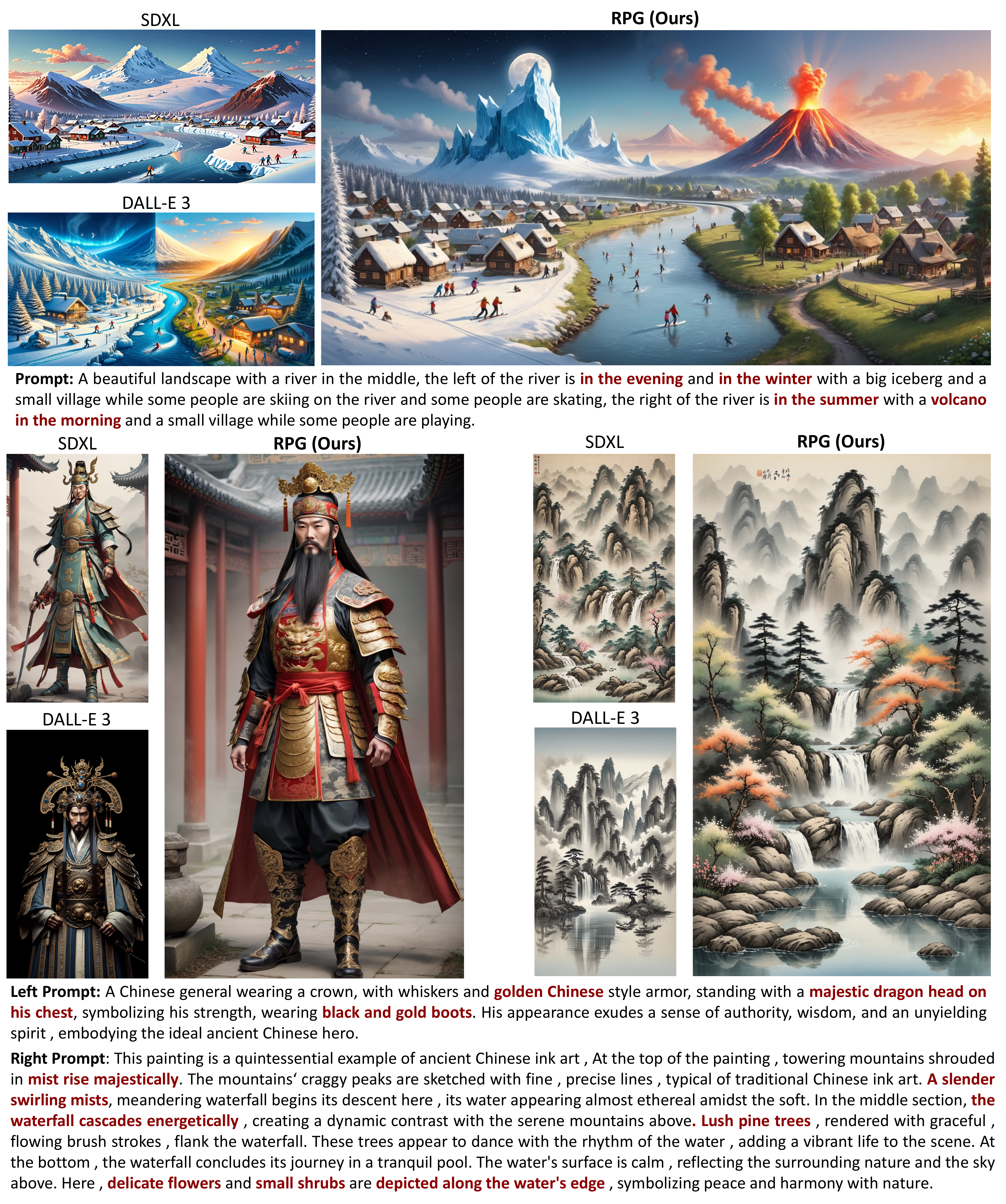}
\caption{\textbf{Compared to previous SOTA models, RPG \cite{yang2024mastering} exhibits a superior ability to express intricate and compositional text prompts within generated images (colored text denotes critical part).}}
\label{fig:rpg-result}
\end{figure*}

A new interesting line of diffusion model research is to leverage the pre-trained text-to-image diffusion model for more complex or fine-grained control of synthesis results. DreamBooth \cite{ruiz2022dreambooth} presents the first technique that tackles the new challenging problem of subject-driven generation, allowing users, from just a few casually captured images of a subject, to recontextualize
subjects, modify their properties, original art renditions, and more. 
Different from those image diffusion models conditioned on text prompts, ControlNet \cite{zhang2023adding} attempts to control pre-trained large
diffusion models to support additional semantic maps, like edge maps, segmentation maps, keypoints, shape normals, depths, etc. However, most methods often face challenges when handling complex text prompts involving multiple objects with multiple attributes and relationships. To this end, RPG \cite{yang2024mastering} proposes a brand new training-free text-to-image generation/editing framework harnessing the powerful chain-of-thought reasoning ability of multimodal LLMs \cite{zhang2023multimodal} to enhance the compositionality of text-to-image diffusion models. This new RPG framework unifies both text-guided image generation (in \cref{fig:rpg-generation})  and image editing (in \cref{fig:rpg-edit}) tasks in a closed-loop fashion. Notably, as demonstrated in \cref{fig:rpg-result}, RPG outperforms all SOTA methods, such as SDXL \cite{podell2023sdxl} and DALL-E 3 \cite{betker2023improving}, demonstrate its superiority. Furthermore, RPG framework is user-friendly, and can generalize to different MLLM architectures and diffusion backbones (e.g., ControlNet).

Recent advances have also explored building upon diffusion-based language model (DLLM) foundations for enhanced multi-modal generation capabilities. MMaDA \cite{yang2025mmada} leverages pre-trained DLLM backbones and extends them with masked autoencoder architectures to achieve superior multi-modal discrete data generation. By building on established DLLM foundations, MMaDA demonstrates how existing diffusion language models can be effectively adapted and enhanced for complex multi-modal synthesis tasks, opening new possibilities for foundation model reuse and extension in the diffusion domain.

\subsubsection{Scene Graph-to-Image Generation}
\begin{figure*}[ht]
\centering
    \includegraphics[width=14cm]{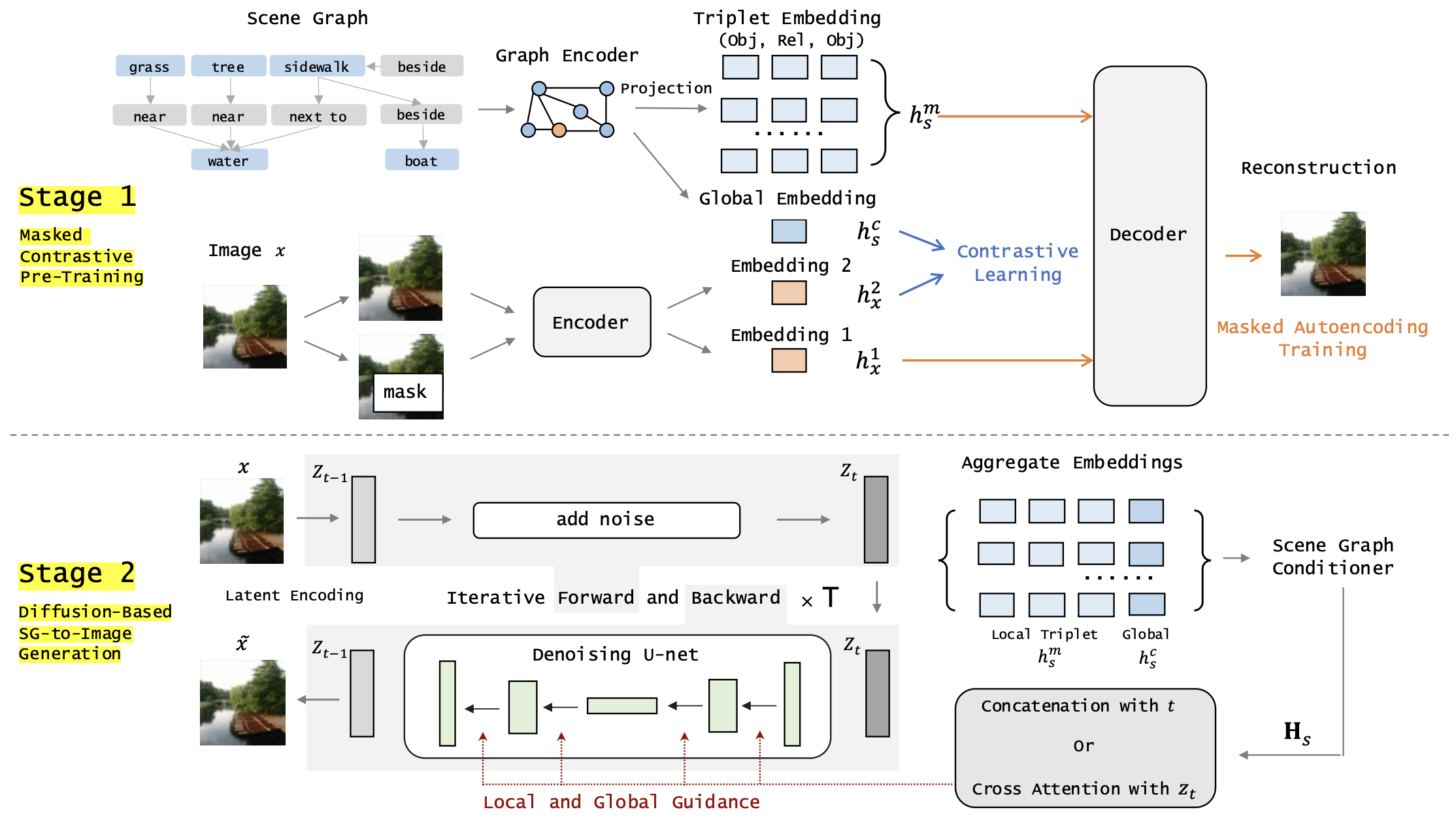}
        \caption{\textbf{SGDiff \cite{yang2022sgdiff} leverages masked contrastive pre-training for scene graph-based image diffusion generation.}}
\label{fig:sgdiff}
\end{figure*}

Despite text-to-image generation models has made exciting progress from natural language descriptions, they struggle to faithfully
reproduce complex sentences with many objects and relationships. Generating images from scene graphs (SGs) is an important and challenging task for generative models \cite{johnson2018image}. Traditional methods \cite{johnson2018image,li2019pastegan,herzig2020learning} mainly predict an image-like layout from SGs, then generate images based on the layout. However, such intermediate representations would lose some semantics in SGs, and recent diffusion models \cite{rombach2022high} are also unable to address this limitation. SGDiff \cite{yang2022sgdiff} proposes the first diffusion model specifically for image generation from scene graphs (\cref{fig:sgdiff}), and learns a continuous SG embedding to condition the latent diffusion model, which has been globally and locally semantically-aligned between SGs and images by the designed masked contrastive pre-training. SGDiff can generate images that better express the intensive and complex relations in SGs compared with both non-diffusion and diffusion methods. However, high-quality paired SG-image datasets are scarce and small-scale, how to leverage large-scale text-image datasets to augment the training or provide a semantic diffusion prior for better initialization is still an open problem.   



\subsubsection{Text-to-3D Generation}
3D content generation \cite{poole2022dreamfusion,lin2022magic3d,xu2022dream3d,jun2023shap} has been in high demand for a wide range of applications, including gaming, entertainment, and robotics simulation.
Augmenting 3D content generation with
natural language could considerably help with both novices and experienced artists. DreamFusion \cite{poole2022dreamfusion} adopts a pre-trained 2D text-to-image diffusion model to
perform text-to-3D synthesis.
It optimizes a randomly-initialized 3D model (a Neural Radiance Field, or NeRF) with a probability density
distillation loss, which utilizes a 2D diffusion model as a prior for optimization of a parametric image generator.
To obtain fast and high-resolution optimization of NeRF, Magic3D \cite{lin2022magic3d} proposes a two-stage diffusion framework built on cascaded low-resolution image diffusion prior and high-resolution latent diffusion prior.
In order to achieve high-fidelity 3D creation, Make-It-3D \cite{tang2023make} optimizes a neural radiance field by incorporating constraints from the reference image at the frontal view and diffusion prior at novel views, enhancing the coarse model into textured point clouds and increasing realism with diffusion prior and high-quality textures from the reference image.
ProlificDreamer \cite{wang2023prolificdreamer} presents Variational Score Distillation (VSD), optimizing a distribution of 3D scenes based on textual prompts as random variables to closely align the distribution of rendered images from all perspectives with a pretrained 2D diffusion model, using KL divergence as the measure.
IPDreamer \cite{zeng2023ipdreamer} further proposes a novel 3D object synthesis framework that enables users to create controllable and high-quality 3D objects effortlessly. It excels in synthesizing a high-quality 3D object which can greatly align with a provided
complex image prompt.

Modeling compositional 3D data distribution is a fundamental and critical task for generative models. Current feed-forward methods \citep{shue20233d,shi2024mvdream}  are primarily capable of generating single objects and face challenges when creating more complex scenes containing multiple objects due to limited training data. 
Recently, a series of learnable-layout compositional methods have been proposed \citep{epstein2024disentangled,vilesov2023cg3d,han2024reparo,chen2024comboverse,gao2024graphdreamer} . These methods combine multiple object-ad-hoc radiance fields and then optimize the positions of the radiance fields from external feedback. For example,
\citet{epstein2024disentangled} propose learning a distribution of reasonable layouts based solely on the knowledge from a large pre-trained text-to-image model.   
\citet{vilesov2023cg3d} introduce an optimization method based on Monte-Carlo sampling and physical constraints. 

\subsubsection{Text-to-Motion Generation}
Human motion generation is a fundamental task in computer animation, with applications covering
from gaming to robotics \cite{zhang2022motiondiffuse}. The generate motion is usually a sequence of human poses represented by joint rotations and positions.
Motion Diffusion Model (MDM) \cite{tevet2022human} adapts a classifier-free diffusion-based generative model for the human motion generation, which is transformer-based, combining insights from motion generation literature, and regularizes the model with geometric losses on the locations and velocities of the motion. FLAME \cite{kim2022flame} involves a transformer-based diffusion to better handle motion data, which manages variable-length motions and well attend to free-form text. Notably, it can edit the parts of the motion, both frame-wise and joint-wise, without any fine-tuning. 

\begin{figure*}[htp]
\centering
\footnotesize
\includegraphics[width=0.75\textwidth]{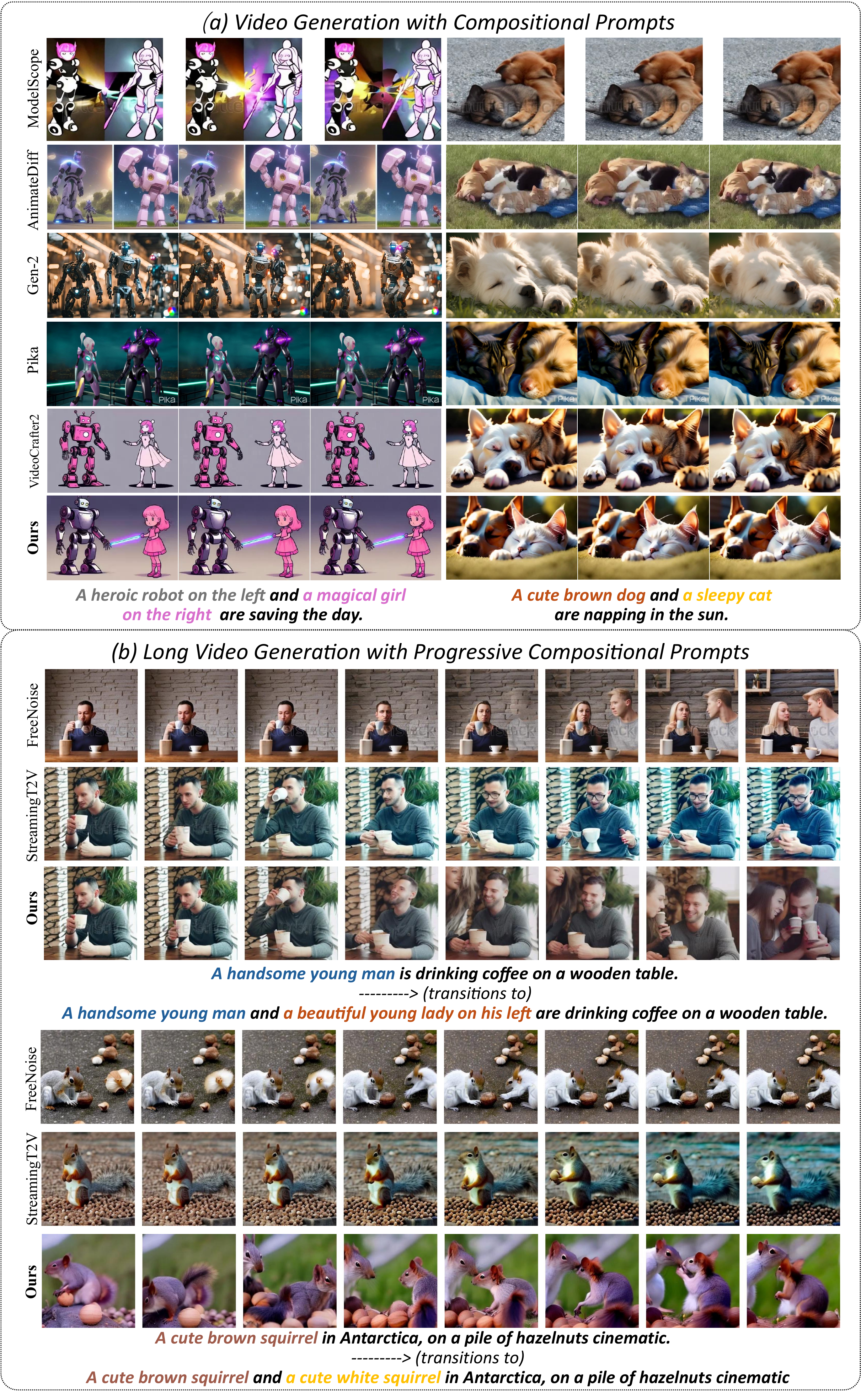}
\caption{\textbf{Comparing VideoTetris \cite{tian2024videotetris} with open-sourced or commercial T2V models in short and long video generation.}}
\label{fig:videotetris}
\end{figure*}

\subsubsection{Text-to-Video Generation} \label{text-to-video}
Tremendous recent progress in text-to-image diffusion-based generation motivates the development of text-to-video generation \cite{singer2022make,ho2022imagen,wu2022tune}. Make-A-Video \cite{singer2022make} proposes to extend a diffusion-based text-to-image model
to text-to-video through a spatiotemporally factorized diffusion model. It leverages joint text-image prior to bypass the need for paired text-video data, and further presents super-resolution strategies for
high-definition, high frame-rate text-to-video generation. Imagen Video \cite{ho2022imagen}  generates
high definition videos by designing a cascaded video diffusion models, and transfers some findings that work well in the text-to-image setting to video generation,
including frozen T5 text encoder and classifier-free guidance.
Tune-A-Video \cite{wu2022tune} introduces one-shot video tuning for text-to-video generation, which eliminates the burden
of training with large-scale video datasets. It employs efficient attention tuning and structural inversion to significantly enhance temporal consistency.
Text2Video-Zero \cite{khachatryan2023text2video} achieves zero-shot text-to-video synthesis using a pretrained text-to-image diffusion model, ensuring temporal consistency through motion dynamics in latent codes and cross-frame attention. Its goal is to enable affordable text-guided video generation and editing without additional fine-tuning.
FateZero \cite{qi2023fatezero} is the first framework for temporal-consistent zero-shot text-to-video editing using pre-trained text-to-image diffusion model. It fuses the attention maps in the DDIM inversion and generation processes to maximally preserve the consistency of motion and structure during editing.
ContextDiff \cite{yang2024crossmodal} incorporates the cross-modal context information about the interactions between text condition and video sample into forward and reverse processes, forming a forward-backward consistent video diffusion model for text-to-video generation.

Most of text-to-video diffusion models are trained on fixed-size video datasets, and thus are often limited to generating a relatively small number of frames, leading to significant degradation in quality when tasked with generating longer videos.
Several advancements \citep{zhuang2024vlogger,streamingt2v,tian2024videotetris} have sought to overcome this limitation through various strategies. Vlogger \citep{zhuang2024vlogger} employs a masked diffusion model for conditional frame input facilitating longer video generation, and StreamingT2V \citep{streamingt2v} utilizes a ControlNet-like conditioning mechanism to enable auto-regressive video generation.  Recent VideoTetris \cite{tian2024videotetris} introduces a Spatio-Temporal Compositional Diffusion method for handling scenes with multiple objects and following progressive complex prompts (i.e., compositional text-to-video generation). Besides, VideoTetris develops a new video data preprocessing method and a consistency regularization method called Reference Frame Attention to improve auto-regressive long video generation through enhanced motion dynamics and prompt semantics. Qualitative comparisons in \cref{fig:videotetris} show that VideoTetris not only generates superior quality compositional videos, but also produces high-quality long videos that align with compositional prompts while maintaining the best consistency.

\subsubsection{Text-to-Audio Generation}
Text-to-audio generation is the task to transform normal language texts to voice outputs \cite{wu2021itotts,levkovitch2022zero}. Grad-TTS \cite{popov2021grad} presents a novel text-to-speech model with a score-based decoder and diffusion models. It gradually transforms noise
predicted by the encoder and is further aligned with text input
by the method of Monotonic Alignment Search \cite{rabiner1989tutorial}. Grad-TTS2 \cite{kim2022guided} improves Grad-TTS in an adaptive way.
Diffsound \cite{yang2022diffsound} presents a non-autoregressive decoder based on the discrete diffusion model \cite{sohl2015deep,austin2021structured}, which predicts all the mel-spectrogram tokens in every single step, and then refines the predicted tokens in the following steps.
EdiTTS \cite{tae2021editts} leverages the score-based text-to-speech model to refine a mel-spectrogram prior that is coarsely modified.
Instead of estimating the gradient
of data density, ProDiff \cite{huang2022prodiff} parameterizes the denoising diffusion model by directly predicting the clean data.
\subsection{Temporal Data Modeling}
\subsubsection{Time Series Imputation}
Time series data are widely used with many important real-world applications \cite{Oreshkin2020N-BEATS:,zhang2022cross,eldele2021time,yang2022unsupervised}. Nevertheless, time series usually contain missing values for multiple reasons, caused by mechanical or artificial errors \cite{silva2012predicting,yi2016st,tan2013tensor}.
Recent years, imputation methods have been greatly for both deterministic imputation \cite{cao2018brits,che2018recurrent,luo2018multivariate} and probabilistic imputation \cite{fortuin2020gp}, including diffusion-based approaches.
Conditional Score-based Diffusion models for Imputation (CSDI) \cite{tashiro2021csdi} presents a novel time series imputation method that leverages score-based diffusion models. Specifically, for the purpose of exploiting correlations within temporal data, it adopts the form of self-supervised training to optimize diffusion models.
Its application in some real-world datasets reveals its superiority over previous methods.  Controlled Stochastic Differential Equation (CSDE) \cite{park2021neural} proposes a novel probabilistic framework for modeling stochastic dynamics with a neural-controlled
stochastic differential equation. Structured State Space Diffusion (SSSD) \cite{alcaraz2022diffusion} integrates conditional diffusion models and structured state-space models \cite{gu2021efficiently} to particularly capture long-term dependencies in time series.
It performs well in both time series imputation and forecasting tasks.

\begin{figure*}[t]
	\centering
        \includegraphics[width=\textwidth]{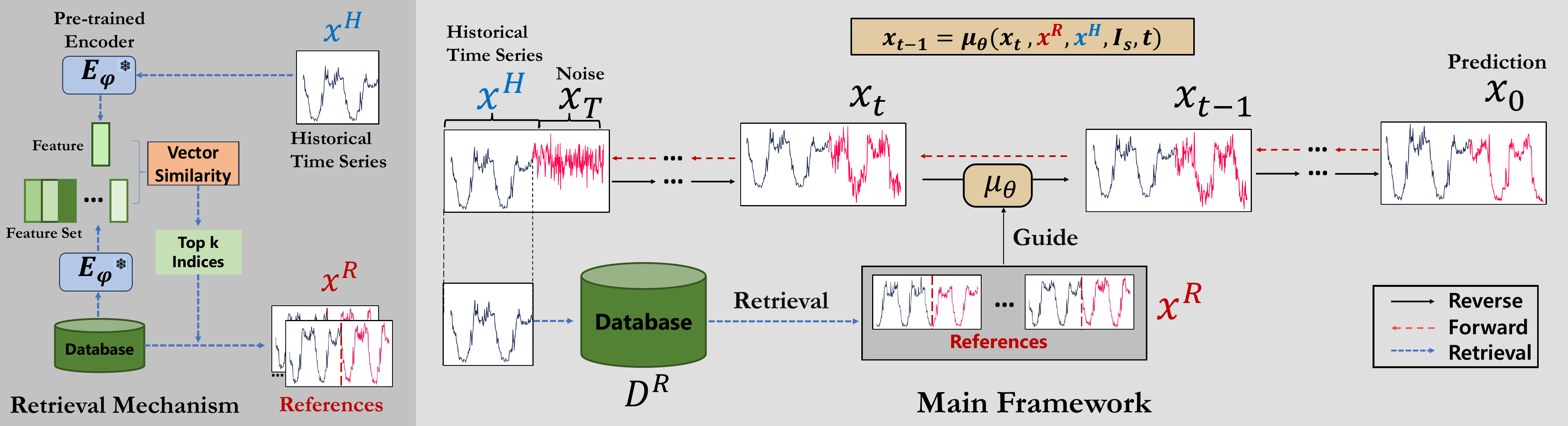}  
\caption{Overview of retrieval-augmented diffusion models
for time series forecasting (RATD \cite{liu2024retrieval}). The historical time series $x^H$ is inputted into the retrieval module to for the corresponding references $x^R$. After that, $x^H$ is concatenated with the noise as the main input for the model $\mu_{\theta}$. $x^R$ will be utilized as the guidance for the denoising process.}
\label{fig:ratd}
\end{figure*}

\subsubsection{Time Series Forecasting}
Time series forecasting is the task of forecasting or predicting the future value over a period of time. Neural methods have recently become widely-used for solving the prediction problem with univariate point forecasting methods \cite{oreshkin2019n} or
univariate probabilistic methods \cite{salinas2019high}. In
the multivariate setting, we also have point forecasting methods \cite{li2019enhancing} as well as probabilistic
methods, which explicitly model the data
distribution using Gaussian copulas \cite{salinas2020deepar},
GANs \cite{yoon2019time}, or normalizing flows \cite{rasul2020multivariate}.
TimeGrad \cite{rasul2021autoregressive} presents an autoregressive model for forecasting multivariate probabilistic time series, which samples from the data
distribution at each time step through estimating its gradient. It utilizes diffusion probabilistic models, which are closely connected with score matching and energy-based methods. Specifically, it learns gradients by optimizing a variational bound on the data likelihood and transforms white noise into a sample of the distribution of interest through a Markov
chain using Langevin sampling \cite{song2019generative} during inference time. To handle complex time series forecasting, as illustrated in \cref{fig:ratd}, Liu et al. (2024) for the first time introduce Retrieval-
Augmented Time series Diffusion (RATD) \cite{liu2024retrieval}, allowing for greater utilization of the dataset
and providing meaningful guidance in the denoising process. 
\subsubsection{Waveform Signal Processing}
In electronics, acoustics, and some related fields, the waveform of a signal is denoted by the shape of its graph as a function of time, independent of its time and magnitude scales.
WaveGrad \cite{chen2020wavegrad} introduces a conditional model for waveform generation that estimates gradients of the data density. It receives a Gaussian white noise signal as input and iteratively refines the signal with a gradient-based sampler.
WaveGrad naturally trades inference speed for sample quality by adjusting the number of refinement steps, and
make a connection between non-autoregressive and autoregressive models with respect to audio quality.
DiffWave \cite{kong2020diffwave} presents a versatile and effective diffusion probabilistic model for conditional or unconditional waveform generation.
The model is non-autoregressive and is efficiently trained by
optimizing a variant of variational bound on the data likelihood. %
Moreover, it produces high-fidelity audio in different waveform generation tasks, such as class-conditional generation and unconditional generation.
\subsection{Robust Learning}
Robust learning is a class of defense methods that help learning networks that are robust to adversarial perturbations or noises \cite{pidhorskyi2020adversarial,blau2022threat,wu2022guided,wang2022guided,nie2022diffusion,yoon2021adversarial}.
While adversarial training \cite{madry2018towards} is viewed as a
standard defense method against adversarial attacks for image classifiers, adversarial purification has
shown significant performances as an alternative defense method \cite{yoon2021adversarial}, which purifies attacked images into clean images with a standalone purification model.
Given an adversarial example, DiffPure \cite{nie2022diffusion} diffuses it with a small amount of noise following a forward diffusion process and then restores the clean image with a reverse generative process.
Adaptive Denoising Purification (ADP) \cite{yoon2021adversarial} demonstrates that an EBM trained with denoising score matching \cite{vincent2011connection} can effectively purify attacked images within just a few steps.
It further proposes an effective randomized purification scheme, injecting random noises into images before purification.
Projected Gradient Descent (PGD) \cite{blau2022threat} presents a novel stochastic diffusion-based pre-processing robustification, which aims to be a model-agnostic adversarial defense and yield a high-quality denoised outcome. In addition, some works propose to apply a guided diffusion process for advanced adversarial purification \cite{wang2022guided,wu2022guided}.

\subsection{Interdisciplinary Applications}
\subsubsection{Drug Design and Life Science}
Graph Neural Networks \cite{hamilton2017inductive,wu2020comprehensive,zhou2020graph,yang2020dpgn} and corresponding representation learning \cite{hamilton2017representation} techniques have achieved great success \cite{bielak2021graph,thakoor2021bootstrapped,xu2021self,yang2022omni,zhu2020deep,wu2020graph} in many areas, including modeling molecules/proteins in various tasks ranging from property prediction \cite{gilmer2017neural,duvenaud2015convolutional} to molecule/protein generation \cite{jin2018junction,shi2020graphaf,jumper2021highly,luo2021predicting}, where a molecule is naturally represented by a node-edge graph.
On one hand, recent works propose to pre-train GNN/transformer \cite{zhou2023uni,luo2022one} specifically for molecules/proteins with biomedical or physical insights \cite{liu2022molecular,zaidi2022pre}, and achieve remarkable results. On the other hand, more works begin to utilize graph-based diffusion models for enhancing molecule or protein generation. Torsional diffusion \cite{jing2022torsional} presents a new diffusion framework that makes operations on the space of torsion angles with a diffusion process on the hyperspace and an extrinsic-to-intrinsic scoring model.
GeoDiff \cite{xu2021geodiff} demonstrates that Markov chains evolving with equivariant Markov kernels can produce an invariant distribution, and further design blocks for the Markov kernels to preserve the desirable equivariance property. There are also other works incorporate the equivariance property into 3D molecule generation \cite{hoogeboom2022equivariant}  and protein generation \cite{anand2022protein,berman2000protein}.
Motivated by the classical force field methods for simulating molecular dynamics, ConfGF \cite{shi2021learning} directly estimates the gradient fields of the log density of atomic coordinates in molecular conformation generation.
\begin{figure*}[htp]
\centering
\footnotesize
\includegraphics[width=0.98\textwidth]{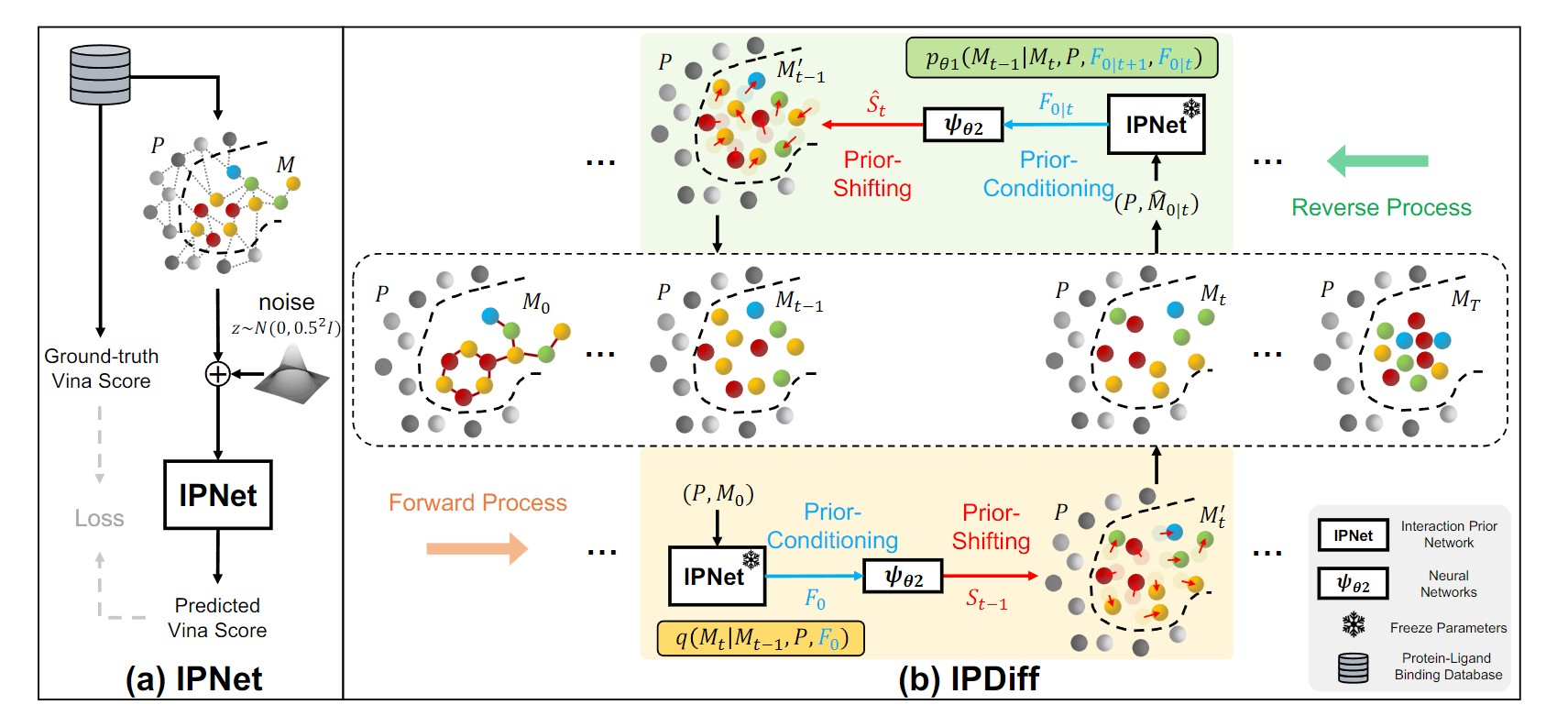}
\caption{\textbf{IPDiff \cite{huang2024proteinligand} incorporates protein-ligand interactions into both forward and reverve processes of molecular diffusion models.}}
\label{fig:ipdiff}
\end{figure*}

\begin{figure*}[htp]
\centering
\footnotesize
\includegraphics[width=0.8\textwidth]{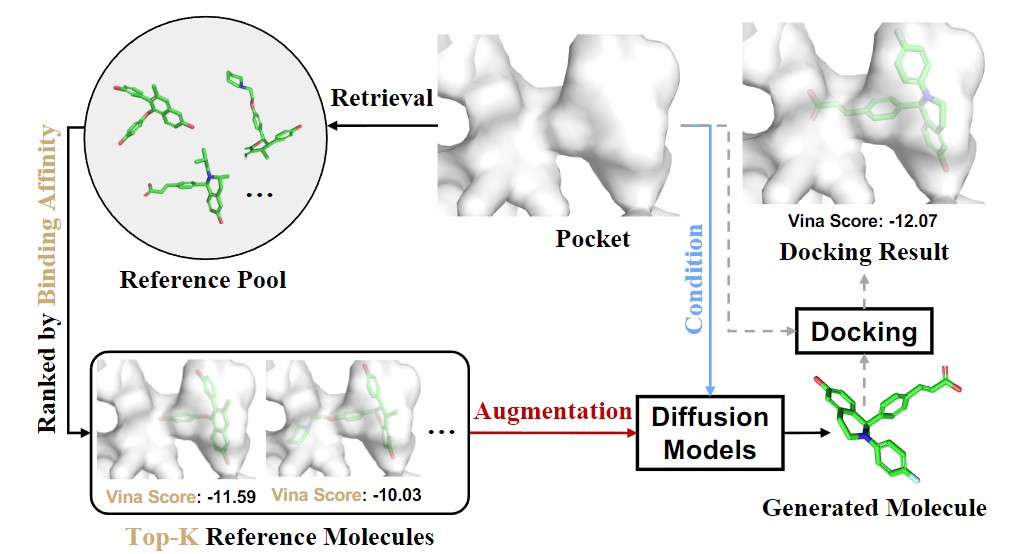}
\caption{\textbf{IRDiff \cite{huang2024interactionbased} designs a interaction-based retrieval-augmented generation frameowrk for SBDD}.}
\label{fig:irdiff}
\end{figure*}

Recently, given a target protein, the design of 3D small drug molecules that can closely bind to the target begins to be promoted by diffusion models \cite{guan20233d,huang2024proteinligand,huang2024interactionbased}. 
IPDiff \cite{huang2024proteinligand} proposes a novel 3D molecular diffusion model for structure-based drug design (SBDD). As illustrated in \cref{fig:ipdiff}, the pocket-ligand interaction is explicitly considered in both forward and reverse processes with the proposed prior-conditioning and prior-shifting mechanisms.
Notably, IPDiff beats all previous diffusion-based and autoregressive generation models regarding binding-related metrics and molecular properties.
BindDM \cite{huang2024binddm} proposes a hierarchical complex-subcomplex diffusion model for SBDD tasks, which incorporates essential binding-adaptive subcomplex for 3D molecule diffusion generation. 
IRDiff \cite{huang2024proteinligand} proposes an interaction-based retrieval-augmented 3D molecular
diffusion model named IRDIFF for SBDD tasks. 
As deonstrated in \cref{fig:irdiff}, this model
guides 3D molecular generation using informative external
target-aware references, designing two novel
augmentation mechanisms, i.e., retrieval augmentation and
self augmentation, to incorporate essential protein-molecule
binding structures for target-aware molecular generation.

There are also studies that use diffusion models for protein generation, such as DiffAb. DiffAb \cite{luo2022antigen} proposes for the first time a diffusion-based 3D antibody design framework that models both the sequence and structure of the complementarity-determining regions (CDRs) that determine antibody complementarity. Experiments show that DiffAb can be used for various antibody design tasks, such as jointly generating sequence-structure, designing CDRs with fixed frameworks, and optimizing antibodies.
SMCDiff \cite{trippe2022diffusion} proposes to first
learn a distribution over diverse and longer protein backbone structures via an E(3)-equivariant graph neural network, and then efficiently samples scaffolds from this distribution given a motif. The generation results demonstrates the designed backbones is well aligned with AlphaFold2-predicted structures.
\subsubsection{Material Design}
Solid state materials are the critical
foundation of numerous key technologies \cite{butler2018machine}.
Crystal Diffusion Variational Autoencoder (CDVAE) \cite{xie2021crystal} incorporates stability as an inductive bias by proposing a noise conditional score network, which simultaneously utilizes permutation, translation, rotation, and periodic invariance properties.
Luo et al. (2022) \cite{luo2022antigen} model
sequences and structures of complementarity-determining regions with equivariant diffusion, and explicitly
target specific antigen structures to generate antibodies at atomic resolution.

\subsubsection{Medical Image Reconstruction}
An inverse problem is to recover an unknown signal from observed measurements, and it is an important problem in medical image reconstruction of  Computed Tomography (CT) and Magnetic Resonance Imaging (MRI) \cite{song2021solving,chung2022mr,peng2022towards,xie2022measurement,chung2022come}. Song et al. (2021) \cite{song2021solving} utilize a
score-based generative model to reconstruct an image consistent with both the prior and the observed measurements.
Chung et al. (2022) \cite{chung2022score} train a continuous time-dependent score function with denoising score matching, and iterate between the numerical SDE solver and data consistency step for reconstruction at the evaluation stage.
Peng et al. (2022) \cite{peng2022towards} perform MR
reconstruction by gradually guiding the reverse-diffusion process given observed
k-space signal, and propose a coarse-to-fine sampling algorithm for efficient sampling.

\section{Future Directions}
\label{sec8}
Research on diffusion models is in its early stages, with much potential for improvement in both theoretical and empirical aspects.
As discussed in early sections, key research directions include efficient sampling and improved likelihood, as well as exploring how diffusion models can handle special data structures, interface with other types of generative models, and be tailored to a range of applications. 
In addition, we foresee that future research on diffusion models will likely expand to the following avenues.
\paragraph{Revisiting Assumptions}
Numerous typical assumptions in diffusion models need to be revisited and analyzed.  
For example, the assumption that the forward process of diffusion models completely erases any information in data and renders it equivalent to a prior distribution may not always hold.
In reality, complete removal of information is unachievable in finite time. 
It is of great interest to understand when to halt the forward noising process in order to strike a balance between sampling efficiency and sample quality \cite{franzese2022much}. 
Recent advances in Schrödinger bridges and optimal transport \cite{chen2021likelihood,de2021simulating,de2021diffusion,shi2022conditional,song2022applying} provide promising alternative solutions, suggesting new formulations for diffusion models that are capable of converging to a specified prior distribution in finite time.

\paragraph{Theoretical Understanding}
Diffusion models have emerged as a powerful framework, notably as the only one that can rival generative adversarial networks (GANs) in most applications without resorting to adversarial training.
Key to harnessing this potential is an understanding of why and when diffusion models are effective over alternatives for specific tasks. 
It is important to identify which fundamental characteristics differentiate diffusion models from other types of generative models, such as variational autoencoders, energy-based models, or autoregressive models. 
Understanding these distinctions will help elucidate why diffusion models are capable of generating samples of excellent quality while achieving top likelihood. 
Equally important is the need to develop theoretical guidance for selecting and determining various hyperparameters of diffusion models systematically. 

\paragraph{Latent Representations}
Unlike variational autoencoders or generative adversarial networks, diffusion models are less effective for providing good representations of data in their latent space. 
As a result, they cannot be easily used for tasks such as manipulating data based on semantic representations. 
Furthermore, since the latent space in diffusion models often possesses the same dimensionality as the data space, sampling efficiency is negatively affected and the models may not learn the representation schemes well \cite{jing2022subspace}.


\paragraph{AIGC and Diffusion Foundation Models}
From Stable Diffusion to ChatGPT, Artificial Intelligence Generated Content (AIGC) has gained much attention in both academic and industrial circles. Generative Pre-Training is the core technique in GPT-1/2/3/4 \cite{radford2018improving,radford2019language,ouyang2022training,openai2023gpt4} and (Visual) ChatGPT \cite{wu2023visual}, which exhibits promising generation performance and surprising emergent abilities \cite{wei2022emergent} equipped with Large Language Models (LLMs) \cite{touvron2023llama} and Visual Foundation Models \cite{bommasani2021opportunities,yu2022coca,yuan2021florence}. 

It is interesting to transfer the generative pre-training (decoder-only) from GPT series to diffusion model class, evaluate the diffusion-based generation performance at scale, and analyse the emergent abilities of diffusion foundation models. Recent advances have demonstrated the potential of building specialized diffusion foundation models for different domains and applications. MMaDA \cite{yang2025mmada} exemplifies this trend by leveraging diffusion-based language model foundations to create a multi-modal generation framework that exhibits emergent cross-modal reasoning capabilities. Similarly, TraceRL \cite{wang2025revolutionizing} explores how diffusion foundation models can be adapted for reinforcement learning scenarios, generating coherent action sequences while maintaining the scalable properties of foundation models.

Furthermore, combining LLMs with diffusion models has been proved to be a new promising direction \cite{yang2024mastering,yang2024editworld}, opening up possibilities for more sophisticated AIGC systems \cite{liu2025preacher} that leverage the complementary strengths of both paradigms. These developments suggest that the future of AIGC lies not only in scaling individual model types but also in creating hybrid architectures that can harness different generative principles for enhanced performance and broader applicability.
\section{Conclusion}
\label{sec9}
We have provided a comprehensive look at diffusion models from various angles. We began with a self-contained introduction to three fundamental formulations: DDPMs, SGMs, and Score SDEs. We then discussed recent efforts to improve diffusion models, highlighting three major directions: sampling efficiency, likelihood maximization, and new techniques for data with special structures. We also explored connections between diffusion models and other generative models and outlined potential benefits of combining the two. A survey of applications across six domains illustrated the wide-ranging potential of diffusion models. Finally, we outlined possible avenues for future research.


\bibliographystyle{ACM-Reference-Format}
\bibliography{main}

\end{document}